\documentclass[journal]{IEEEtran}
\ifCLASSINFOpdf
\else
\fi

\usepackage{amsmath}
\usepackage{amssymb}
\usepackage{booktabs}
\usepackage{xcolor}
\usepackage{multirow}
\usepackage{graphicx}

\usepackage{url}

\usepackage{fancyhdr}
\usepackage{mathrsfs}
\usepackage{amsthm}
\usepackage{mathtools}

\def\bbR{\mathbb{R}}

\newcommand{\R}{\mathbb{R}}

\definecolor{darkgreen}{rgb}{0.0, 0.56, 0.0}

\usepackage[inline]{enumitem}
\usepackage{array}
\newcolumntype{H}{>{\setbox0=\hbox\bgroup}c<{\egroup}@{}}

\newcommand{\gc}{\textcolor{black}}
\newcommand{\mt}{\textcolor{black}}

\usepackage{rotating}

\usepackage{hyperref}

\colorlet{blueblack}{blue} 

\begin{document}
%
\title{Graph Neural Networks for Graph Drawing}
%
%
%

\author{Matteo~Tiezzi, 
Gabriele~Ciravegna
and~Marco~Gori~\IEEEmembership{Fellow,~IEEE}
\IEEEcompsocitemizethanks{\IEEEcompsocthanksitem M. Tiezzi and M. Gori are with the Department of Information Engineering and Mathematics, University of Siena, 53100 Siena, Italy.  
M. Gori and G. Ciravegna are with MAASAI, Inria, I3S, CNRS, Université Côte d'Azur, Nice, France.\protect\\
Matteo Tiezzi is the corresponding author (mtiezzi@diism.unisi.it).

\IEEEcompsocthanksitem Accepted for publication in  Transaction of Neural Networks and Learning Systems (TNNLS), Special Issue on Deep Neural Networks for Graphs: Theory, Models, Algorithms and Applications.

(DOI: \url{https://doi.org/10.1109/tnnls.2022.3184967})
}
\thanks{2022 IEEE. Personal use of this material is permitted. Permission from IEEE must be obtained for all other uses, in any current or future media, including reprinting/republishing this material for advertising or promotional purposes, creating new collective works, for resale or redistribution to servers or lists, or reuse of any copyrighted component of this work in other works.}
}

\maketitle

\begin{abstract}


Graph Drawing techniques have been developed in the last few years with the purpose of producing aesthetically pleasing node-link layouts. Recently, the employment of differentiable loss functions has paved the road to the massive usage of Gradient Descent and related optimization
algorithms. In this paper, we propose a novel framework for the development of {\em Graph Neural Drawers} (GND), machines that rely on neural computation for constructing efficient and complex maps. GNDs are Graph Neural Networks (GNNs) whose learning process can be driven by any provided loss function, such as the ones commonly employed in Graph Drawing. Moreover, we prove that this mechanism can be guided by loss functions computed by means of Feedforward Neural Networks, on the basis of supervision hints that express beauty properties, like the minimization of crossing edges. In this context, we show that GNNs can nicely be enriched by positional features to deal also with unlabelled vertexes. We provide a proof-of-concept by constructing a loss function for the edge-crossing and provide quantitative and qualitative comparisons among different GNN models working under the proposed framework. 
\end{abstract}


%
\IEEEpeerreviewmaketitle

\section{Introduction}

\IEEEPARstart{V}{isualizing} complex relations and interaction patterns among entities is a crucial task, given the increasing interest in structured data representations\cite{bronstein2021geometric}. The Graph Drawing \cite{battista1998graph} literature aims at developing algorithmic techniques to construct drawings of graphs -- i.e. mathematical structures capable to efficiently represent the aforementioned relational concepts with nodes and edges connecting them -- for example via the node-link paradigm~\cite{saket2014node, tamassia2013handbook, didimo2019survey}. 
The readability of graph layouts can be evaluated following some aesthetic criteria such as  the number of crossing edges, minimum  crossing angles, community preservation, edge length variance,  etc. \cite{ahmed2020graph}.
The final goal is to find suitable coordinates for the node positions, and this often requires to explicitly express and combine these criteria through complicated mathematical formulations ~\cite{cox2008multidimensional}. Moreover, effective approaches such as energy-based models~ \cite{jacomy2014forceatlas2, kamada1989algorithm} or spring-embedders~\cite{frick1994fast, brandes2008experimental} require hands-on expertise and trial and error processes to achieve certain desired visual properties. 
Additionally, such methods define loss or energy functions that must be optimized for each new graph to be drawn, often requiring to adapt  algorithm-specific parameters.
Lately, two interesting directions have emerged in the Graph Drawing community. The former one leverages the power of Gradient Descent to explore the manifold given by pre-defined loss functions or combinations of them. Stochastic Gradient Descent (SGD) can be used to move sub-samples of vertices couples in the direction of the gradient of spring-embedder losses~\cite{zheng2018graph} substituting complicated techniques such as Majorization~\cite{de1988convergence}.
This approach has been extended to arbitrary optimization goals, or combinations of them, which can be optimized via Gradient Descent if the corresponding criterion can be expressed via  smooth functions~\cite{ahmed2020graph}.
The latter novel direction consists in the exploitation of Deep Learning models. Indeed, the flexibility of neural networks and their approximation capability can come in handy also when dealing with the Graph Drawing scenario. For instance,
Neural networks are capable to learn the layout characteristics from plots produced by other graph drawing techniques~\cite{brandes2006eigensolver, wang2019deepdrawing}, as well as the underlying distribution of data~\cite{kwon2019deep}. 
Very recently, the node positions produced by graph drawing frameworks~\cite{brandes2006eigensolver} have been used as an input to Graph Neural Networks (GNNs)~\cite{DBLP:journals/tnn/ScarselliGTHM09, DBLP:journals/tnn/WuPCLZY21} to produce pleasing layout that minimize combinations of aesthetic losses~\cite{wang2021deepgd}.

We propose a framework, \textit{Graph Neural Drawers} (GND), which embraces both the aforementioned directions. We borrow the representational capability and computational efficiency of neural networks to prove that (1) differentiable loss functions guiding the common Graph Drawing pipeline can be provided directly by a neural network, a Neural Aesthete, \gc{even when the required aesthetic criteria cannot be directly optimized.}   
In particular, we propose a proof-of-concept where we focus on the criteria of edge crossing, proving that a neural network can learn to identify if two arcs are crossing or not \gc{and provide a differentiable loss function towards non-intersection}. \gc{Otherwise, in fact, this simple aesthetic criterion cannot be achieved through direct optimization, because it is non-differentiable. Instead, the Neural Aesthete} provides a useful and flexible gradient direction that can be exploited by (Stochastic) Gradient Descent methods.
Moreover, (2) we prove that \mt{GNNs}, even in the non-attributed graph scenario if enriched with appropriate node positional features, can be used to process the topology of the input graph with the purpose of 
mapping the obtained node representation in a 2D layout. We compare various commonly used GNN models ~\cite{kipf2017semisupervised, velivckovic2018graph, xu2019powerful}, proving that the proposed framework is flexible enough to give these models the ability to learn a wide variety of solutions. In particular, GND is capable to draw graphs (\textit{i}) from supervised coordinates, i.e. emulating Graph Drawing Packages, (\textit{ii}) minimizing common aesthetic loss functions and, additionally,  (\textit{iii}) by descending towards the gradient direction provided by the Neural Aesthete. 

The paper is organized as follows. Section \ref{sec:related} introduces some basics on the Graph Drawing scenario as well as references on Gradient Descent approaches.  Section \ref{sec:proposal_1} introduces the Neural Aesthete and provides a proof-of-concept on the edge crossing task. Section \ref{sec:proposal_2} describes the requirements of using GNNs to draw graphs in the non-attributed scenario, as well as the problem definition and the experimental evaluation. Finally, conclusions are drawn in Section \ref{sec:conclusion}.

\section{Related work}
\label{sec:related}


There exists a large variety of methods in literature to improve graph readability. 
A straight-forward approach, that has been proved to be effective in improving the human understanding of the graph topology, consists in minimizing the number of crossing edges~\cite{purchase1997aesthetic}. However, the computational complexity of the problem is NP-hard, and several authors proposed complex solutions and algorithms to address this problem \cite{abrego2013rectilinear}. In~\cite{shabbeer2010optimal}, authors employ an Expectation-Maximization algorithm based on the decision boundary surface built by a Support Vector Machine. The underlying idea is that two edges do not cross if there is a line separating the node coordinates. Further aesthetic metrics have been explored, such as the minimization of node occlusions~\cite{chrobak1996convex}, neighborhood preservation, the maximization of crossing edges angular width~\cite{kaufmann2018heuristic} and many more~\cite{gibson2013survey, ahmed2020graph}. 
Given the graph drawing categorization depicted in surveys~\cite{gibson2013survey} (i.e. force-directed, dimension reduction, multi-level techniques), interesting and aesthetically pleasing layouts are produced by methods regarding a graph as a physical system, with forces acting on nodes with attraction and repulsion dynamics up to a stable equilibrium state~\cite{kamada1989algorithm}. Force-directed techniques inspired many subsequent works, from spring-embedders~\cite{fruchterman1991graph} to energy-based approaches~\cite{jacomy2014forceatlas2}.
The main idea is to obtain the final layout of the graph minimizing the \textit{Stress} function (see Eq. \ref{eq:stress}). The forces characterizing this formulation can be thought of as springs connecting pairs of nodes. This very popular formulation, exploited for graph layout in the seminal work by Kamada and Kawai~\cite{kamada1989algorithm}, was optimized with the localized 2D Newton-Raphson method. Further studies employed various complicated optimization techniques, such as the \textit{Stress Majorization} approach which produces graph layout through an iterative resolution of simpler functions, as proposed by Gasner et al.~\cite{gansner2004graph}.
In this particular context, some recent contributions highlighted the advantages of using gradient-based methods to solve graph drawing tasks. The SGD method was successfully applied to efficiently minimize the Stress function in Zheng et al.~\cite{zheng2018graph}, displacing pairs of vertices following the direction of the gradient, computed in closed form. 
A recent framework, $(GD)^2$, leverages Gradient Descent to optimize several readability criteria at once~\cite{ahmed2020graph}, as long as the criterion can be expressed by smooth functions. Indeed, thanks to the powerful auto-differentiation tools available in modern machine learning frameworks~\cite{paszke2017automatic},  several criteria such as ideal edge lengths, Stress Majorization, node occlusion, angular resolution and many others can be easily optimized.
We build our first contribution upon these ideas, proving that neural networks can be used to learn decomposed single criteria (i.e., edge crossing) approximating smooth functions, with the purpose of providing a useful descent direction to optimize the graph layout. 


Deep Learning has been successfully applied to data belonging to the non-Euclidean domain, e.g. graphs, in particular thanks to \mt{GNNs}~\cite{DBLP:journals/spm/BronsteinBLSV17, DBLP:journals/tnn/WuPCLZY21}. The seminal work by Scarselli et al.~\cite{DBLP:journals/tnn/ScarselliGTHM09} proposes a model
based on an information diffusion process involving the whole graph, fostered by the iterative application of an \textit{aggregation function} among neighboring nodes up to an equilibrium point.
The simplification of this computationally expensive mechanism was the goal of several works which leverage alternative recurrent neural models~\cite{DBLP:journals/corr/LiTBZ15} or constrained fixed-point formulations. This problem was solved via Reinforcement Learning algorithms as done in \mt{Stochastic Steady-state Embedding (SSE)}\cite{DBLP:conf/icml/DaiKDSS18} or cast to the Lagrangian framework and optimized by a gradient descent-ascent approach, like in \mt{Lagrangian-Propagation GNNs (LP-GNNs)}~\cite{DBLP:conf/ecai/TiezziMMMG20}, even with the advantage of multiple layers of feature extraction~\cite{tiezzi2021deep}.
The iterative nature of the aforementioned models inspired their classification into the umbrella of RecGNNs in recent surveys~\cite{DBLP:journals/tnn/WuPCLZY21, DBLP:journals/nn/BacciuEMP20}. 

In addition to RecGNNs, several other flavours of GNN models have been proposed, such as the ConvGNNs~\cite{kipf2020deep} or Attentional GNNs~\cite{velivckovic2018graph, monti2017geometric, bresson2017residual}. All such models fit into the powerful concept of message exchange, the foundation on which is built the very general framework of Message Passing Neural Networks (MPNNs)\cite{gilmer2017neural, gilmer2020message}.

Recent works analyze the expressive capabilities of GNNs and their aggregation functions, following the seminal work on graph isomorphism by Xu et al. ~\cite{DBLP:journals/corr/abs-1810-00826}. The model proposed by the authors,  Graph Isomorphism Network (GIN), leverage an injective aggregation function with  the same representational power of the Weisfeiler-Leman (WL) Test~\cite{weisfeiler1968reduction}. Subsequent works (sometimes denoted with the term WL-GNNs) try to capture higher-order graph properties \cite{morris2019weisfeiler,maron2018invariant, corso2020principal, bodnar2021weisfeiler}.
Bearing in mind that we deal with the non-attributed graph scenario, i.e., graphs lacking node features, we point out the importance of the nodal feature choice. Several recent works investigated this problem~\cite{cui2021positional, srinivasan2019equivalence, murphy2019relational}. We borrow the highly expressive Laplacian Eigenvector-based positional features described by Dwivedi et al. \cite{dwivedi2020benchmarking}.

There have been some early attempts in applying Deep Learning models and GNNs to the Graph Drawing scenario.
Wang et al.~\cite{ wang2019deepdrawing} proposed a graph-based LSTM model able to learn and generalize the coordinates patterns produced by other graph drawing techniques. 
\mt{ However, this approach is limited by the fact that the model drawing ability is highly dependent on the training data, such that processing different graph classes or layout styles requires re-collecting and re-training procedures. We prove that our approach is more general, given that we are able to learn both drawing styles from graph drawing techniques and to draw by minimizing aesthetic losses.
Another very recent work,  DeepGD \cite{wang2021deepgd}, consists in a message-passing GNN which process starting positions produced by graph drawing frameworks~\cite{brandes2006eigensolver}, to construct pleasing layouts that minimize combinations of aesthetic losses (Stress loss combined with others).
Both DeepDraw and DeepGD share the common need of transforming the graph topology into a more complicated one: DeepDrawing~\cite{wang2019deepdrawing} introduces skip connections (\textit{fake edges}) among nodes in order to process the graph via a bidirectional-LSTM; DeepGD converts the input graph to a complete one, so that each node couples is directly connected, and requires to explicitly provide the shortest path between each node couple as an edge feature. The introduction of additional edges into the learning problem increase the computational complexity of the problem, hindering the model ability to scale to bigger graphs. More precisely, in the DeepGD framework the computational complexity grows quadratically in the number of nodes $O(N^2)$. Conversely, we show that the GNN are capable of producing aesthetically pleasing layouts without inserting additional edges, by simply leveraging powerful positional-structural features. 
Additionally, we introduce a novel neural-based mechanism, the \textit{Neural Aesthete}, capable to express differentiable aesthetic losses delivering flexible gradient direction also for non-differentiable goals. We show that this mechanism can be exploited by Gradient-descent based graph drawing techniques and by the proposed GND framework.
}
Finally, GNN-based Encoder-Decoder architectures can learn a generative model 
of the underlying distribution of data from a collection of graph layout examples~\cite{kwon2019deep}.

\section{Loss functions and Neural Aesthetes}
\label{sec:proposal_1}
\subsection{ \gc{Graph Drawing Algorithms}}
Graph drawing algorithms typically optimize functions that somehow express a sort of beauty index, leveraging information visualization techniques, graph theory concepts, topology and geometry to derive a graphical visualization of the graph at hand in a bidimensional or tridimensional space~\cite{battista1998graph, di1994algorithms, yoghourdjian2018exploring}.
Amongst others, typical beauty indexes are those of measuring the degree of edge crossings~\cite{purchase1997aesthetic}, the measurements to avoid small angles between adjacent or crossing edges, and  measurements to express a degree of uniform allocation of the vertexes~\cite{gibson2013survey, ahmed2020graph}.
All these requirements inherently assume that the graph drawing only consists of the allocation of the vertexes in the layout space, since the adjacent matrix of the graph can drive the drawing of the arcs as segments. However, 
we can also choose to link pairs of vertexes through a spline by involving some associated variables in the optimization process ~\cite{vismara2000experimental}.

Without loss of generality, in this work we restrict our objective to the vertex coordinates optimization, but the basis ideas can be extended also to the case of appropriate arc drawing.

As usual, we denote a graph by $G=(\mathcal{V},\mathcal{E})$, where  $\mathcal{V}= \{v_1, \dots, v_N\}$ is a finite set of N {\em nodes} and $E \subseteq V \times V$ collects the {\em arcs} connecting them. The neighborhood of node $v_i$ is denoted by $\mathcal{N}_i$.  
We  denote the coordinates of each vertex  with $p_i : \mathcal{V} \mapsto {\rm I\!R^2} $, for a node $i$ mapped to a bi-dimensional space. We denote with $P \in {\rm I\!R^{N \times 2}}$  the  matrix of the node coordinates.


One of the techniques that empirically proved to be  very effective for an aesthetically pleasing  node coordinates selection is the \textit{Stress} function~\cite{kamada1989algorithm},
\begin{equation}
    \textsc{stress}(P) = \sum_{i < j} w_{ij}\big(||p_i -p_j|| - d_{ij}\big)^2
    \label{eq:stress}
\end{equation}

where \mt{$p_i, p_j$}  are the coordinates of vertices $i$ and $j$, respectively, $d_{ij}$ is the graph theoretic distance (or shortest-path) between node $i$ and $j$, and $w_{ij}$ is a weighting factor leveraged to balance the influence of certain pairs given their theoretical distance. Usually, it is defined as $w_{ij} = d_{ij}^{-\alpha}$ with $\alpha \in [0, 1,2]$.
The optimization of this function is generally carried out leveraging complicated resolution methods (i.e., 2D Newton-Raphson method, Stress Majorization, etc.) that hinder its efficiency.

Recently, Gradient Descent methods were employed to produce graph layouts~\cite{zheng2018graph} by minimizing the Stress function, and noticeably, Ahmed et al.~\cite{ahmed2020graph} proposed a similar approach employing auto-differentiation tools. The advantage of this solution is that, as long as  aesthetic criteria are characterized by smooth differentiable functions, it is possible to undergo an iterative optimization process\footnote{The variables of the optimization process  are the node coordinates~$P$.} following, at each variable update step,  the gradient of the criteria.



Clearly, the definition of aesthetic criteria as smooth functions could be hard to express. 
For instance, while we can easily count the number of arc intersections, \gc{devising a smooth function that may drive a continuous optimization of this problem is not trivial~\cite{shabbeer2010optimal, ahmed2020graph}. 
Indeed, finding the intersection of two lines,\footnote{\gc{Obiously, in the case of arcs one should also check whether the intersection point, if exists, lies on one of the two segments.}} is as simple as solving the following equation system:} 
\begin{equation}
    \begin{cases}
    \gc{
        {a_1}x + {b_1}y + {c_1} = 0,} \\  
    \gc{ {a_2}x + {b_2}y + {c_2} = 0
        }
    \end{cases}
    \label{eq:inters}
\end{equation}
\gc{By employing the classic Cramer's rule we can see that there is an intersection only in case of a non-negative determinant of the coefficient matrix  $A, Det(A) = a_1b_2 - a_2b_1 \neq 0$. Clearly, the previous formula cannot be employed as a loss function in an optimization problem since it does not provide gradients.
To tackle this issue and provide a scoring function optimizable via gradient descent, we propose the \textit{Neural Aesthete}.
}
\subsection{\gc{The Neural Aesthete}}
A major contribution of this paper is that of introducing the notion of \textit{Neural Aesthete}, which is in fact a neural network that learns beauty from examples with the perspective of generalizing to unseen data. The obtained modelled function that is expressed by the Neural Aesthete is smooth and differentiable by definition and offers a fundamental heuristic for graph drawing. As a proof-of-concept, we focus on edge crossing. 
In this case, we define the Neural Aesthete as a machine which  processes two arcs as inputs and returns the information on whether or not they intersect one each other.
Each arc is identified by the coordinates of the corresponding pair of vertices, $e_{u} = (p_i, p_j)$ for $e_{u} \in \mathcal{E}$.
Hence, the Neural Aesthete 
$\nu(\cdot,\cdot,\cdot): {\cal E}^{2} \times \bbR^{m} \to \bbR$ 
operates on the concatenation of two arcs, 
$e_{u}$ and  $e_{v}$ and returns
\begin{equation}
    y_{e_u, e_v} = \nu \big(\theta,e_{u}, e_{v})
\end{equation}

where  $\theta \in \bbR^{m}$ is the vector which represents
the weights of the neural network. 
The Neural Aesthete is learned by optimizing a cross-entropy loss function $L(y_{e_u, e_v}, \hat{y}_{e_u, e_v})$  over the arcs $(e_u, e_v) \in \mathcal{E}$, which is defined as: 
\begin{equation}
\begin{split}
L({y}_{e_u, e_v}, \hat{y}_{e_u, e_v}) = & - \big(\hat{y}_{e_u, e_v} \cdot log(y_{e_u, e_v}) \\
& \quad \ + (1 - \hat{y}_{e_u, e_v}) \cdot log(1 - y_{e_u, e_v})\big),
\end{split}
\end{equation}
where $\hat{y}_{e_u, e_v}$ is the target:
\begin{equation}
    \hat{y}_{e_u, e_v} =    \begin{cases}
    0, & \text{if } \quad  (e_{u}, e_{v})  \text{  do not intersect}  \\
    1, &  \text{otherwise} 
    \end{cases}
    \label{eq:target}
\end{equation}
\gc{ and the intersection of $e_u, e_v$ is automatically computed, e.g., by solving Eq. \ref{eq:inters}.
}

Notice that the learning process from a finite set of supervised examples yields weights that allows us to estimate the probability of intersection of any two arcs. 
Basically, the learned output of the neural network can be regarded as a degree of intersection between any arc couple.
Once learned, this characteristic of the Neural Aesthete comes in handy for the computation of the gradient of a loss function for Graph Drawing. In general, we want to move the extreme nodes defining the two arcs towards the direction of \textit{non-intersection}.

Hence, for the Graph Drawing task, the Neural Aesthete is able to process an unseen edge couple $(e_{u}, e_{v})$ randomly picked from the edge list $\mathcal{E}$, and to predict their degree of intersection $y_{e_u, e_v}$.
We define the loss function $L(\cdot,\cdot)$ on this edge pair as the cross-entropy with respect to the target \textit{no-intersection}, $\hat{y}_{e_u{//}e_v} = 0$

\begin{equation}
    H_{u, v} = L(y_{e_u, e_v}, \hat{y}_{e_u{//}e_v}) = - log(1 - y_{e_u, e_v})
    \label{eq:dir_no_target}
\end{equation}

This smooth and differential loss function foster the utilization of Gradient Descent methods to optimize the problem variables, i.e. the arc node coordinates $(e_u, e_v)$.

This same procedure can be replicated to all the graph edges, 

\begin{equation}
    H(P) = \sum_{(e_u, e_v) \in \mathcal{E}} L(y_{e_u, e_v}, \hat{y}_{e_u// e_v})
    \label{eq:total_loss_neu}
\end{equation}

Overall, a possible graph drawing scheme is the one which returns
\begin{equation}
	 P^{\star} = \arg\min_{\cal P} H( P).
\end{equation}
This can be carried out by classic optimization methods. 
For instance, a viable solution is by gradient descent as follows:
\[
	 P \leftarrow  P - \eta \nabla_P H( P).
\]
where $\eta$ specifies the learning rate.

It is worth mentioning that, overall, this approach  leverages the computational efficiency and parallelization capabilities of neural networks. Hence, the prediction of the edge-crossing degree can be carried out for many edge couples in parallel.

Moreover, this same approach can be conveniently combined with other aesthetic criteria, for instance coming from other Neural Aesthetes or from classical loss function (e.g., Stress).
For example, we could consider 
\begin{align}
	E = H(P) +\lambda_{A} A(\cdot)+ \lambda_{B} B(\cdot)
\end{align}
where $A(\cdot)$ and $B(\cdot)$ denotes other aesthetic criteria characterized by smooth differentiable functions.

\subsection{Example: Neural Aesthete on small-sized Random Graphs}
\label{sec:prop1_ex}
\begin{figure}

    $\qquad $\textsc{start} $\qquad$  \textsc{stress} $\qquad$ \textsc{na-cross} $\qquad$  \textsc{combined} $\quad $ \\
    \centering
    \includegraphics[width=0.22\columnwidth, trim=40 30 20 20 , clip]{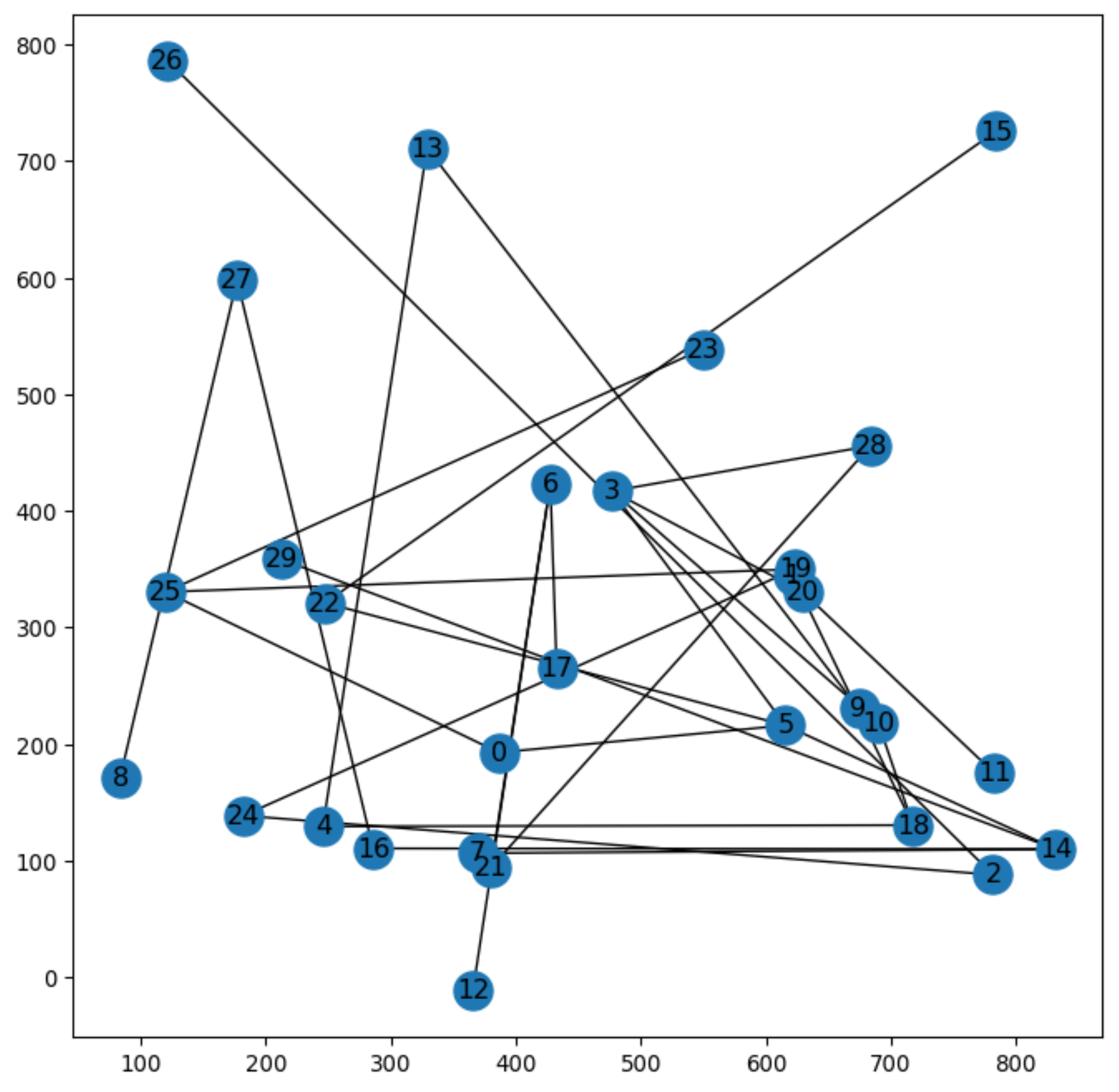} $\ $
    \includegraphics[width=0.22\columnwidth, trim=20 20 60 30 , clip]{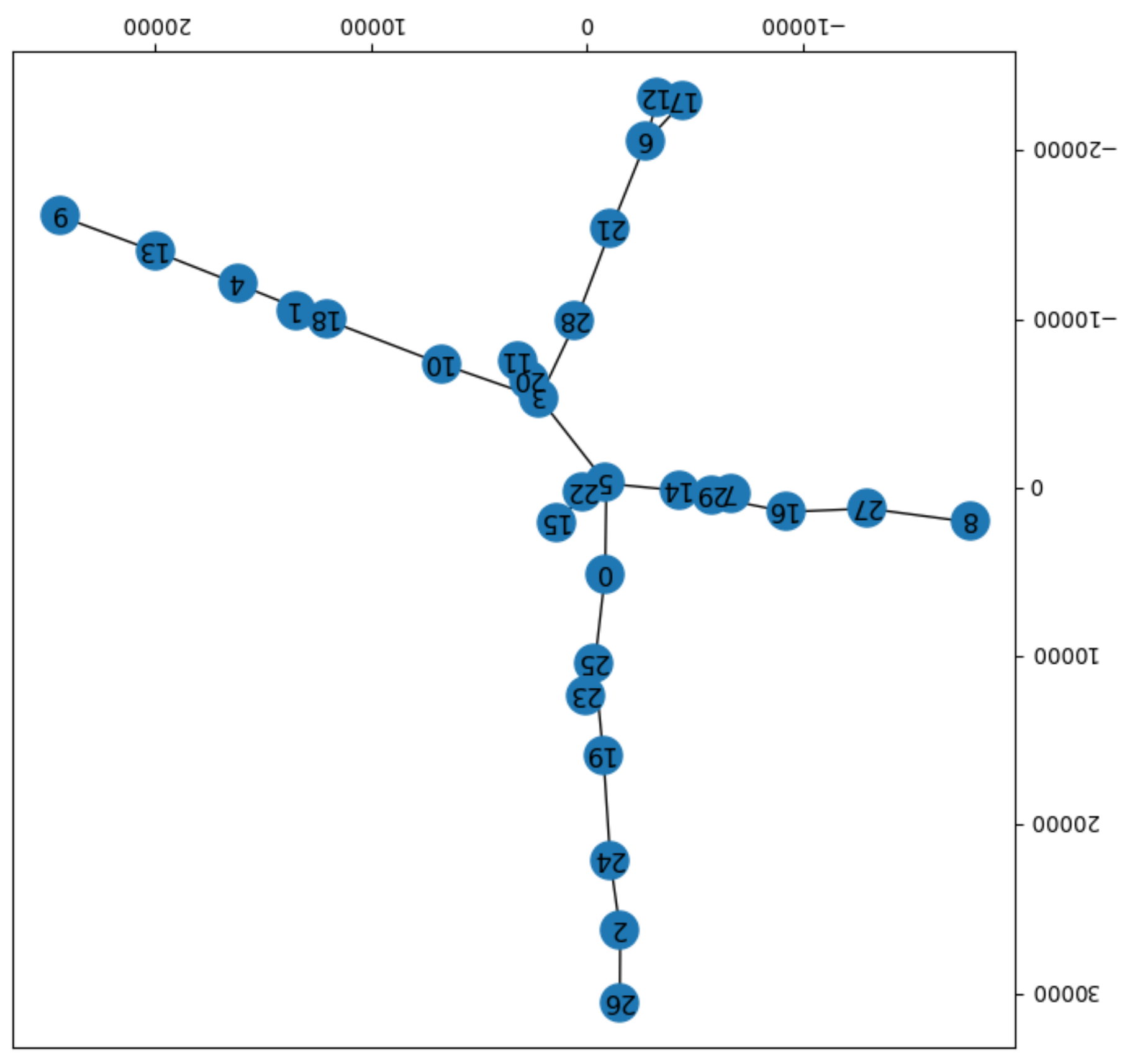}   $\ $
    \includegraphics[width=0.22\columnwidth, trim=60 30 20 20 , clip]{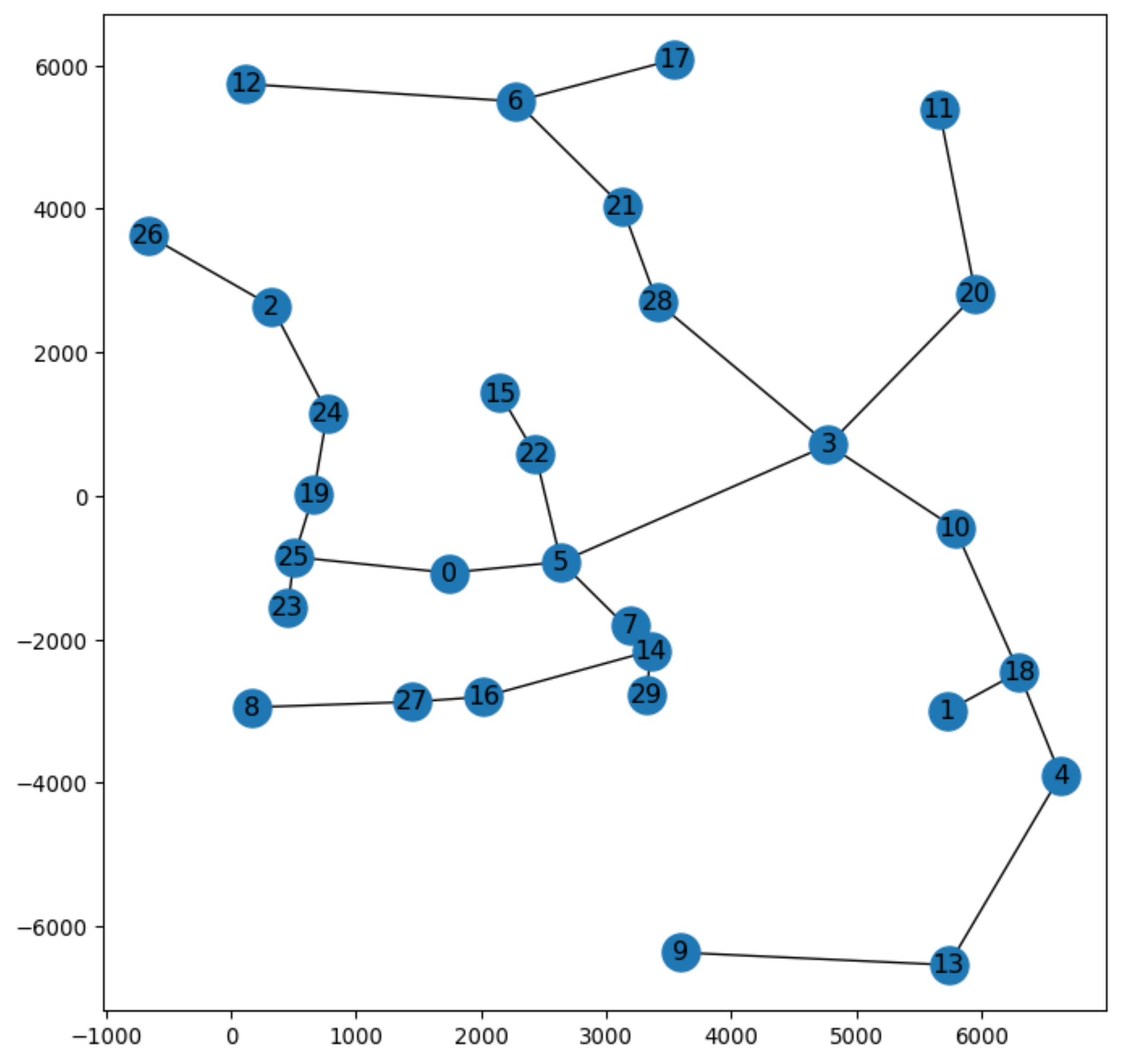} $\ $
    \includegraphics[width=0.22\columnwidth, trim=60 30 20 20 , clip]{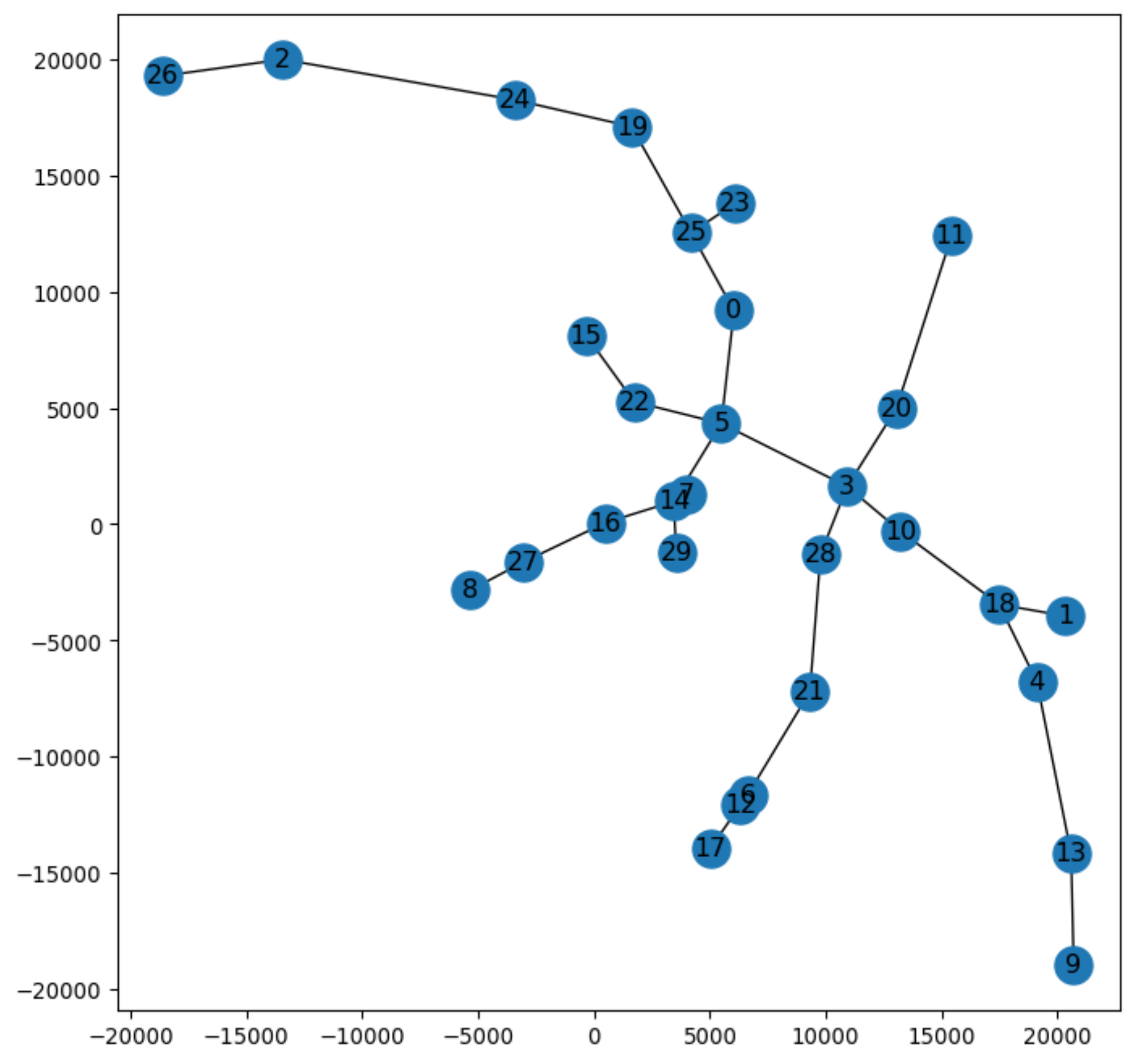} \\ \vspace{10 pt}
    \includegraphics[width=0.22\columnwidth, trim=40 30 20 20 , clip]{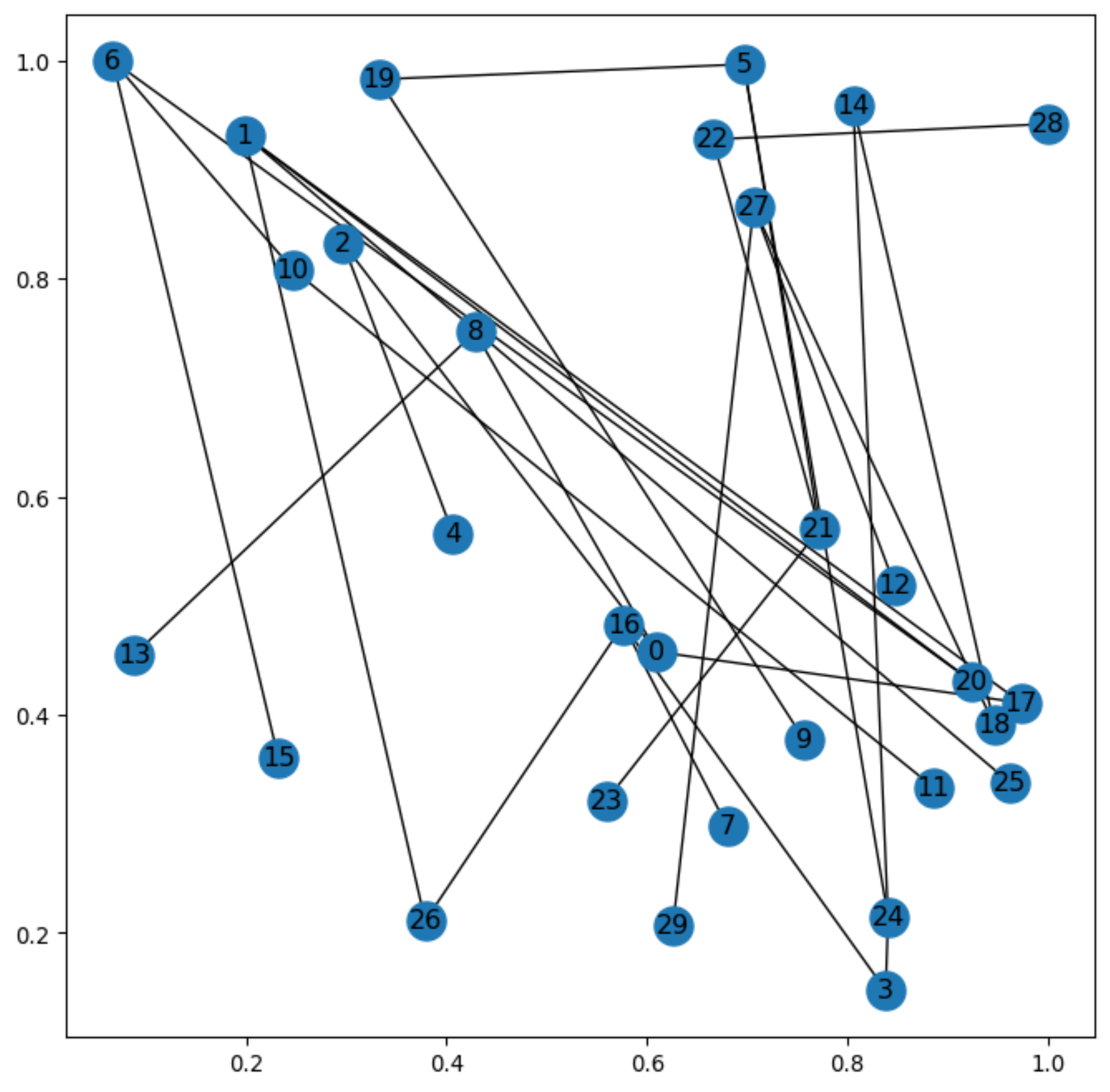} $\ $
    \includegraphics[width=0.22\columnwidth, trim=50 30 20 20 , clip]{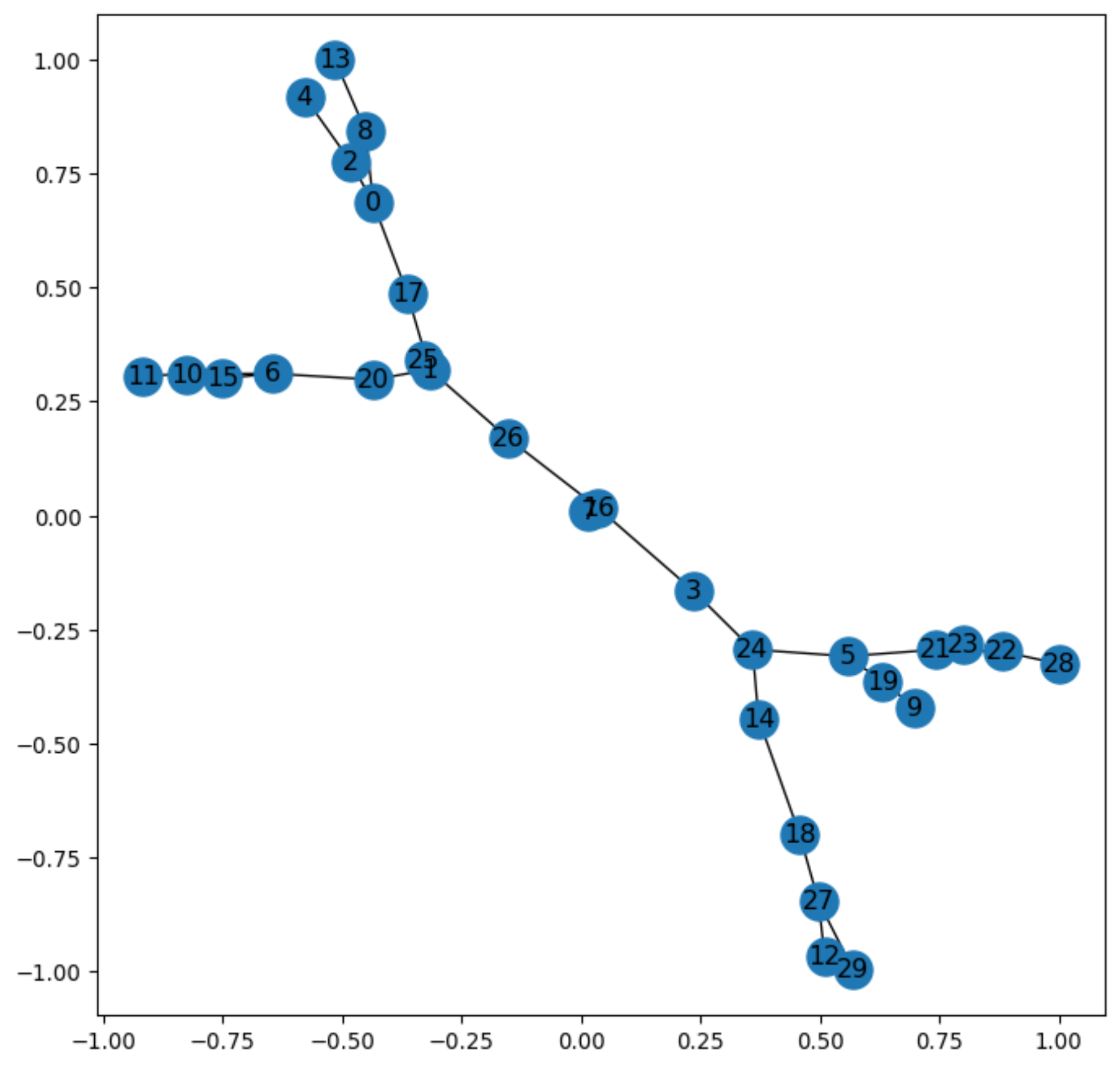}   $\ $
    \includegraphics[width=0.22\columnwidth, trim=20 30 20 20 , clip]{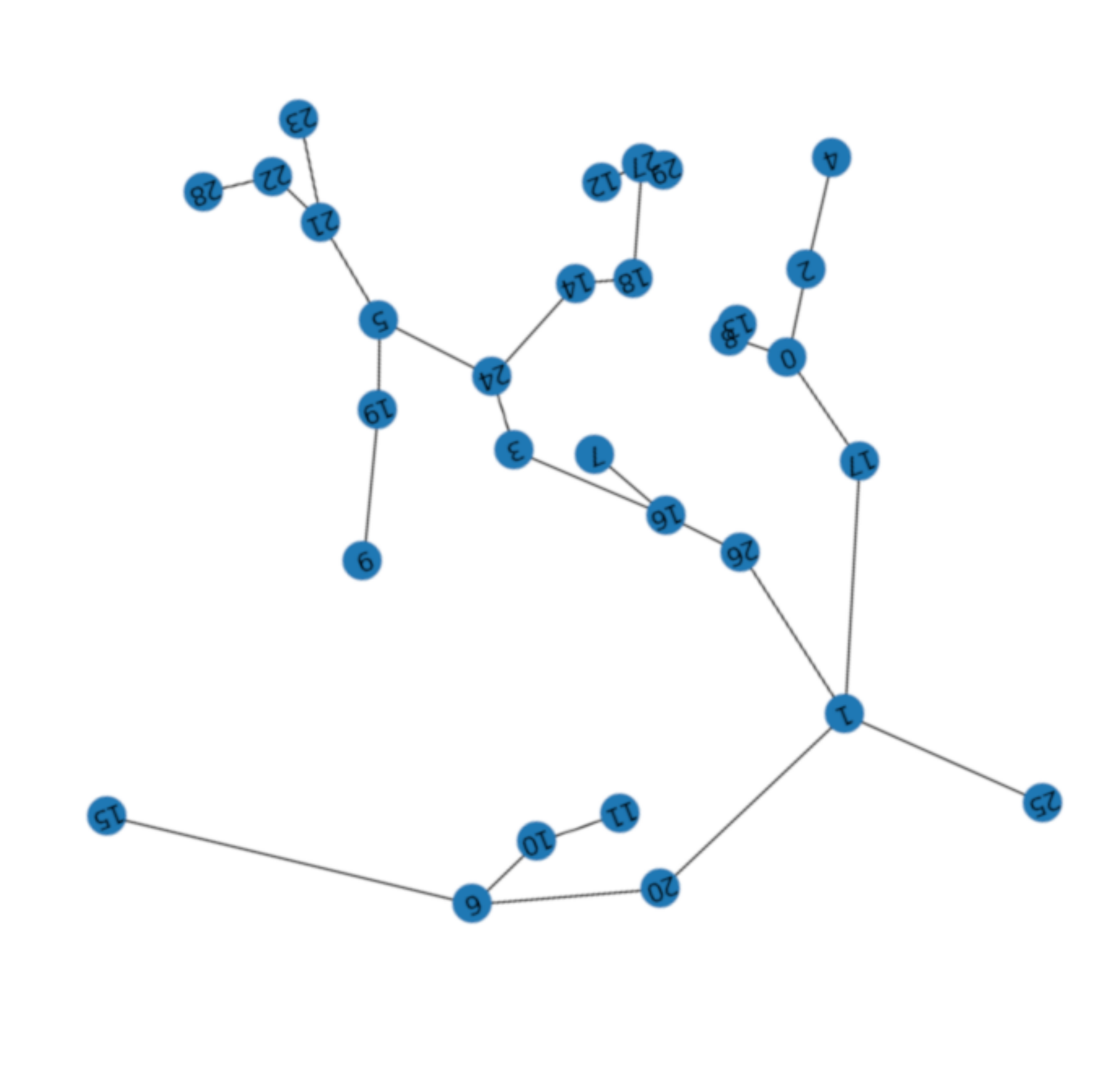} $\ $
    \includegraphics[width=0.22\columnwidth, trim=50 20 10 30 , clip]{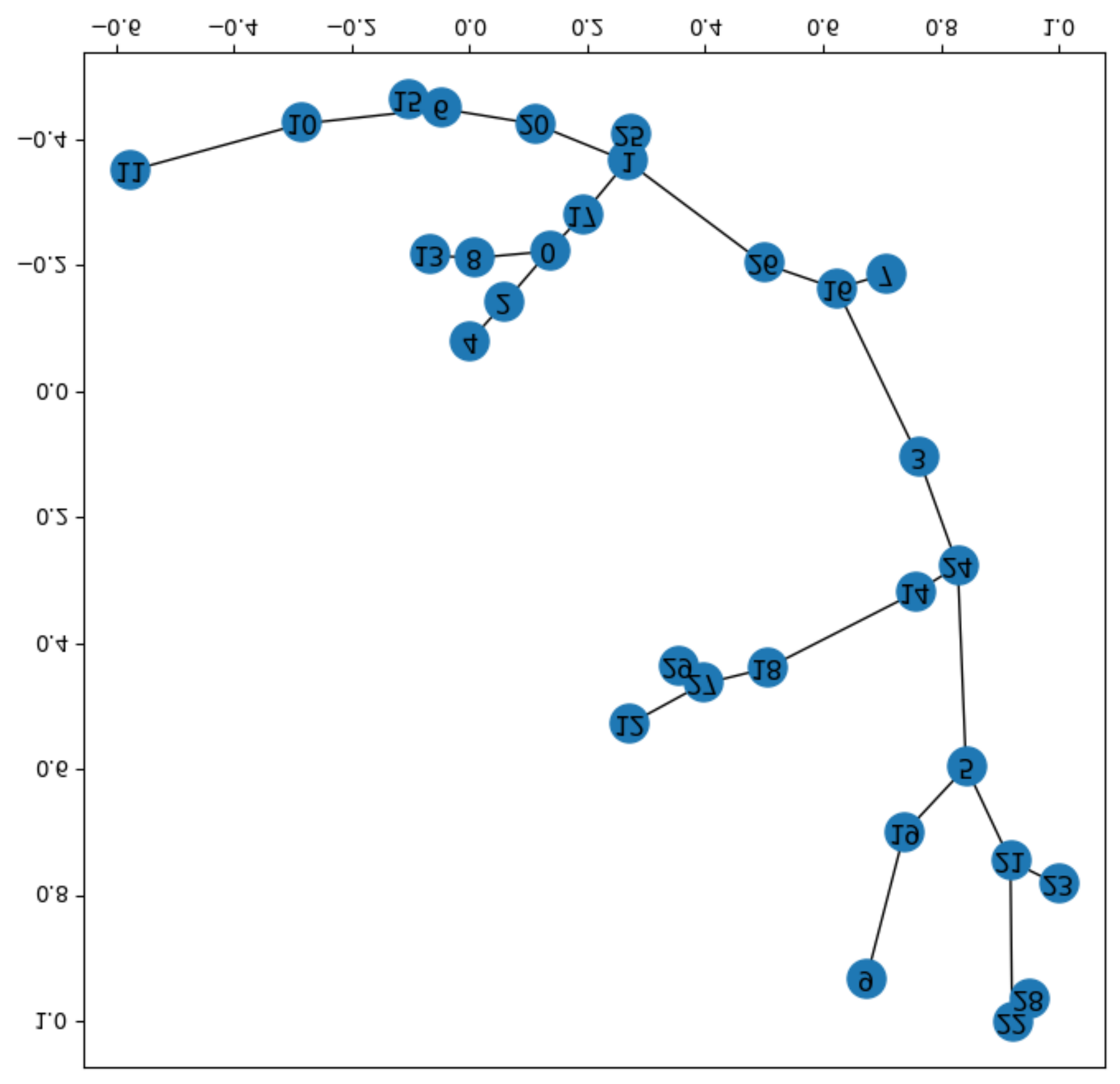} \\ \vspace{10 pt}
    \includegraphics[width=0.22\columnwidth, trim=40 30 20 20 , clip]{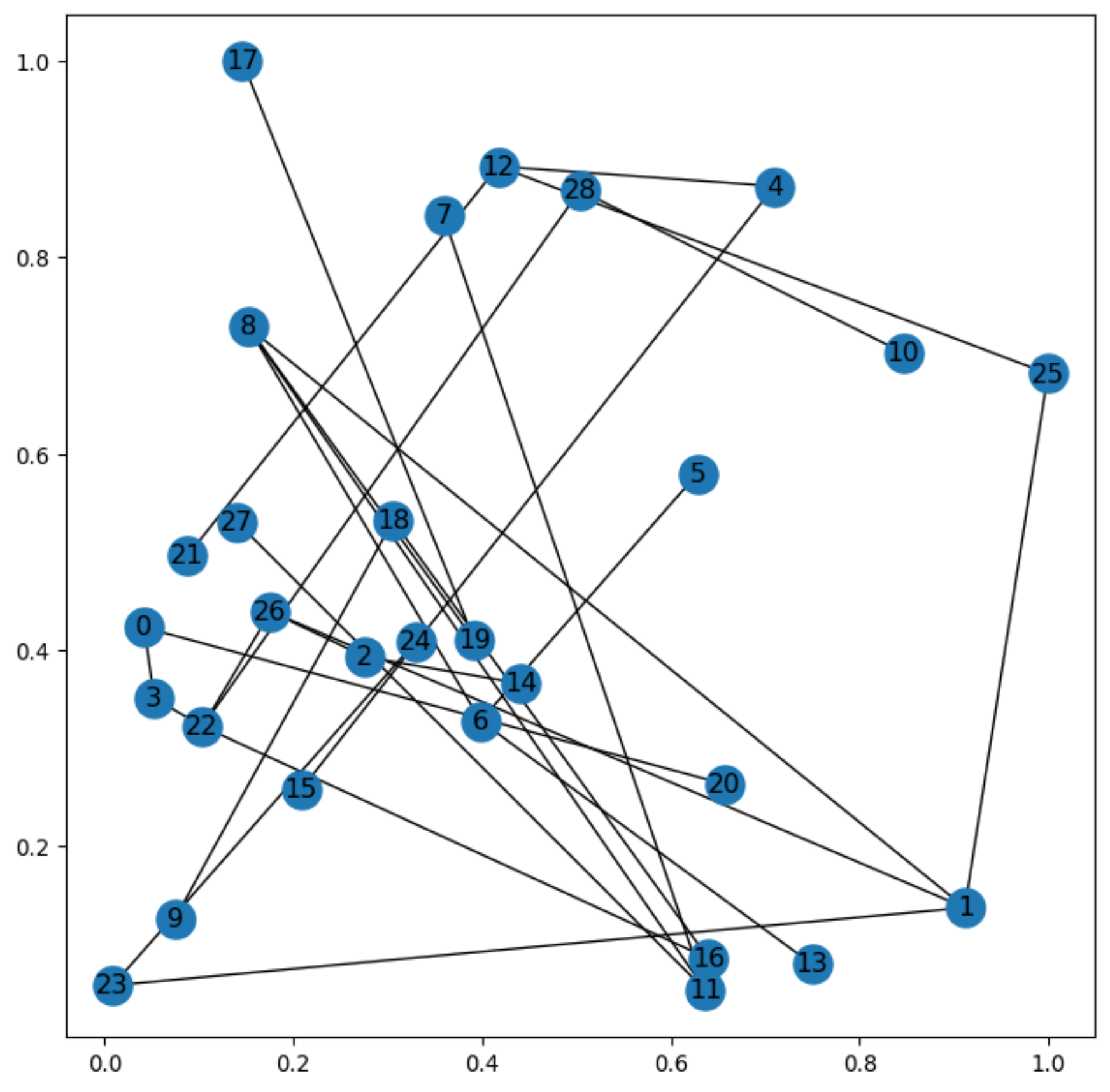} $\ $
    \includegraphics[width=0.22\columnwidth, trim=20 30 50 20 , clip]{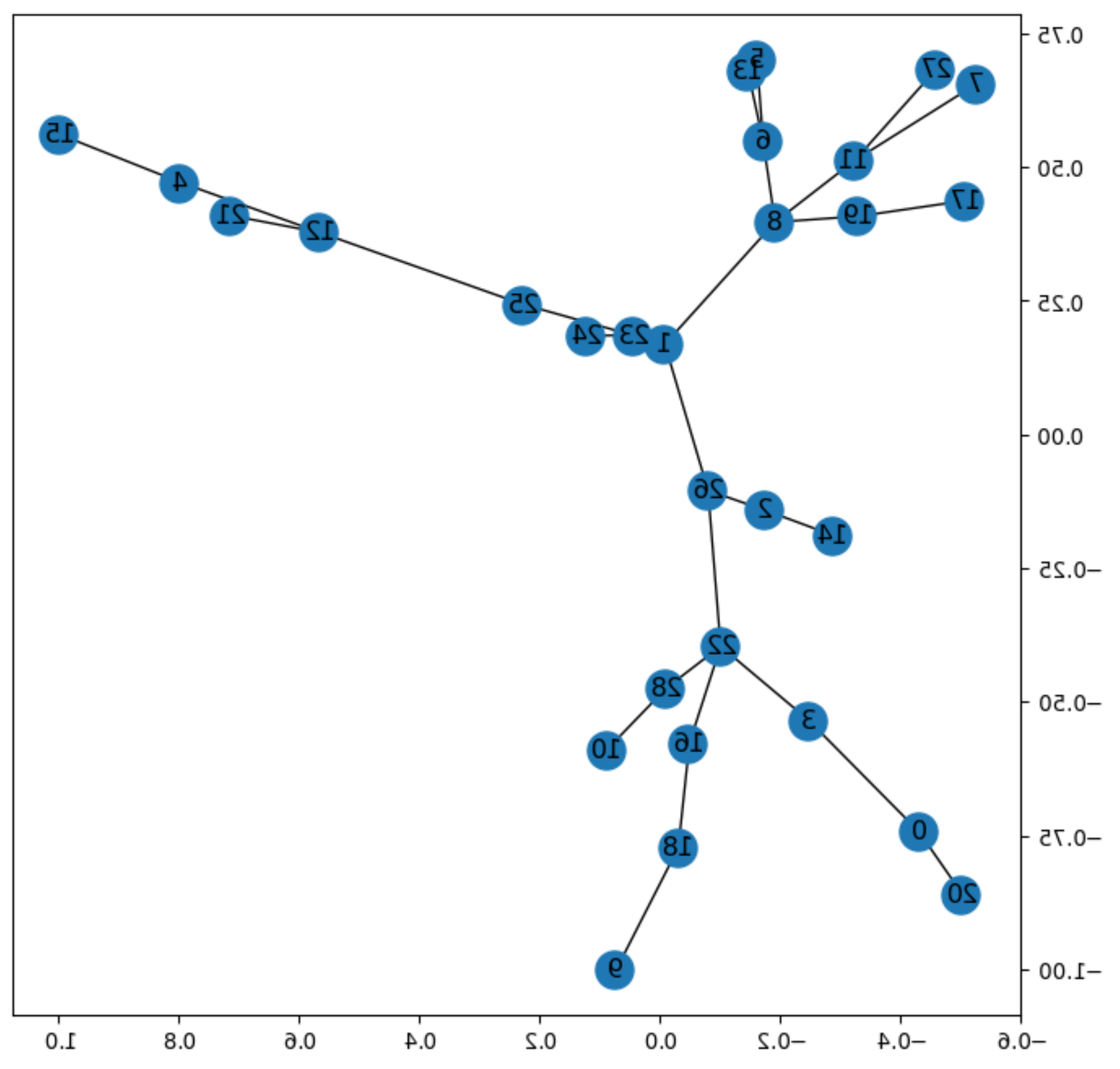}   $\ $
    \includegraphics[width=0.22\columnwidth, trim=50 30 20 20 , clip]{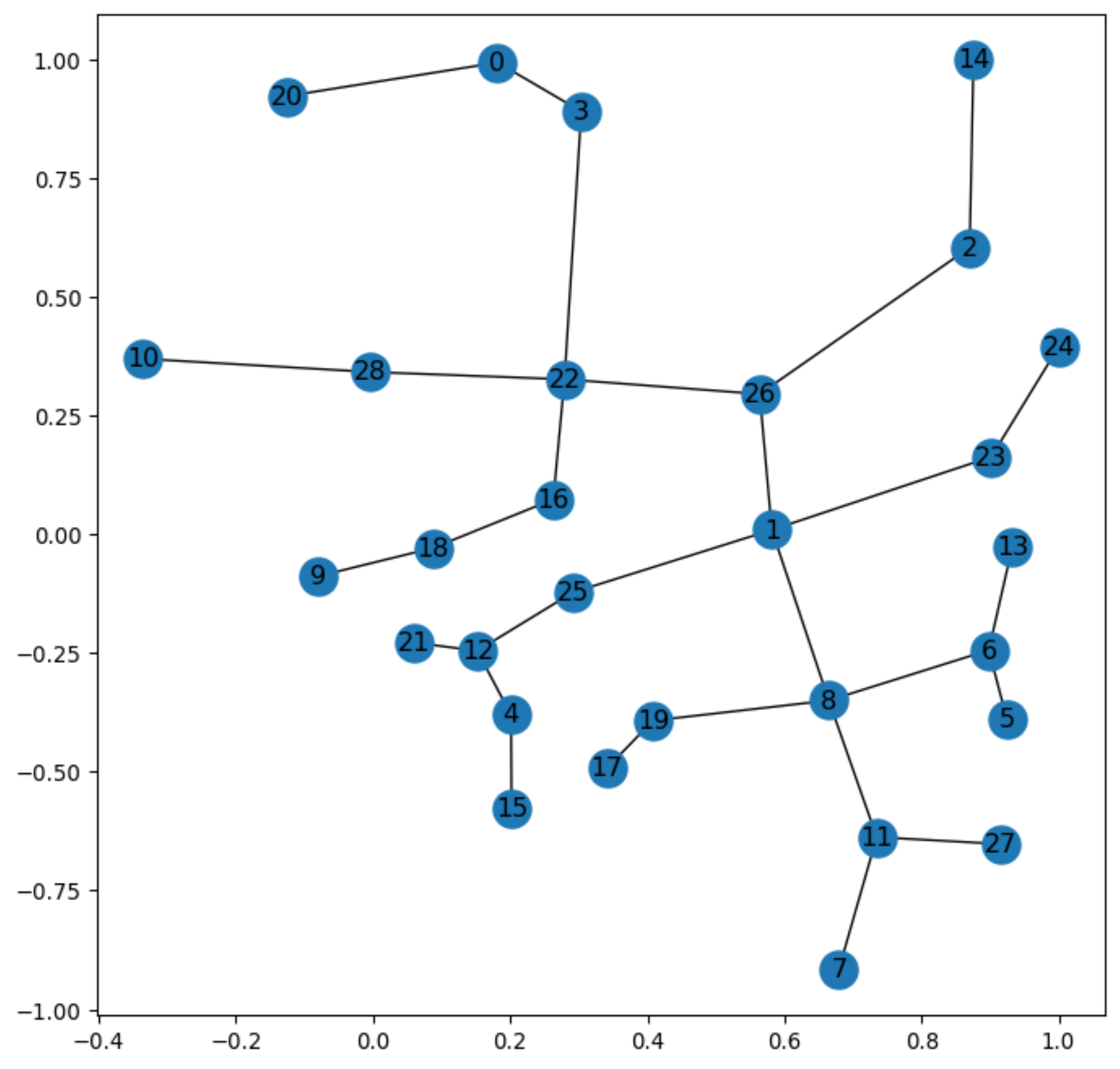} $\ $
    \includegraphics[width=0.22\columnwidth, trim=50 30 15 10 , clip]{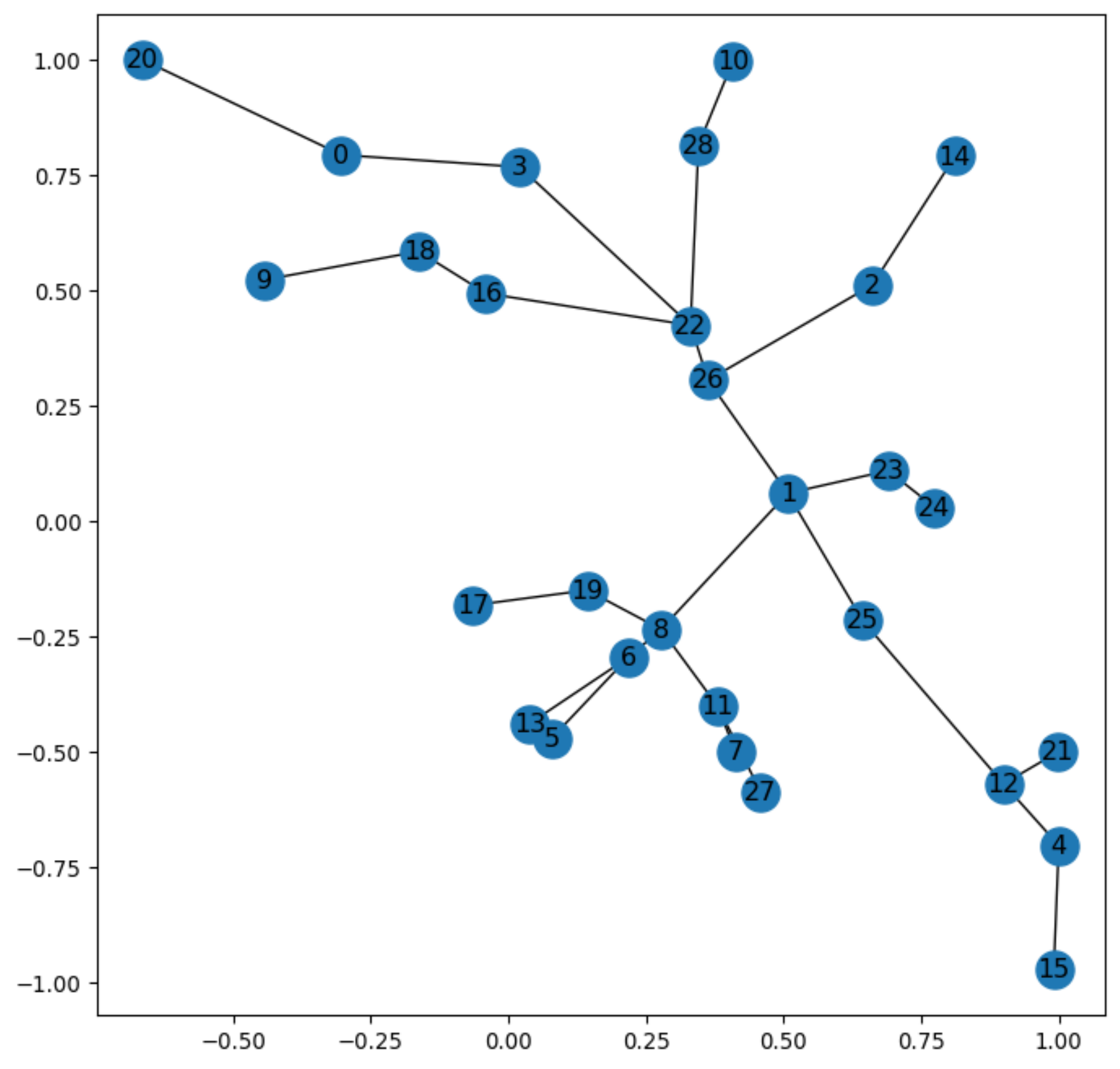} \\ \vspace{10 pt}
        \caption{\textsc{Neural Aesthete for edge crossing}. Left-to right, Graph layouts with starting random node coordinates (\textsc{start}), optimized by minimizing stress function with Gradient Descent (\textsc{stress}), optimized by Gradient Descent applied on the Neural Aesthete for edge-crossing loss (\textsc{NA-cross}), optimized by alternating stress loss and Neural Aesthete loss in subsequent iterations (\textsc{combined}). We report the graph layouts generated in three random sparse graphs, one for each row.}
    \label{fig:edge_cross}
\end{figure}

We provide a qualitative proof-of-concept example for the aforementioned Neural Aesthete for edge-crossing in Figure~\ref{fig:edge_cross}.

We built an artificial dataset composed of 100K entries to train the Neural Aesthete. Each entry of the dataset is formed by an input-target couple $(x, \hat{y})$.
The input pattern $x$ corresponds to the Neural Aesthete arcs input positions\footnote{Which are defined as  $x := [e_u, e_v] = [p_i, p_j, p_h, p_k]$, given two arcs $e_u = (p_i, p_j)$ and $e_v = (p_h, p_k)$.} as defined in Section \ref{sec:proposal_1}, whose node coordinates are randomly picked inside the interval $[0,1]$.
The corresponding target  $\hat{y}$ is defined as in Eq. \ref{eq:target}.

We balanced the dataset composition in order to have a comparable number of samples between the two classes (cross/no-cross).
We trained a Neural Aesthete implemented as a Multi-Layer Perceptron (MLP) with two hidden layers of 100 nodes each and ReLu activation functions, minimizing the cross-entropy loss function with respect to the targets and leveraging the Adam optimizer \cite{kingma2014adam}.
We tested the generalization capabilities of the learned model on a test dataset composed of 50K entries, achieving a test accuracy of $97\%$\footnote{\mt{For comparison, a Decision Tree model, trained on the same dataset, only reaches a test accuracy of 78.7\%.}}. 
Hence, the learned model constitutes the Neural Aesthete for the task of edge crossing.
Given an unseen input composed of a couple of arcs, the learned model outputs a probability distribution representing a degree of intersection. Following the common pipeline of Graph Drawing methods with Gradient Descent, the Neural Aesthete output represent a differentiable function that provides an admissible descent direction for the problem parameters $P$.

To test the capability of the proposed solution, we leverage an artificial dataset of random graphs with a limited number of nodes $(N \in [20, 40]$).
We generated Erdős-Rényi graphs with the method presented in \cite{batagelj2005efficient} for efficiently creating sparse \footnote{The probability of edge creation has been set to $p=0.01$.} random networks, implemented in NetworkX \cite{hagberg2008exploring}. We selected the connected component of the generated graph having the biggest size (max node number).

Figure \ref{fig:edge_cross} reports a qualitative example of the proposed method in three graphs from the aforementioned dataset. To generate the graph layout, we carried on an optimization process on mini-batches of 10 arc-couples for an amount of 2K iterations (gradient steps). The first column depicts the starting random positions of the nodes; second column reports the graph layout obtained with an in-house implementation of the Stress function (see Eq. \ref{eq:stress}), optimized via Gradient Descent as done in \cite{ahmed2020graph}; third column contains the results obtained by optimizing the loss provided by the proposed Neural Aesthete for edge-crossing; fourth column reports the layouts obtained alternating the optimization of the Stress function and edge-crossing in subsequent update steps. 
It is noticeable to see how the solution provided by our approach is capable to avoid any arc intersection in these simple graphs. Moreover, the fact that the Neural Aesthete output represents a form of degree-of-intersection, seems to provide a good gradient direction that easily moves the arcs into a recognizable angle pattern, even when combined with other criteria (fourth column).
The proposed proof-of-concept proves that Neural Aesthetes represent a feasible, general and efficient solution for Graph Drawing. In the following, we prove that this same approach can be used to guide the training process of different kind of Deep Neural Models.

\section{GNNs for Graph Drawing}
\label{sec:proposal_2}

The increasing adoption of \mt{GNNs} in several research fields and practical tasks \cite{jumper2021highly, zhao2019t} opens the road to an even wider spectrum of problems.  Clearly, Graph Drawing and GNNs seem inherently linked, even if the formalization of this learning process under the GNN framework is not trivial. 
As pointed out in Section \ref{sec:related}, some recent works leveraged  GNNs-inspired models for Graph Drawing. DeepDrawing~\cite{ wang2019deepdrawing} employs a graph-based LSTM model to learn node layout styles from Graph Drawing frameworks. 
DeepGD \cite{wang2021deepgd} is a concurrent work in which an MPNN processes starting node positions to develop pleasing layouts that minimize combinations of aesthetic losses (Stress loss combined with others). Starting node positions, however, needs to be initialized by standard graph drawing frameworks~\cite{brandes2006eigensolver}; in case a random initialization is employed, network performances deteriorate.

One of the drawbacks of both these approaches is the fact that they modify the graph topology, introducing additional connections that were not present in the original graph. This fact entails an increased computational burden for the model. Indeed, a complete graph requires many more message exchanges than a sparse one, being the computational complexity of the GNN propagation \gc{linear in the number of edges} \cite{DBLP:journals/tnn/WuPCLZY21}. Moreover, the complete graph processed by DeepGD is enriched with edge features being the shortest path between the node connected to the edge. This solution gives big advantages in tasks closely connected with the Stress minimization but could prevent the network from generalizing to other tasks. 

We propose an approach to Graph Drawing, GNDs, that leverages the computational efficiency of GNNs and, thanks to informative nodal features (Laplacian eigenvectors, see Sec. \ref{sec:pe}), is general enough to be applied to  several learning tasks. 

\subsection{Graph Neural Networks}

First and foremost, let us introduce some notation. We denote with $l_i$ all the input information (initial set of features) eventually attached to each node $i$ in a graph $G$. The same holds for an arc connecting two nodes $i$ and $j$, whose feature, if available, is denoted with $l_{(i,j)}$.   
Each node $i$ has an associated hidden representation (or state) $x_i\in \R^s$, which in recent models is initialized with the initial features, $x_i = l_i$ (but it is not necessarily the case in  RecGNN models \cite{tiezzi2021deep}). 
Many GNN models can be efficiently described under the powerful umbrella of Message Passing Neural Networks (MPNNs) \cite{gilmer2017neural}, where the node state $x_i$ is iteratively updated at each iteration $t$, through an aggregation of the exchanged information among neighboring nodes $\mathcal{N}_i$, undergoing a message passing process. Formally,
\begin{eqnarray}
    \label{m1} x_{(i,j)}^{(t-1)} &=& \texttt{MSG}^{(t)}\left(x_i^{(t-1)}, x_j^{(t-1)}, l_{(i,j)}\right) \\
    \label{m2} x_i^{(t)} &=& \texttt{AGG}^{(t)}\left(x_i^{(t-1)}, \sum_{j \in \mathcal{N}_i}  x_{(i,j)}^{(t-1)},  l_i\right) 
\end{eqnarray} 

where  $x_{(i,j)}^{(t)}$ represent explicitly the message exchanged by two nodes, computed by a learnable map $\texttt{MSG}^t(\cdot)$.\footnote{Notice that in the case in which arc features $l_{(i,j)}$ are not available they are removed from the problem formulation.}  
Afterwards,  $\texttt{AGG}^t(\cdot)$ aggregates the incoming messages from the neighborhood, eventually processing also local node information such as the node hidden state $x_i$ and its features $l_i$. 
The messaging and aggregation functions $\texttt{MSG}^{(t)}(\cdot)$,  $\texttt{AGG}^{(t)}(\cdot)$ are typically implemented via Multi-Layer Perceptrons (MLPs) learned from data. Apart from RecGNN, other GNN models leverage a different set of learnable parameters for each iteration step. Hence, the propagation process of such models can be described as the outcome of a multi-layer model, in which, for example, the node hidden representation at layer $t$,  $x_i^{(t)}$, is provided as input to the next layer, $t+1$. Therefore an $\ell$-step message passing scheme can be seen as an $\ell$-layered model.

This convenient framework is capable to describe several GNN models \cite{DBLP:journals/tnn/WuPCLZY21}. In this work, we focus our analysis on three commonly used GNN model from literature (i.e., GCN \cite{kipf2017semisupervised}, GAT \cite{velivckovic2018graph}, GIN \cite{DBLP:journals/corr/abs-1810-00826}) whose implementation is given in Table~\ref{tab:gnn}, characterized by different kinds of aggregation mechanisms (degree-norm, attention-based, injective/sum, respectively).   
Following the Table notation, in GCN $c_{u,v}$ denotes a normalization constant depending on node degrees; in GAT $\alpha^{(t-1)}_{u,v}$ is a learned attention coefficient which introduces anisotropy in the neighbor aggregation, $\sigma$ denotes a non-linearity and $W, W_0, W_1$ are learnable weight matrices; in GIN $\epsilon$ is a learnable parameter (which is usually set to zero).

\begin{table}
    \caption{Common implementations of GNN aggregation mechanisms. 
   { 
   See the main text and the referenced papers for further details on the formulations.}
   }
    \centering
     
    \begin{tabular}{lHc}
        \toprule
        Method: Funct.& Reference & Implementation \\
         \midrule
        
        GCN\cite{kipf2017semisupervised}: Mean & Kipf and Welling \cite{DBLP:conf/iclr/KipfW17} & \mt{$\sigma \bigg( c_{v}W^{(t)}  x_v{^{(t-1)}} + \sum_{u \in \mathcal{N}_v} c_{u,v} W^{(t)} x_u{^{(t-1)}}\bigg)$} \\

         
         GAT\cite{velivckovic2018graph}: Att. & {Veličković et al. \cite{velivckovic2018graph}  } & { $\sigma \big(\sum_{u \in \mathcal{N}_v} \alpha^{(t-1)}_{u,v} W^{(t)}x_u^{(t-1)}\big)$}\\

          GIN\cite{DBLP:journals/corr/abs-1810-00826}: Sum  & Xu et al. \cite{DBLP:journals/corr/abs-1810-00826} & $\text{MLP}{^{(t)}} \big((1 + \epsilon)  x_v{^{(t-1)}} + \sum_{u \in \mathcal{N}_v} x_u{^{(t-1)}}$ \big) \\
         
         
         \bottomrule\\
    \end{tabular}
   
    \label{tab:gnn}
\end{table}

\subsection{Problem Formulation}
\label{sec:pe}


Through the \mt{GNDs} framework we propose to employ the  representational and generalization capability of GNNs to learn to generate graph layouts.  
We formulate the problem as a \textit{node-focused} regression task, in which for each vertex belonging to the input graph we want to infer its coordinates in a bi-dimensional plane, conditioned on the graph topology and the target layout/loss function (see Section \ref{sec:dataset_crea}). 
Furthermore, in the GND framework, we propose to employ GNNs to learn to draw by themselves following the guidelines prescribed by Neural Aesthetes. (see Section \ref{sec:neu}). 
In order to be able to properly solve the Graph Drawing task via GNDs, a crucial role is played by the expressive power of the GNN model and the nodal features which are used. In fact, in line with the aforementioned regression task, each node state must be uniquely identified to be afterwards mapped to a different 2D position in the graph layout. This problem is inherently connected with recent studies on the representational capabilities of GNNs (see Section \ref{sec:related} and \cite{you2019position}). 
Standard MP-GNNs have been proved to be less powerful than the 1-WL test \cite{morris2019weisfeiler}, both due to the lack of expressive power of the used aggregation mechanisms and to the existence of symmetries inside the graph. For instance, local isomorphic neighborhoods create indiscernible unfolding of the GNN computational structure. Hence, the GNN embeds isomorphic nodes to the same point in the high dimensional space of the states, hindering the Graph Drawing task. 
Some approaches address this problem proposing novel and more powerful architectures (WL-GNNs) that, however, tend to penalise the computational efficiency of the GNNs \cite{morris2019weisfeiler}.  
Moreover, given the fact that we focus on the task of drawing non-attributed graphs, it is even more important to enrich the nodes with powerful features able to identify both the position of nodes inside the graph (often referred to as Positional Encodings (PEs)\cite{dwivedi2020benchmarking}) and able to describe the neighboring structure.  

\mt{
Recently, it has been shown that the usage of random nodal features theoretically strengthens the representational capability of GNNs \cite{sato2021random,abboud2021surprising}. Indeed, setting random initial node embedding (i.e.,  different random values when processing the same input graph) enable GNNs to better distinguish local substructures, to learn distributed randomized algorithms and to solve matching problems with nearly optimal approximation ratios. 
Formally, the node features can be considered as random variables sampled from a probability distribution $\mu$ with support $D \subseteq \R^s$, 
\begin{equation}
     l_i \sim \mu, \quad \forall i \in \mathcal{V}
\end{equation}
where $\mu$ can be instantiated as the Uniform distribution.
The main intuition is that the underlying message passing process combines such high-dimensional and discriminative nodal features, fostering the detection of fixed substructures inside the graph \cite{sato2021random}. 
These approaches, which hereinafter we refer to as rGNNs, proved that classification tasks can be tackled in a novel way, with a paradigm shift from the importance of task-relevant information (the features values) to the relevance of the relationship among node values. However, the peculiar regression task addressed in this work requires both positional and structural knowledge, which is essential to identify and distinguish  neighboring nodes.
}

To address this issue, we keep standard GNN architectures and leverage Positional features defined as the Laplacian eigenvectors \cite{belkin2003laplacian} of the input graph, as introduced recently in GNNs \cite{dwivedi2020benchmarking}.
Laplacian eigenvectors embed the graphs into the Euclidean space through a spectral technique, are unique and distance-preserving (far away nodes on the graph have large PE distance). Indeed, they can be considered  hybrid positional-structural encodings, as they both define a local coordinate system and preserve the global graph structure.

Formally, they are defined via the factorization of the graph Laplacian matrix:

\begin{align}
L =\textrm{I}-D^{-1/2}AD^{-1/2}=U^T\Lambda U,
 \label{LapEig}
\end{align}
where $I$ is the $N \times N$ identity matrix, $D$ is the node degree matrix,  $A$ is the   adjacency matrix and $\Lambda$ and $U$ correspond respectively to the eigenvalues and eigenvectors. As proposed in \cite{dwivedi2020benchmarking}, we use the $k$ smallest non-trivial eigenvectors to generate a $k$-dimensional feature vector for each node, where $k$ is chosen by grid-search. \mt{Noticeably, given that the smallest eigenvectors provide smooth encoding coordinates of neighboring nodes,  during the message exchange each node receives and implicit feedback on its own positional-structural characteristics from all the nodes with which it is communicating. This process foster the regression task on the node coordinates, which receives useful information from their respective neighborhood. We believe that this is a crucial component of the model pipeline. 
}



\subsection{Experimental setup}
\label{sec:dataset_crea}

We test the capabilities of the proposed framework 
comparing the performances of three commonly used GNN models (see Table \ref{tab:gnn}).
In the following, we describe the learning tasks and the datasets employed for testing the different models.
In Sections \ref{sec:super}, \ref{sec:stress} and \ref{sec:neu}, instead, we will give qualitative and quantitative evaluations for each learning problem, showing the generality of our approach. 

Given the fact that the outputs of GND are node coordinates, we can impose on such predictions heterogeneous loss functions that can be optimized via BackPropagation. In the proposed experiments, we test the GND performances on the loss functions defined as the following: 
\begin{enumerate}[label=(\roman*)]
    \item distance with respect to ground truth node coordinates belonging to certain layouts, produced by Graph Drawing packages (Section \ref{sec:super});
    \item aesthetic loss functions (e.g. Stress) (Section \ref{sec:stress});
    \item loss functions provided by Neural Aesthetes (Section \ref{sec:neu}).
\end{enumerate}
We assume to work with solely the graph topology, hence the node are not characterized by additional features.

We employed two different graph drawing datasets with different peculiarities. We chose to address small-size graphs ($\leq 100$ nodes) to assure the graph layout readability, since prior works highlighted node-link layouts are more suitable to small-size graphs \cite{wang2019deepdrawing, ghoniem2004comparison}.
The former one is the \textsc{Rome} dataset{\footnote{http://www.graphdrawing.org/data.html}, a Graph Drawing benchmarking dataset containing 11534 undirected graphs with heterogeneous structures and connection patterns. We preprocessed the dataset removing three disconnected graphs\footnote{\gc{Stress-based Graph drawing techniques cannot take into account disconnected graphs. However, one can easily draw each connected component separately and then plot them side by side.}}}. Each graph contains a number of nodes between 10 and 100.
Some samples of the dataset are reported in the first column of Figures \ref{fig:sup_exp_kamada} and \ref{fig:sup_exp_spectral}, drawn with different layouts (see the following).

\begin{figure}
    \centering
    \includegraphics[width=0.45\columnwidth]{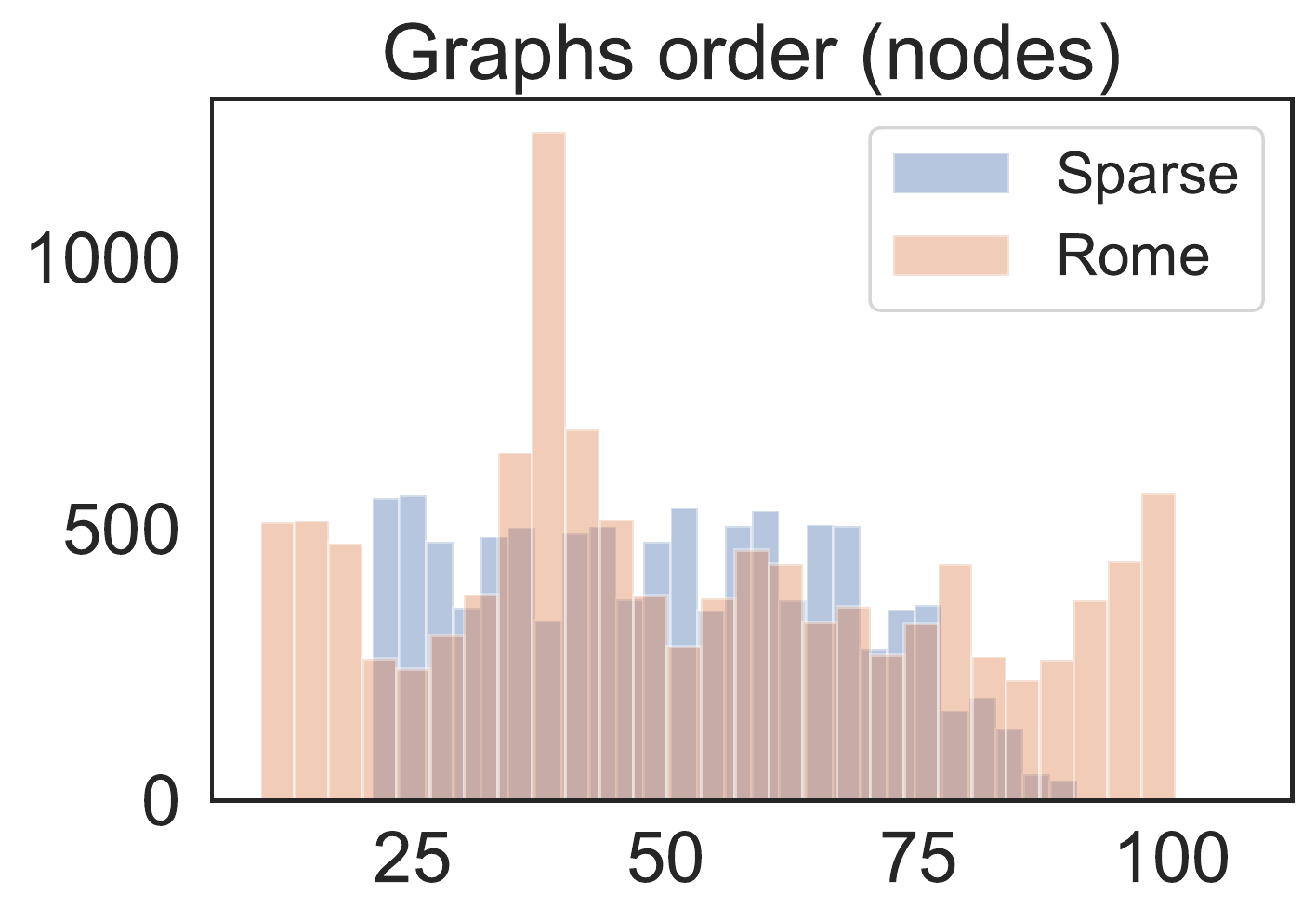}
    \includegraphics[width=0.45\columnwidth]{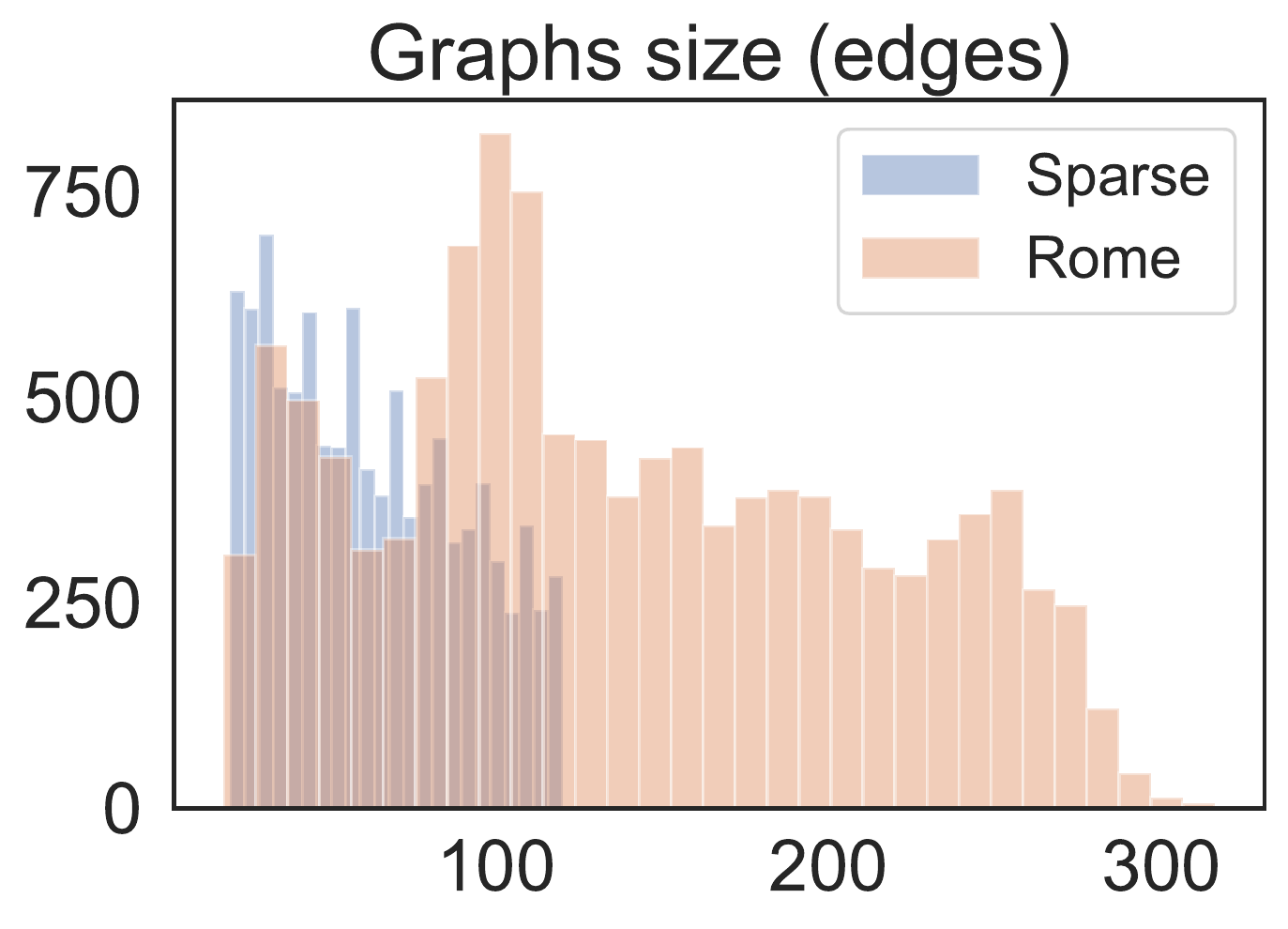}
    \caption{Datasets composition statistics. On the left, the histogram of the graph order (number of nodes for each graph $|\mathcal{V}|$) for both the analyzed datasets. On the right, the histogram of the graphs sizes (number of edges $|\mathcal{E}|$). The \textsc{Sparse} dataset is characterized by a sparse connection pattern.}
    \label{fig:stats_dataset}
\end{figure}

We built a second dataset, which we refer to as \textsc{Sparse}, with the same technique described in Section \ref{sec:prop1_ex}. We generated 10K Erdős-Rényi graphs following the method presented in \cite{batagelj2005efficient} for efficient sparse random networks and implemented in NetworkX \cite{hagberg2008exploring}. 
We randomly picked the probability of edge creation in the interval $(0.01, 0.05)$ and the number of nodes from 20 to 100. To improve the sparsity and readability, we discarded all the created graphs having both more than 60 nodes and more than 120 edges. Afterwards, we selected the connected component of the generated graph having the biggest size (max node number). 
We report in Figure \ref{fig:stats_dataset} a visual description of the datasets composition. 

In order to carry out the training process and afterwards evaluate the obtained performances, we split each of the datasets into three sets, (i.e. training, validation, test) with a ratio of (75\%, 10\%, 15\%).

\subsection{GNNs learn to draw from ground-truth examples}
\label{sec:super}

The first experimental goal is focused on the task of learning to draw graph layouts 
given ground-truths node positions produced by Graph Drawing frameworks. Among several packages, we chose NetworkX \cite{hagberg2008exploring} for its completeness and ease of integration with other development tools. This framework provides several utilities to plot graph appearances. We choose two different classical layouts. The first is the  \textsc{Kamada-Kawai} node layout \cite{kamada1989algorithm} computed by optimizing the Stress function. In few words, this force-directed method models the layout dynamic as springs between all pairs of vertices, with an ideal length equal to their graph-theoretic distance.
The latter is the \textsc{Spectral} layout, which leverages the unnormalized Laplacian $\hat{L}$ and its eigenvalues to build cartesian coordinates for the nodes \cite{beckman1994theory}, \mt{formally:
\begin{equation}
    \hat{L} = D - A = \hat{U}^T\hat{\Lambda} \hat{U},
    \label{eq:spectral}
\end{equation}
where $\hat{\Lambda}$ and $\hat{U}$ correspond to the eigenvalues and eigenvectors, respectively, and using the first two non-trivial eigenvectors ($k=2$) as the actual node coordinates.\footnote{ \mt{The NetworkX Spectral layout adds a rescaling of the node coordinates into the range $(-1,1)$ as a standard step.}}
}
\mt{ We remark that Eq. \ref{LapEig} and \ref{eq:spectral} produce different outputs.} 
This layout tends to highlight clusters of nodes in the graph.\footnote{\mt{See the referenced papers for further details on the layouts properties.}} 

Each training graph is enriched by Positional Encodings  defined as $k$-dimensional Laplacian Eigenvectors (see Section \ref{sec:pe}) and is processed by each of the tested GNN models to predict the node coordinates. Hence, we need a loss function capable to discern if the generated layout is similar to the corresponding ground truth. Furthermore, trained models should generalize the notion of graph layout beyond a simple one-to-one mapping. 
For these reasons, we leverage the Procrustes Statistic \cite{wang2019deepdrawing} as a loss function since it measures the shape difference among graph layouts independently of affine transformations such as translations, rotations and scaling. Given a graph composed of N nodes, the predicted node coordinates $P = (p_1, ..., p_N)$ and the ground-truth positions $\hat{P} = (\hat{p}_1, .., \hat{p}_N)$, the Procrustes Statistic similarity is defined as the squared sum of the distances between $P$ and $\hat{P}$ after a series of possible affine transformations \cite{wang2019deepdrawing}. Formally:

\begin{equation}
 R^2 = 1 - \frac{\big(\mathrm{Tr}\hspace{1pt}(P^T\hat{P}\hat{P}^TP)^{\frac{1}{2}}\big)^2}{\mathrm{Tr}\hspace{1pt}(P^TP)\mathrm{Tr}\hspace{1pt}(\hat{P}^T\hat{P})}
    \label{eq:procru}
\end{equation}

where $\mathrm{Tr}(\cdot)$ denotes the trace operator and the obtained metric $R^2$ assumes values in the interval $[0,1]$, the lower the better.
We will use the Procrustes Statistic-based similarity both as the loss function to guide the model training, and to evaluate its generalization capability on the test set.

We tested the proposed framework comparing the test performances obtained by the three different GNN models described in Table \ref{tab:gnn}, GCN, GAT, GIN.
All models are characterized by the ReLU non-linearity. The GAT model is composed by four attention heads. The $\epsilon$ variable in the GIN aggregation process is set to 0, as suggested in \cite{xu2019powerful}.
We leverage the PyTorch implementation of the models provided by the Deep Graph Library (DGL)\footnote{https://www.dgl.ai/}.

We searched for the best hyper-parameters selecting the models with the lowest validation error obtained during training, in the following grid of values: size of node hidden states $x_i$ in $\{ 10, 25, 50\}$; learning rate $\eta$ in $\{10^{-4}, 10^{-3}, 10^{-2}\}$; the number of GNN layers in $\{ 2, 3, 5\}$;   PE dimension  $k$ in $\{5, 8\}$ (20 is added to the grid in the case of the \textsc{Sparse} dataset, given its greater node number lowerbound); drop-out rate in $\{ 0.0, 0.1\}$. We considered 100 epochs of training with an early stopping strategy given by a patience on the validation loss of 20 epochs. 
For each epoch, we sampled non-overlapping mini-batches composed by $\beta$ graphs, until all the training data were considered. We searched for the best mini-batch size $\beta$ in $\{ 32, 64, 128\}$.
\mt{We devised several competitors in order to asses the performances of the proposed approach. Given that Laplacian PEs available at node-level are powerful descriptors of the neighboring graph structure, we leverage a Multilayer Perceptron (MLP) as a baseline. This neural predictor learns a mapping to the node coordinates, solely exploiting the available local information. 
We compare the performances obtained by GNNs with Laplacian PEs against those achieved by the three corresponding variants of rGNNs,  which we denote with rGCN, rGAT, rGIN.
For a fair comparison, we searched in the same hyperparameter space for all the baseline and competitors. }

\begin{figure*}[th!]
\noindent\rule{\linewidth}{0.4pt} \vspace{-0.85cm}\\

\hspace{4.5cm} \textsc{Rome} $\qquad$ \hspace{6.8cm} \textsc{Sparse} $\quad $ \vspace{-0.65cm}\\

\hspace{.5cm} \noindent\rule{8cm}{0.4pt} $\quad$    \noindent\rule{8.5cm}{0.4pt} \vspace{-0.4cm} \\

\hspace{0.9cm}  \textsc{GT} $\qquad $  \hspace{.5cm}     \textsc{GCN}  \hspace{.7cm} $\quad $ \textsc{GAT} $\quad $   \hspace{.6cm} \textsc{GIN}   \hspace{1.6cm}  \textsc{GT} $\qquad $  \hspace{.5cm}     \textsc{GCN}  \hspace{.6cm} $\quad $ \textsc{GAT} $\quad $   \hspace{.9cm} \textsc{GIN}  \\

    \centering

         \includegraphics[width=0.095\paperwidth, trim= 20 -50 20 20, clip]{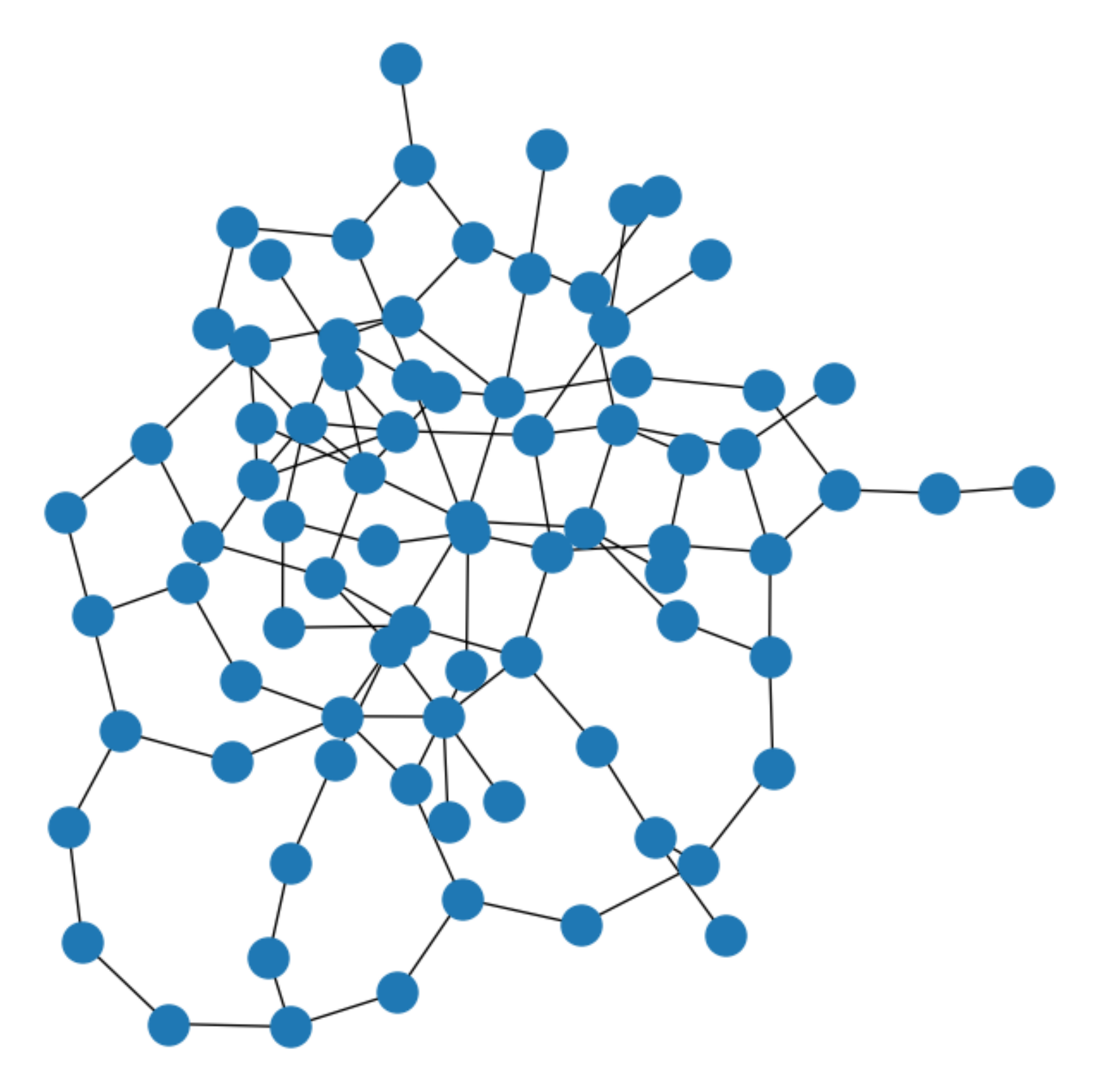}
         \includegraphics[width=0.095\paperwidth, trim= 20 0 10 0, clip]{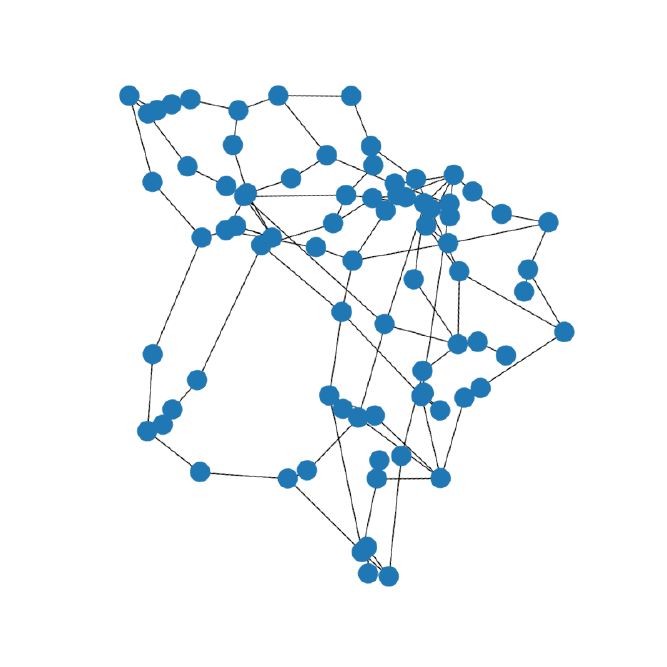}
         \includegraphics[width=0.095\paperwidth, trim= 20 0 20 0, clip]{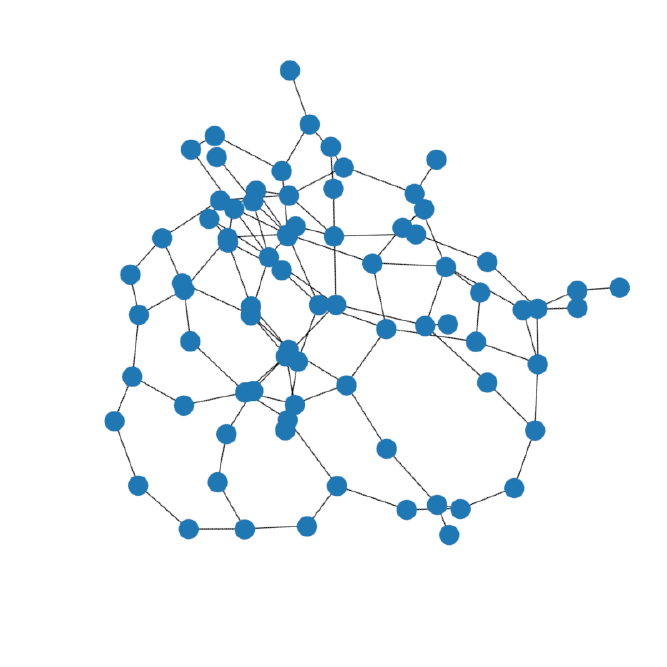}
         \includegraphics[width=0.095\paperwidth, trim= 30 -50 50 20, clip]{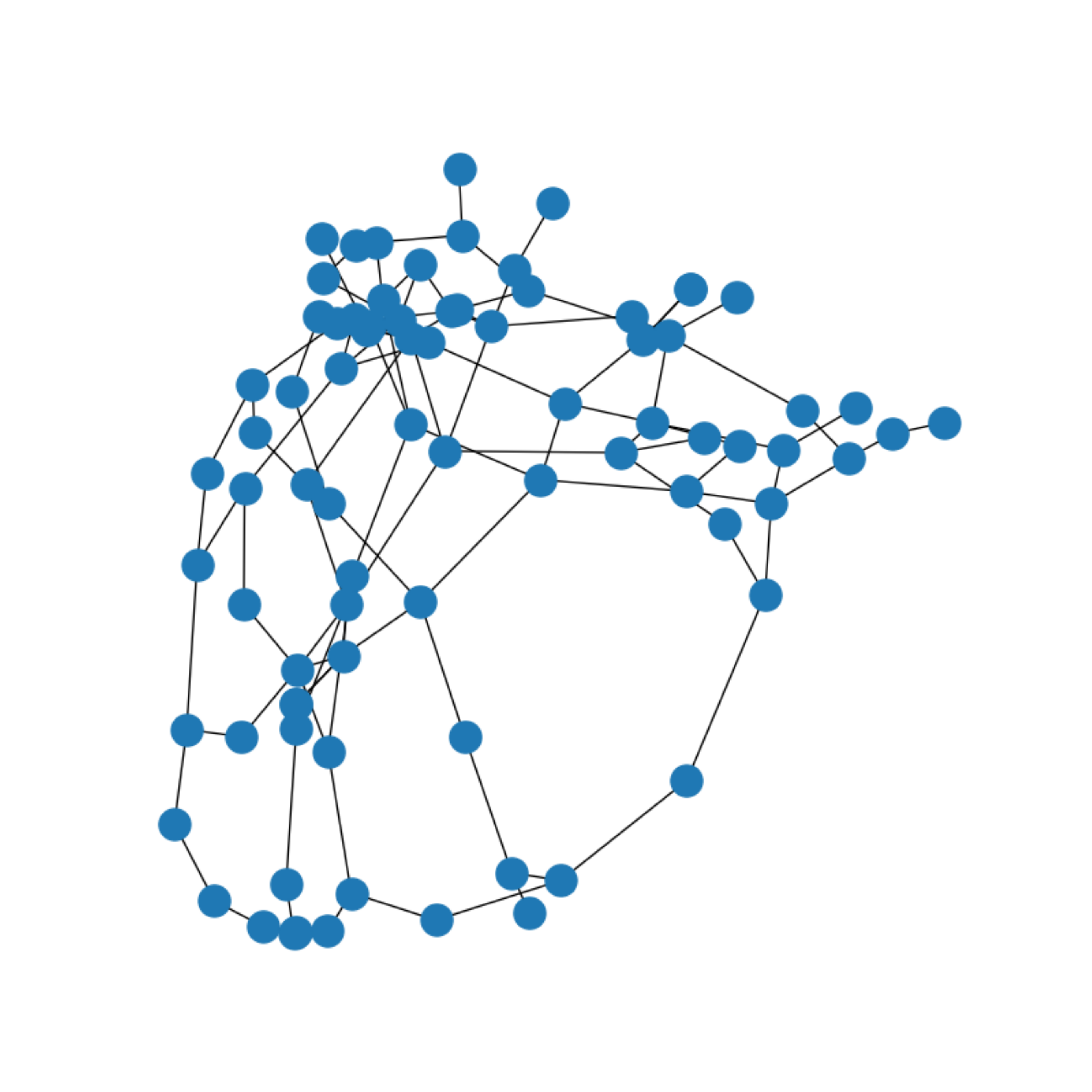} \hspace{0.2cm}
         \includegraphics[width=0.095\paperwidth, trim= 20 0 20 -30, clip]{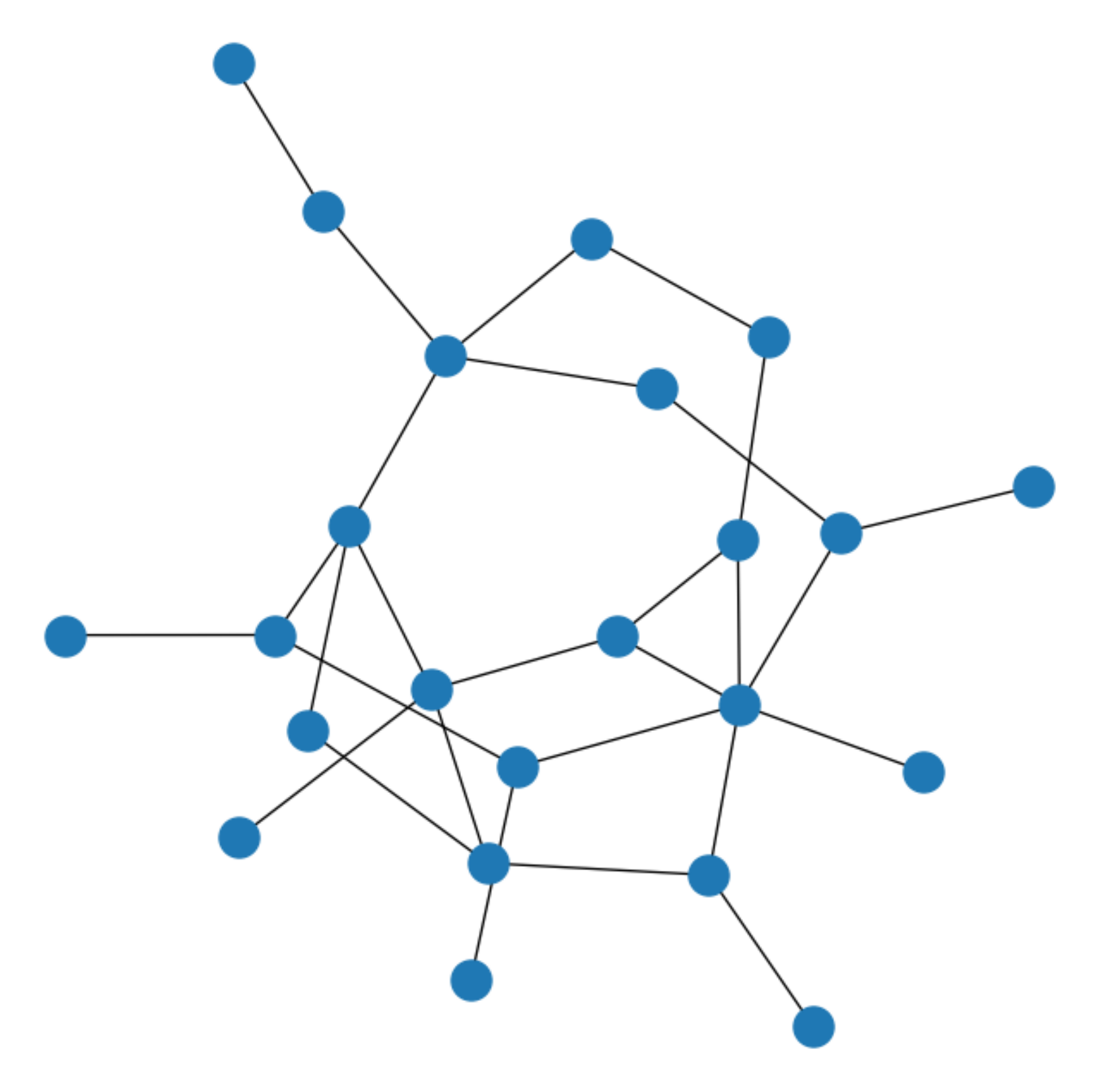}
         \includegraphics[width=0.095\paperwidth, trim= 30 0 30 20, clip]{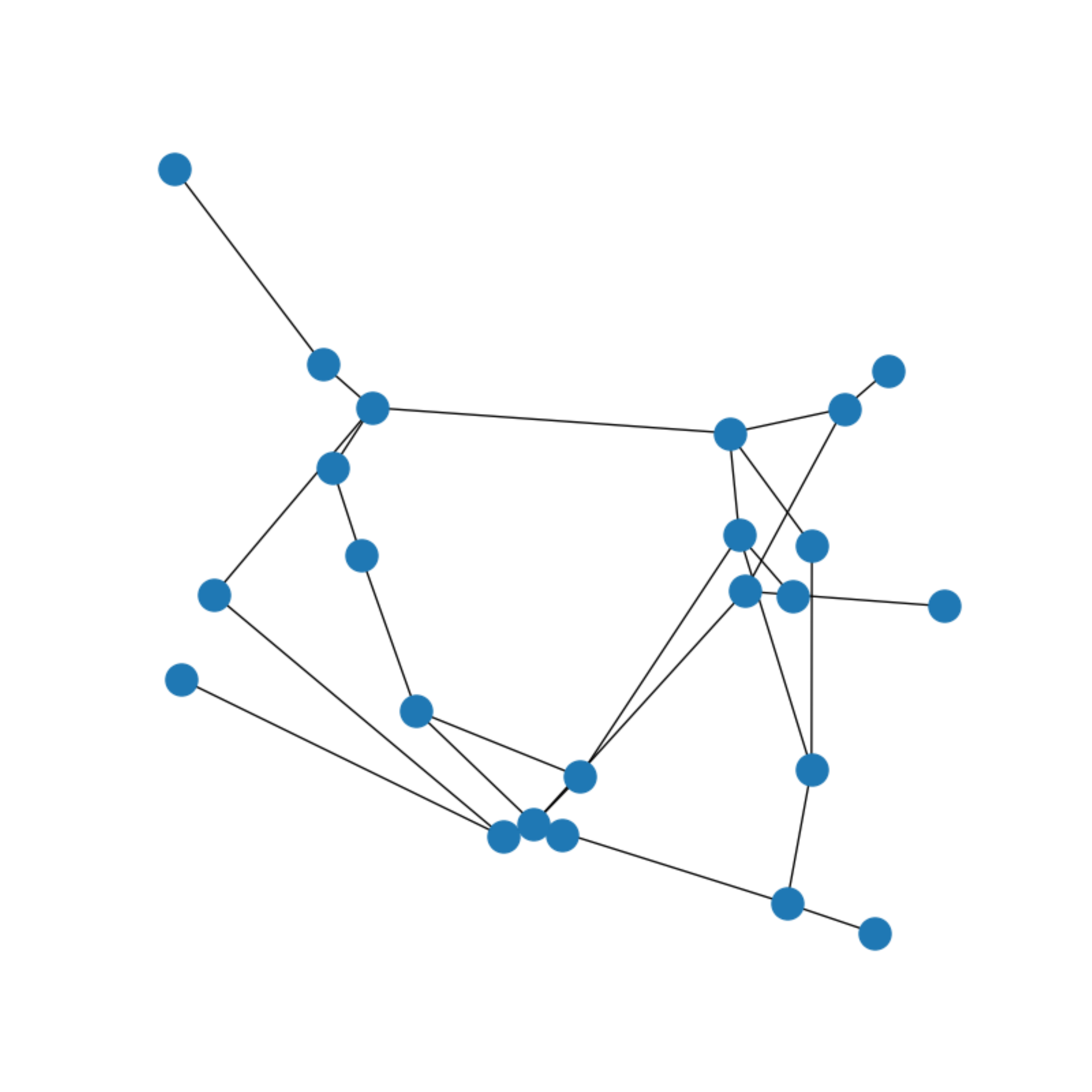}
         \includegraphics[width=0.095\paperwidth, trim= 20 20 20 0, clip]{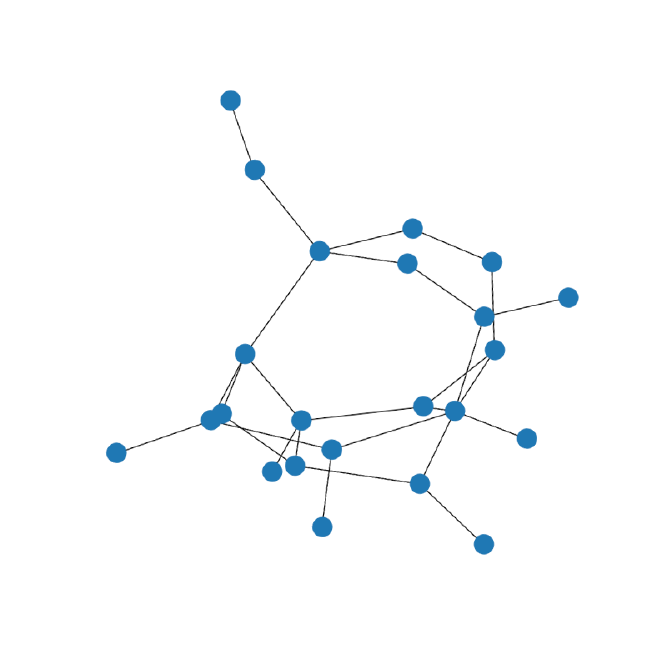}
         \includegraphics[width=0.095\paperwidth, trim= 20 0 10 0, clip]{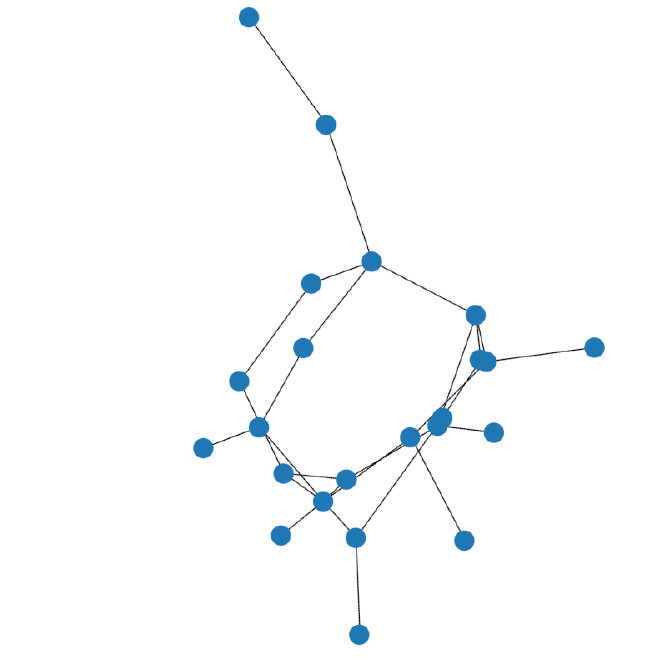}
         \includegraphics[width=0.095\paperwidth, trim= 20 0 20 0, clip]{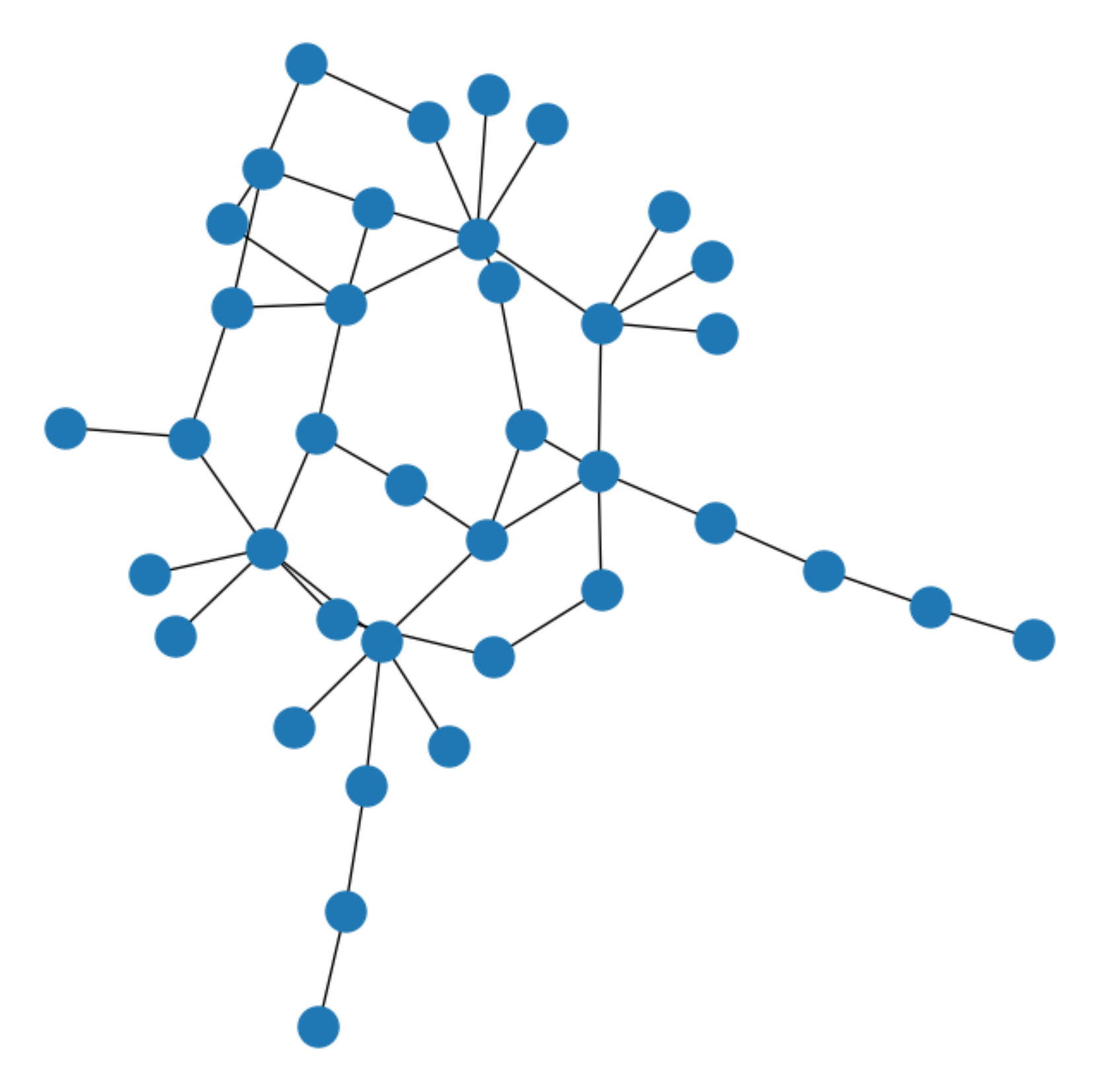}
         \includegraphics[width=0.095\paperwidth, trim= 20 0 10 10, clip]{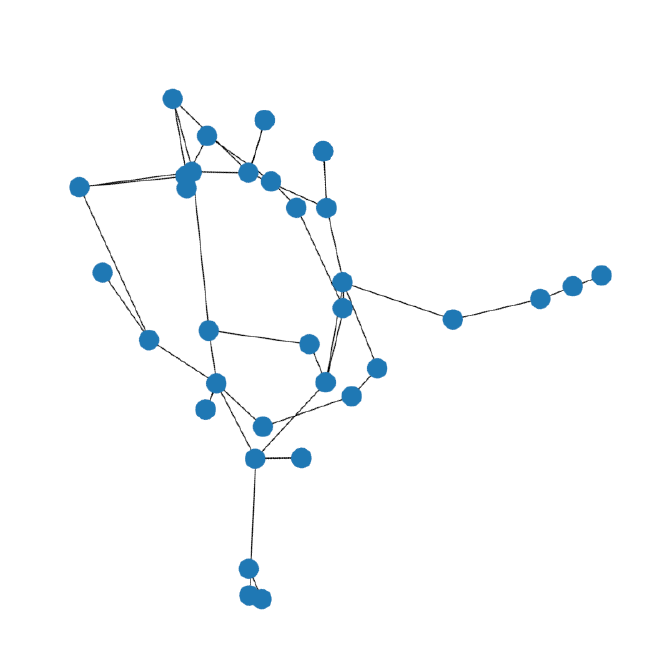}
         \includegraphics[width=0.095\paperwidth, trim= 50 0 50 0, clip]{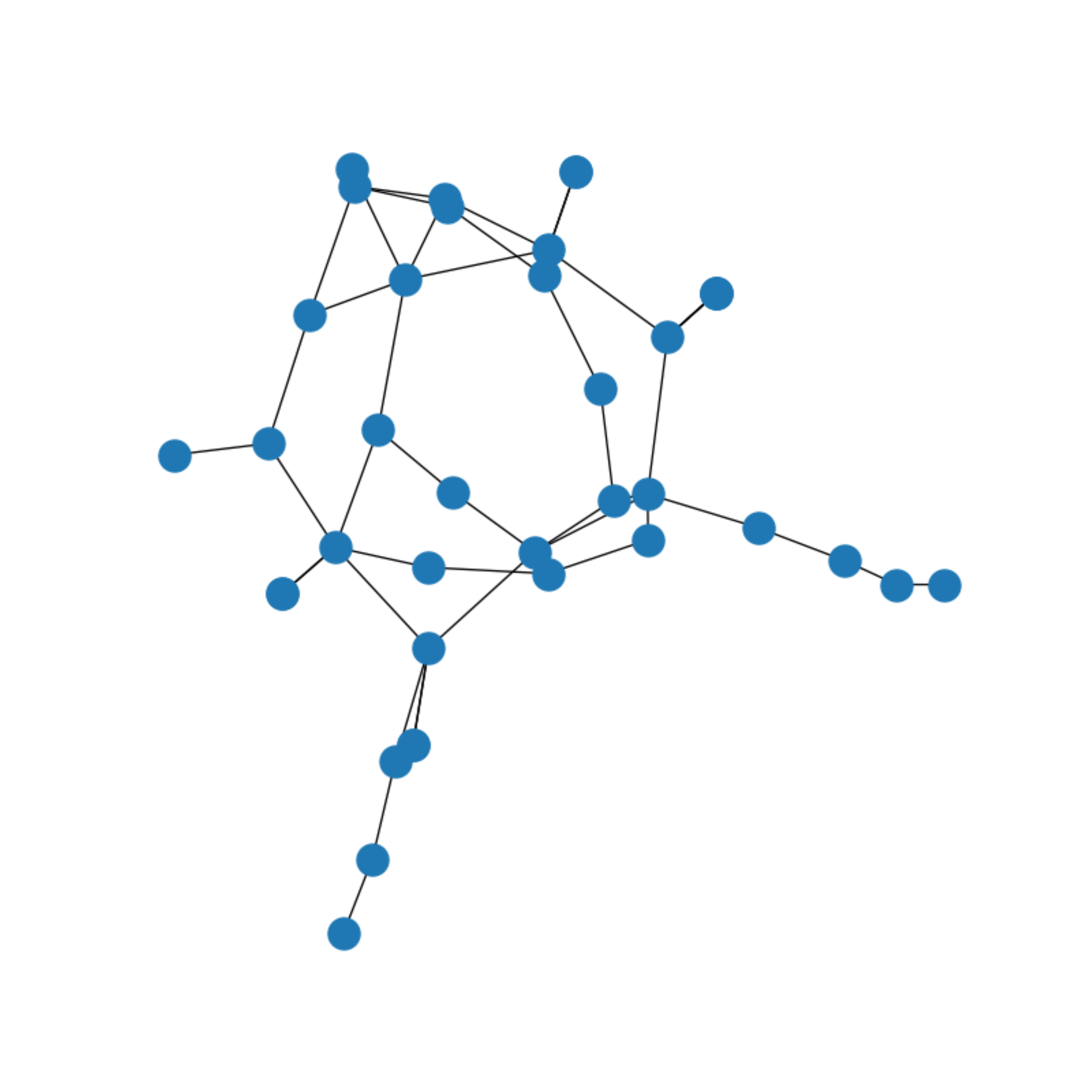}
         \includegraphics[width=0.095\paperwidth, trim= 50 0 50 0, clip]{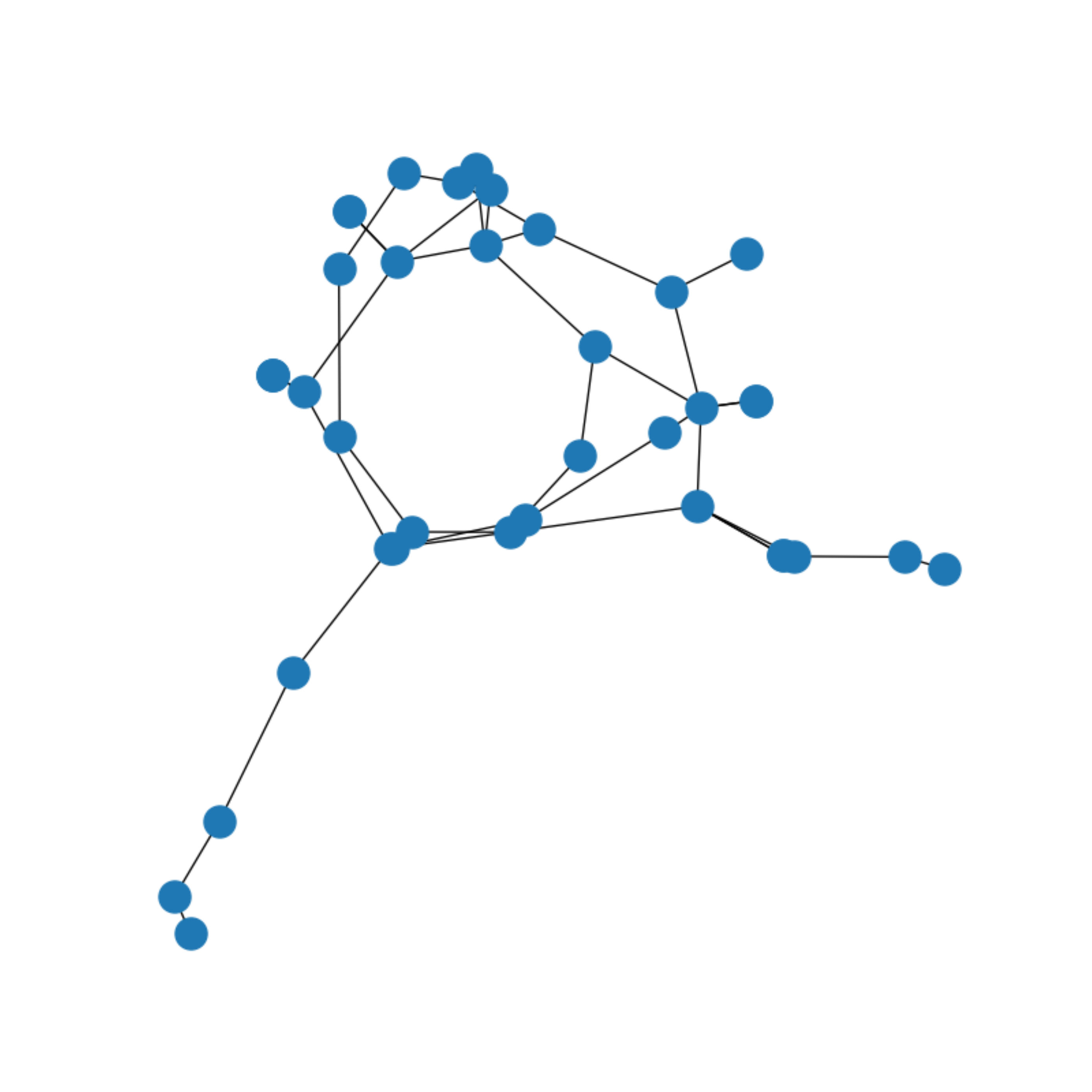} \hspace{0.2cm}
         \includegraphics[width=0.095\paperwidth, trim= 20 0 20 0, clip]{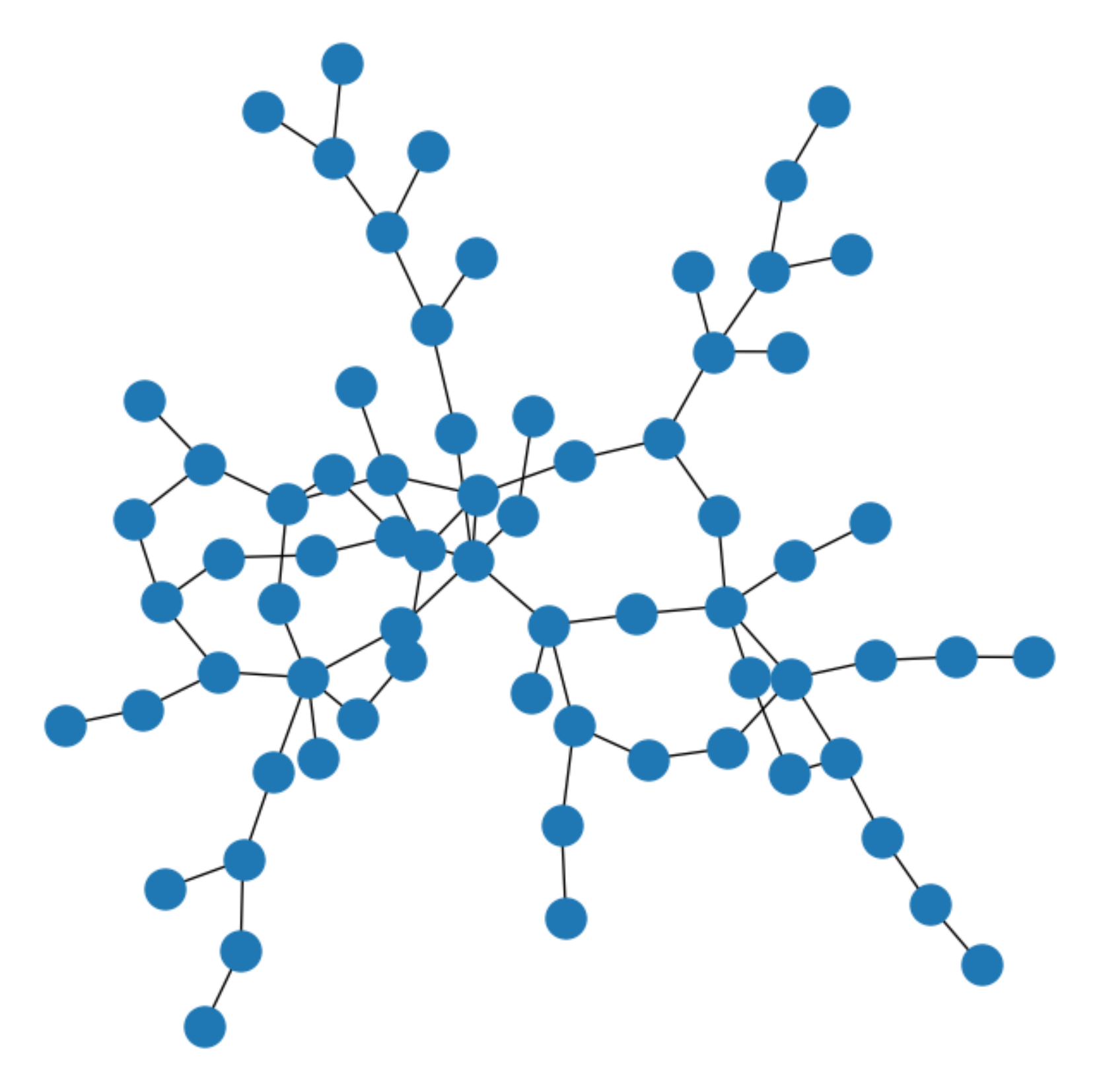}
         \includegraphics[width=0.095\paperwidth, trim= 30 0 30 0, clip]{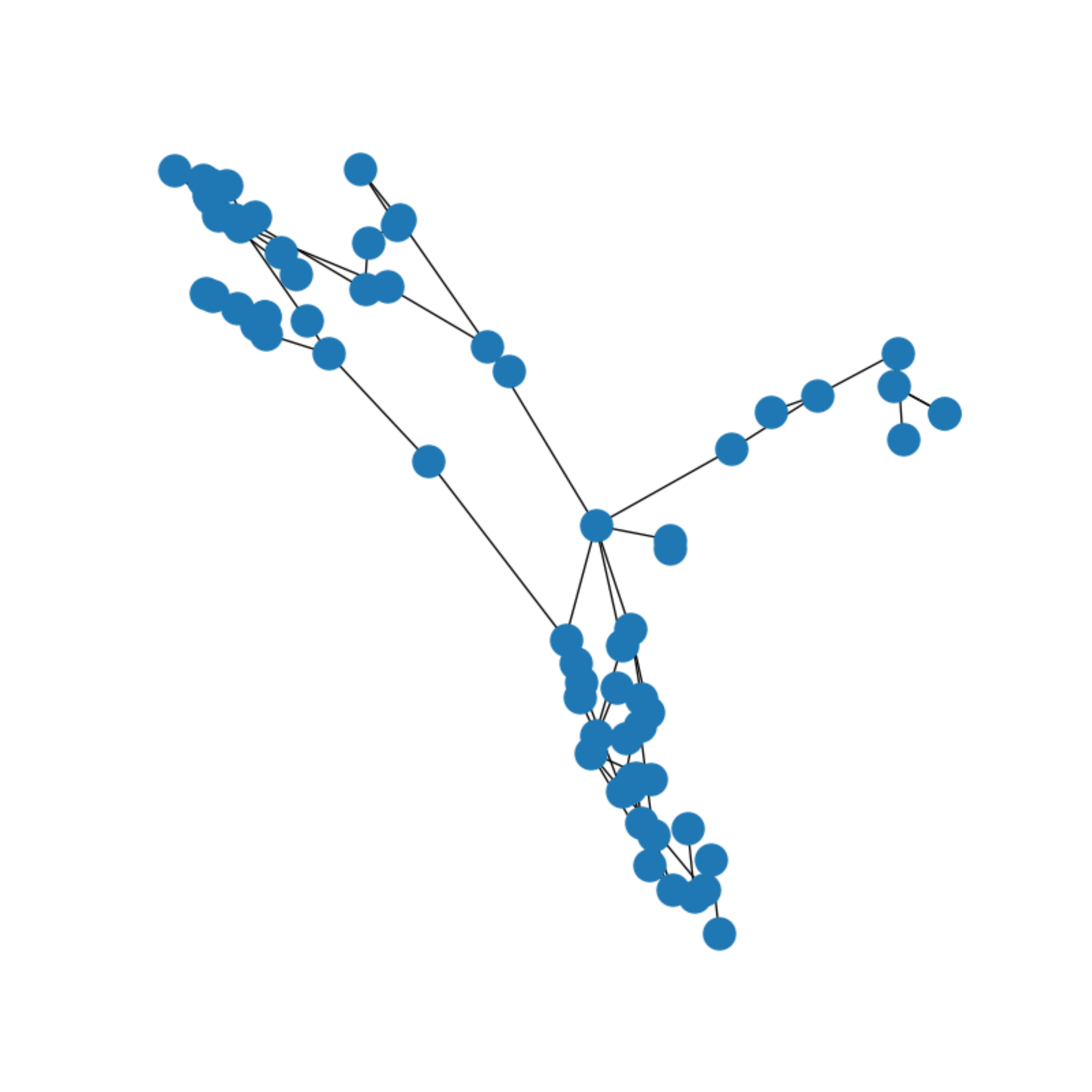}
         \includegraphics[width=0.095\paperwidth, trim= 0 0 40 0, clip]{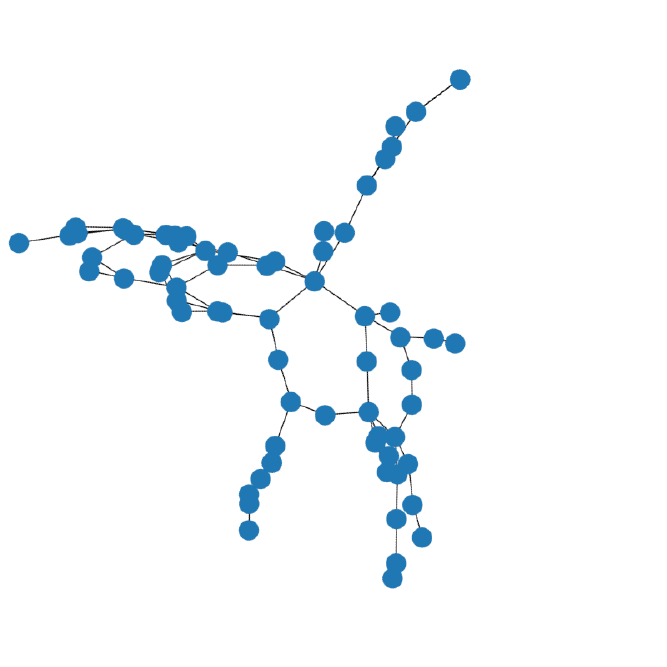}
         \includegraphics[width=0.095\paperwidth, trim= 20 30 20 0, clip]{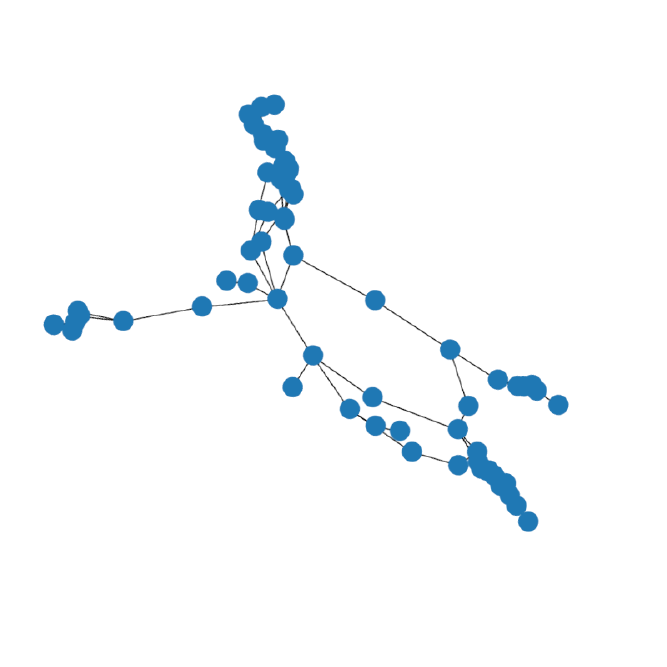}
         \includegraphics[width=0.095\paperwidth, trim= 10 0 10 0, clip]{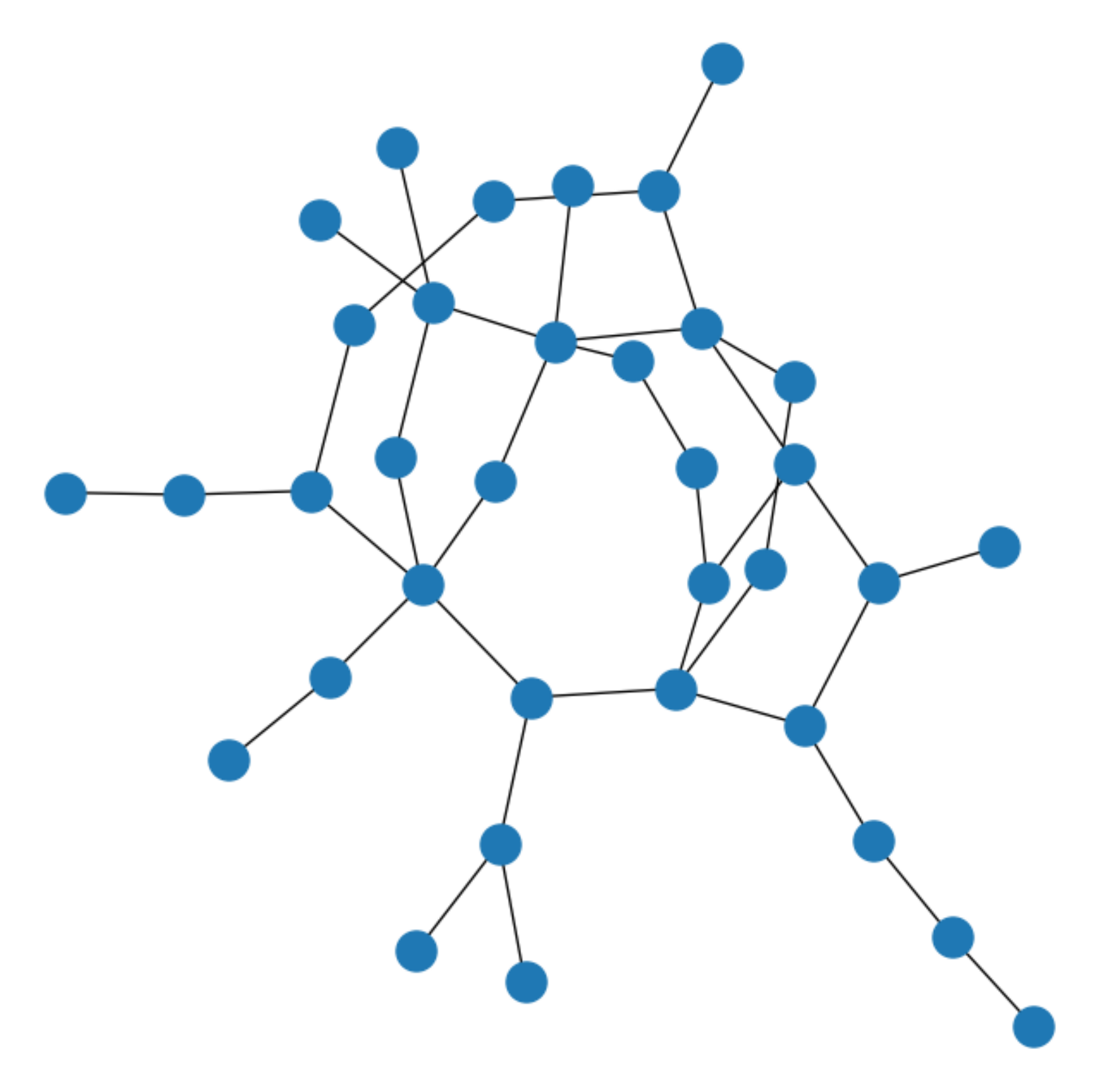}
         \includegraphics[width=0.095\paperwidth, trim= 10 10 10 0, clip]{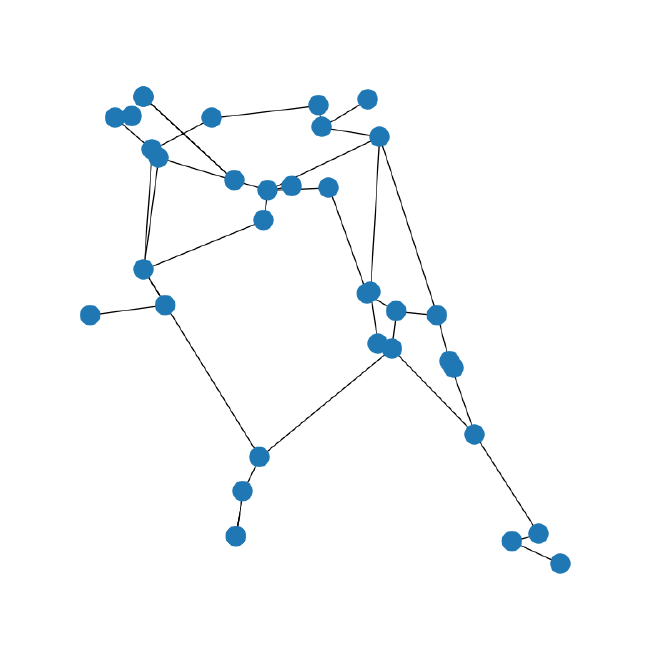}
         \includegraphics[width=0.095\paperwidth, trim= 20 0 20 20, clip]{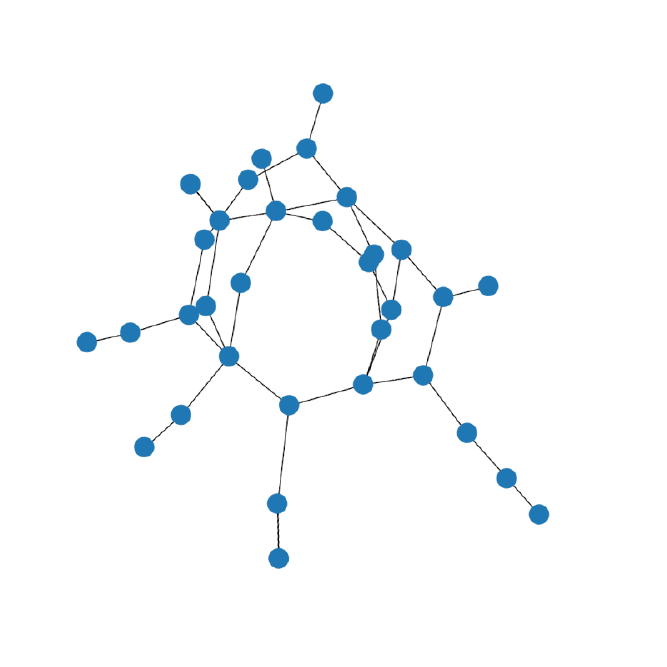}
         \includegraphics[width=0.095\paperwidth, trim= 20 0 20 0, clip]{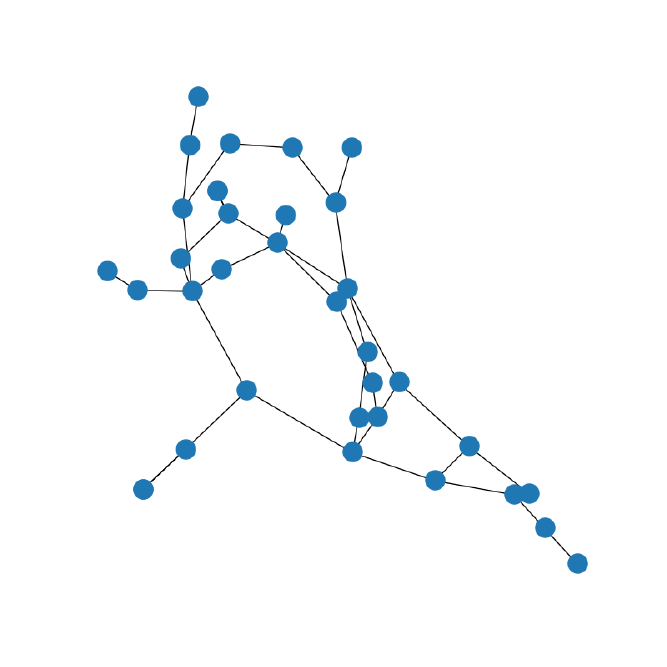} \hspace{0.2cm}
         \includegraphics[width=0.095\paperwidth, trim= 20 0 20 20, clip]{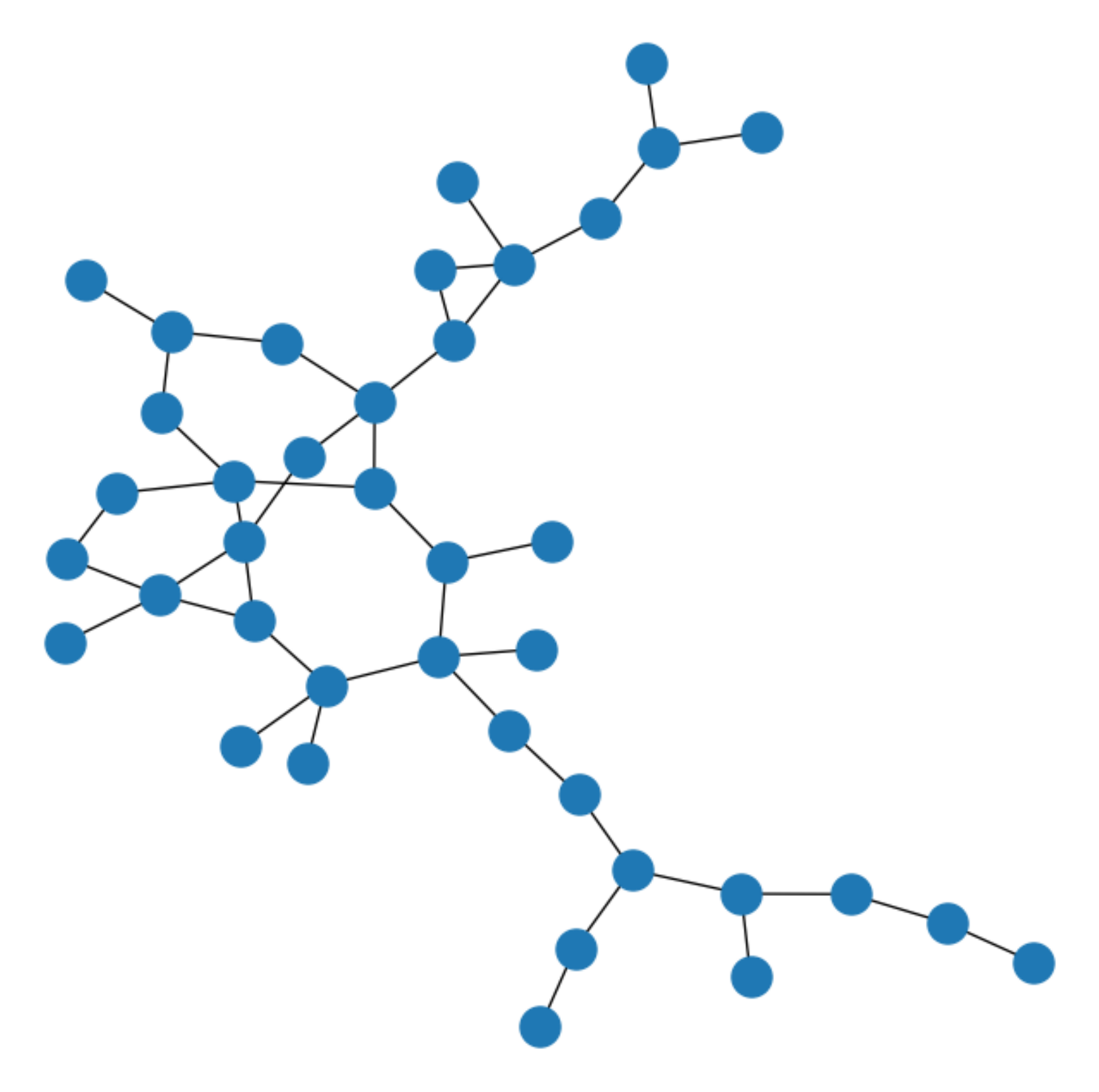}
         \includegraphics[width=0.095\paperwidth, trim= 20 10 20 0, clip]{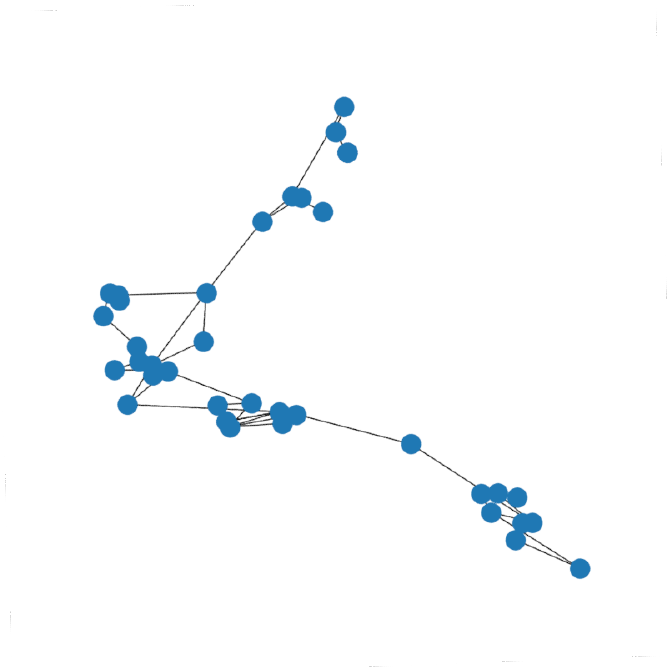}
         \includegraphics[width=0.095\paperwidth, trim= 20 10 30 0, clip]{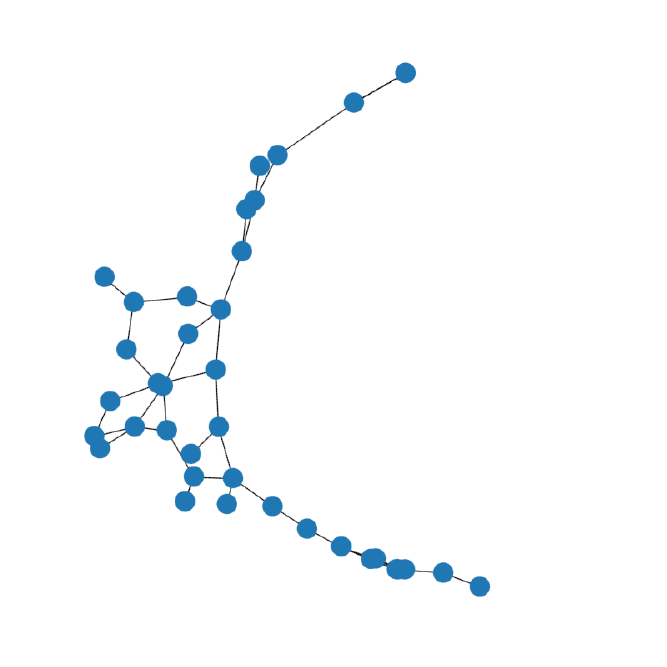}
         \includegraphics[width=0.095\paperwidth, trim= 20 10 30 0, clip]{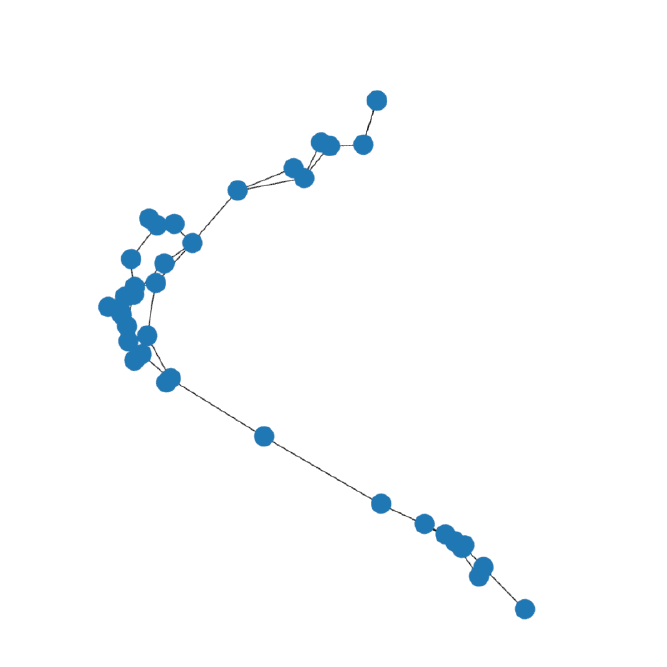}

    \caption{\textsc{Kamada-Kawai} layout. Qualitative example of the predicted node coordinates for both the \textsc{Rome} dataset (first four column) and the \textsc{Sparse} dataset (subsequent four columns).
    Each row depicts the Ground-Truth positions (GT), the graph layout produced by GCN, GAT, GIN model, left-to-right. We report the predictions on three different test graphs (rows).}
    \label{fig:sup_exp_kamada}
\end{figure*}

\begin{table}[th!]

\centering
 \caption{Procrustes Statistic similarity (defined in Eq. \ref{eq:procru}) on the test split of the \textsc{Rome } and \textsc{Sparse} dataset. We compare three GND models with two graph layouts generation, \textsc{Kamada-Kawai} and \textsc{Spectral}. We report the average values and standard deviations over three runs with different weights initialization.}
 
\resizebox{\columnwidth}{!}{%
    \begin{tabular}{llcccc}
    \toprule 
    & \multirow{2}{*}{Model} & \multicolumn{2}{c}{\textsc{Rome}} & \multicolumn{2}{c} {\textsc{Sparse}} \\  
    \cmidrule(l){3-4}   \cmidrule(l){5-6}   
    & &  \textsc{Kamada}  & \textsc{Spectral}   &  \textsc{Kamada}  & \textsc{Spectral}  \\
     \midrule
     
    \multirow{4}{*}{\rotatebox{90}{\textsc{Compet.}}} & \mt{MLP} & 0.291 $\pm$ 0.000 & 0.144 $\pm$ 0.000 & 0.282 $\pm$ 0.001 & 0.131 $\pm$ 0.000 \\
    & \mt{rGCN} & 0.612 $\pm$ 0.009 & 0.527 $\pm$ 0.008 & 0.592 $\pm$ 0.006 & 0.532 $\pm$ 0.009 \\
     &\mt{rGAT} & 0.475 $\pm$ 0.008 & 0.437 $\pm$ 0.008 & 0.465 $\pm$ 0.015 & 0.425 $\pm$ 0.017 \\
    & \mt{rGIN} & 0.685 $\pm$ 0.001 & 0.590 $\pm$ 0.003 & 0.675 $\pm$ 0.010 & 0.603 $\pm$ 0.030 \\
    \midrule
    \multirow{3}{*}{\rotatebox{90}{\textsc{\textbf{GND}}}}    & GCN & 0.241 $\pm$ 0.001 & $0.078 \pm 0.002$ & $0.228 \pm 0.000$ &  $0.060 \pm 0.000$ \\ 
        & GAT & \textbf{0.186} $\pm$ 0.000  & \textbf{0.057} $\pm$ 0.000 & \textbf{0.177} $\pm$ 0.001 &  \textbf{0.045} $\pm$ 0.001 \\ 
    &    GIN & $0.240 \pm 0.002$ & $0.076 \pm 0.003$ & $0.229$ $\pm$ 0.003 &  $0.059 \pm 0.001$ \\ 

    \bottomrule
    \end{tabular}
}
 \label{tab:sup_exp}
\end{table}

\begin{figure*}[th!]
\noindent\rule{\linewidth}{0.4pt} \vspace{-0.85cm}\\

\hspace{4.5cm} \textsc{Rome} $\qquad$ \hspace{6.8cm} \textsc{Sparse} $\quad $ \vspace{-0.65cm}\\

\hspace{.5cm} \noindent\rule{8cm}{0.4pt} $\quad$    \noindent\rule{8.5cm}{0.4pt} \vspace{-0.4cm} \\

\hspace{0.9cm}  \textsc{GT} $\qquad $  \hspace{.5cm}     \textsc{GCN}  \hspace{.7cm} $\quad $ \textsc{GAT} $\quad $   \hspace{.6cm} \textsc{GIN}   \hspace{1.6cm}  \textsc{GT} $\qquad $  \hspace{.5cm}     \textsc{GCN}  \hspace{.6cm} $\quad $ \textsc{GAT} $\quad $   \hspace{.9cm} \textsc{GIN}  \\

    \centering

         \includegraphics[width=0.095\paperwidth, trim= 20 0 20 10, clip]{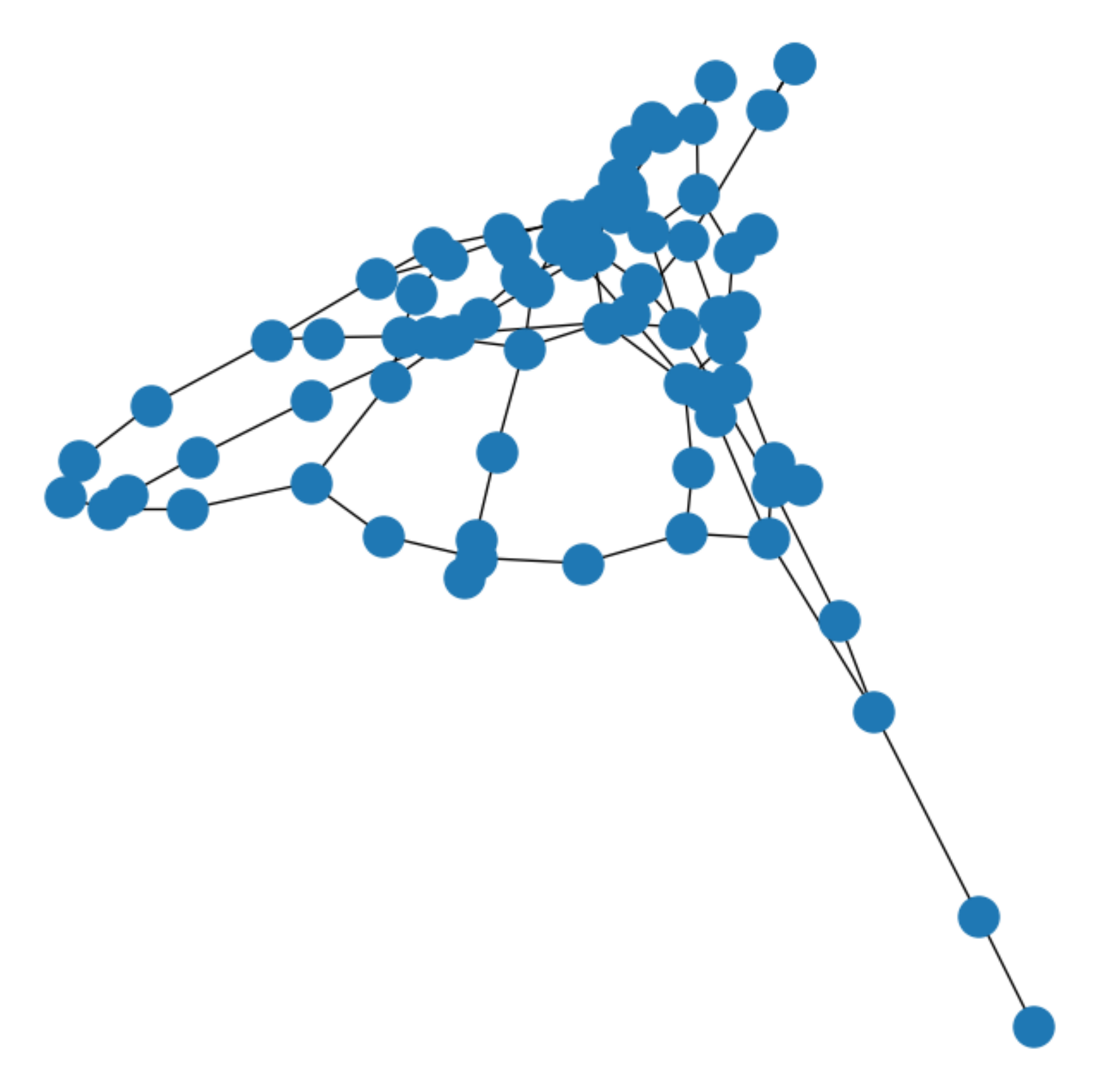}
         \includegraphics[width=0.095\paperwidth, trim= 0 0 20 0, clip]{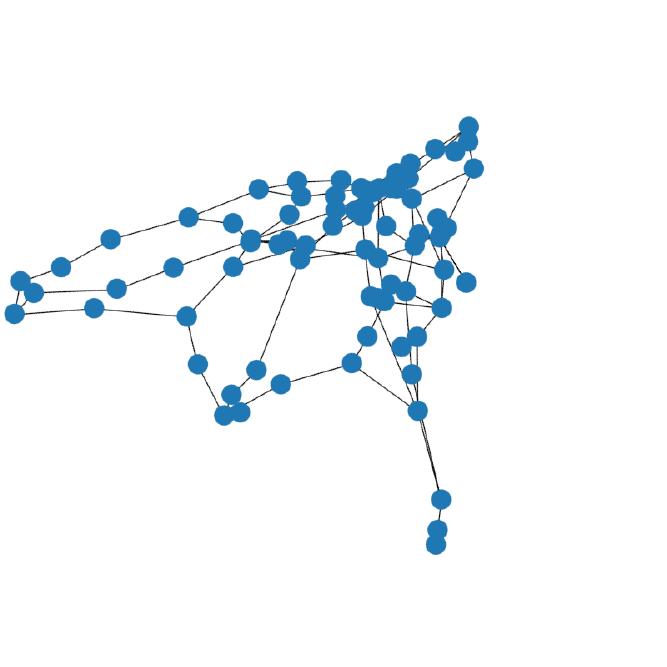}
         \includegraphics[width=0.095\paperwidth, trim= 0 0 20 0, clip]{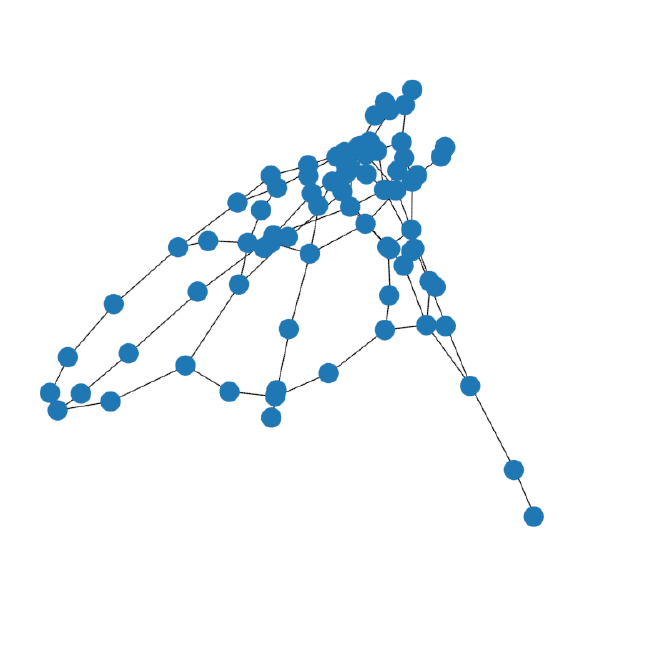}
         \includegraphics[width=0.095\paperwidth, trim= 0 0 20 0, clip]{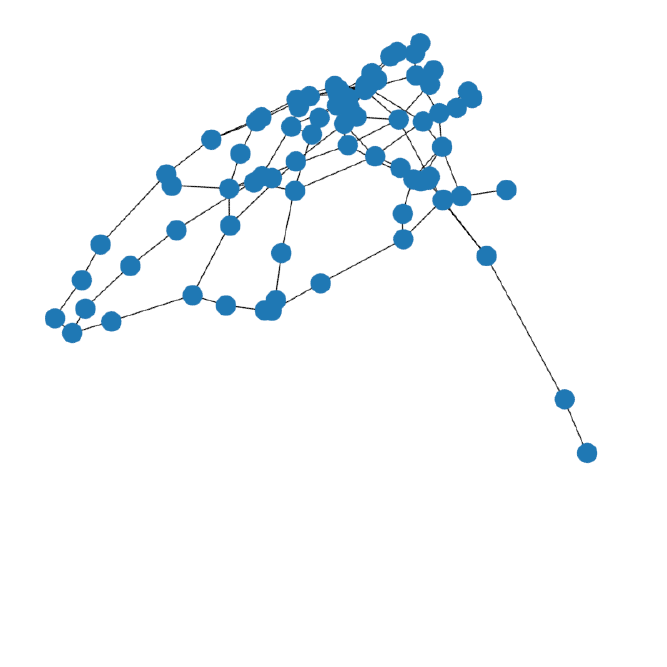} \hspace{0.2cm}
         \includegraphics[width=0.095\paperwidth, trim= 20 -30 20 0, clip]{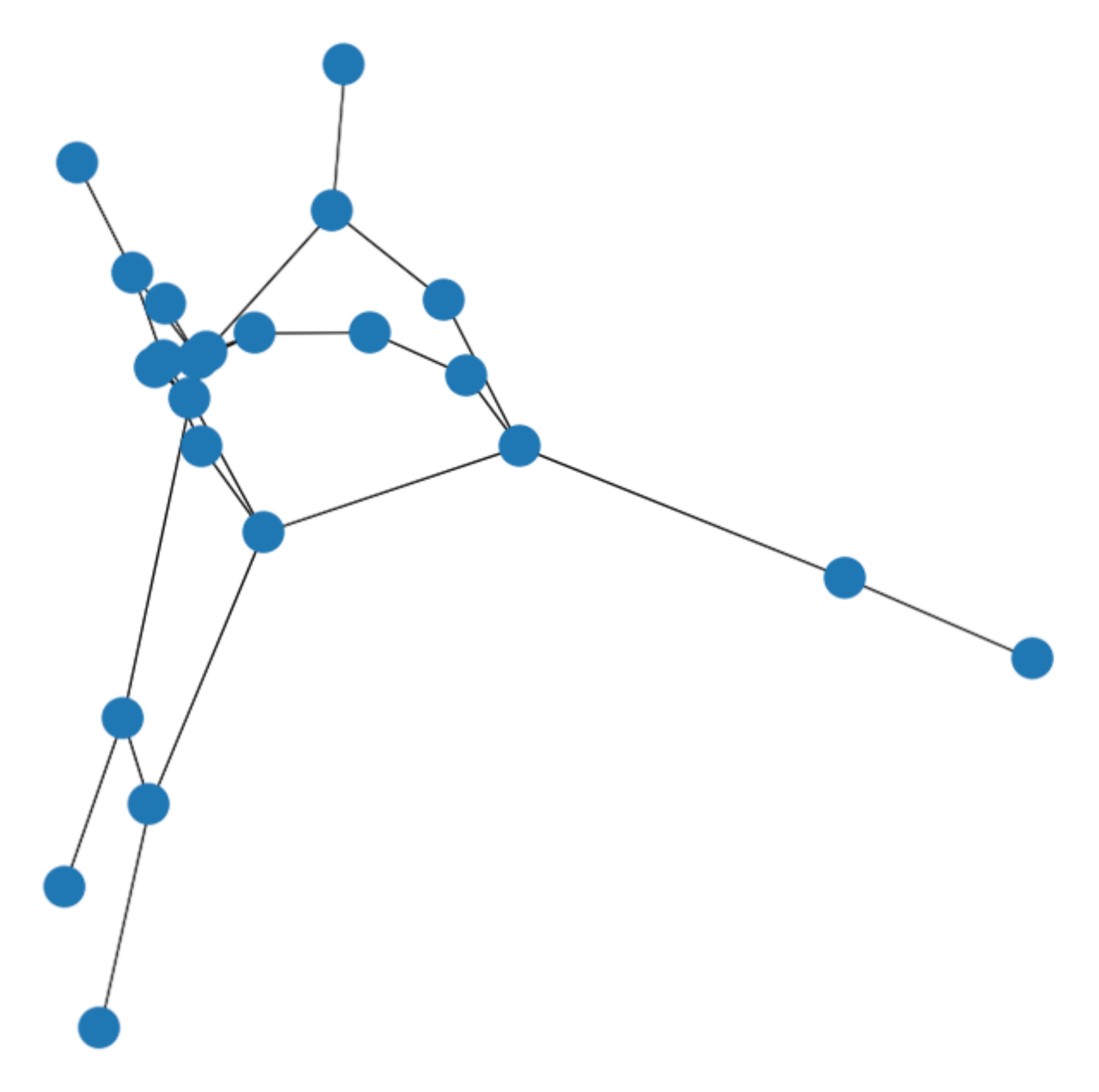}
         \includegraphics[width=0.095\paperwidth, trim= 10 0 10 0, clip]{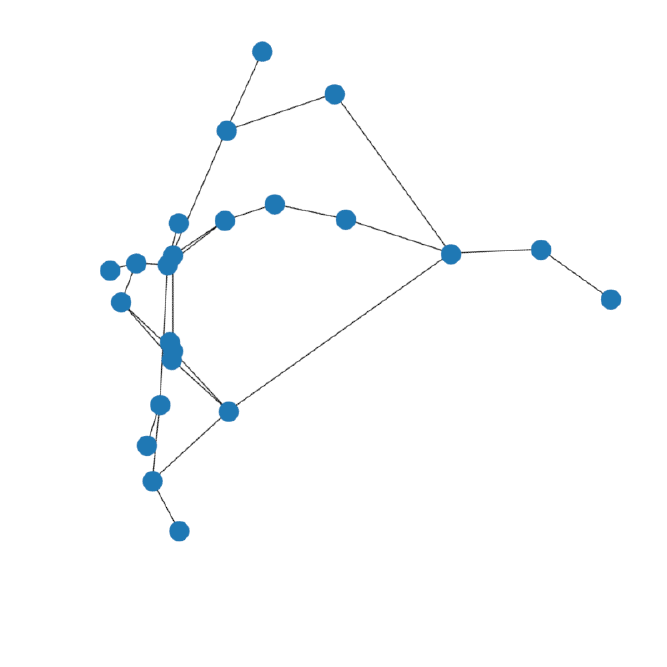}
         \includegraphics[width=0.095\paperwidth, trim= 20 0 20 0, clip]{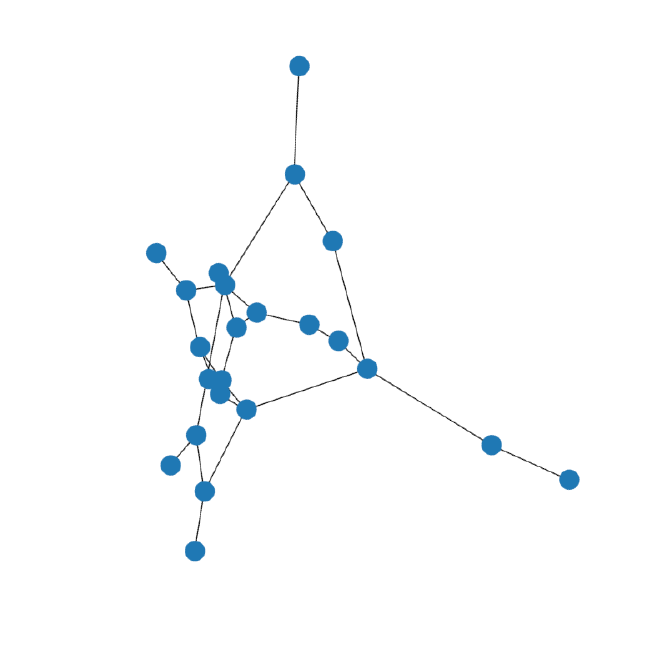}
         \includegraphics[width=0.095\paperwidth, trim= 20 0 0 0, clip]{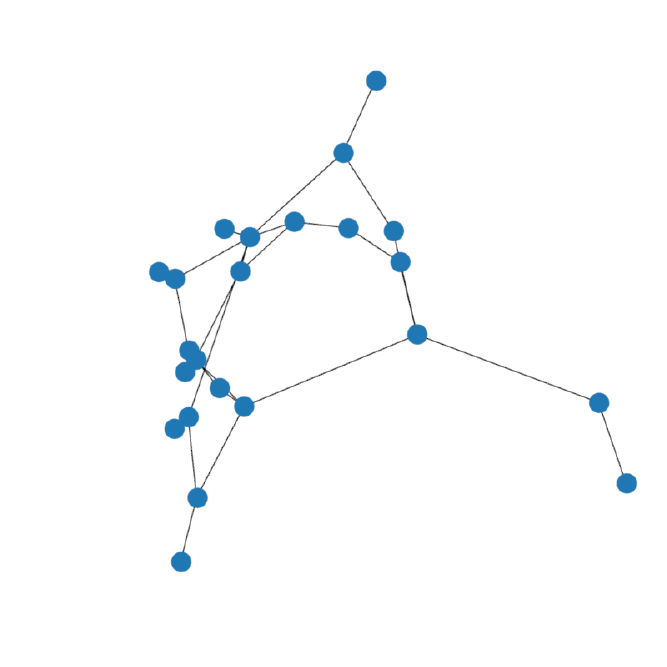}
         \includegraphics[width=0.095\paperwidth, trim= 20 -30 20 0, clip]{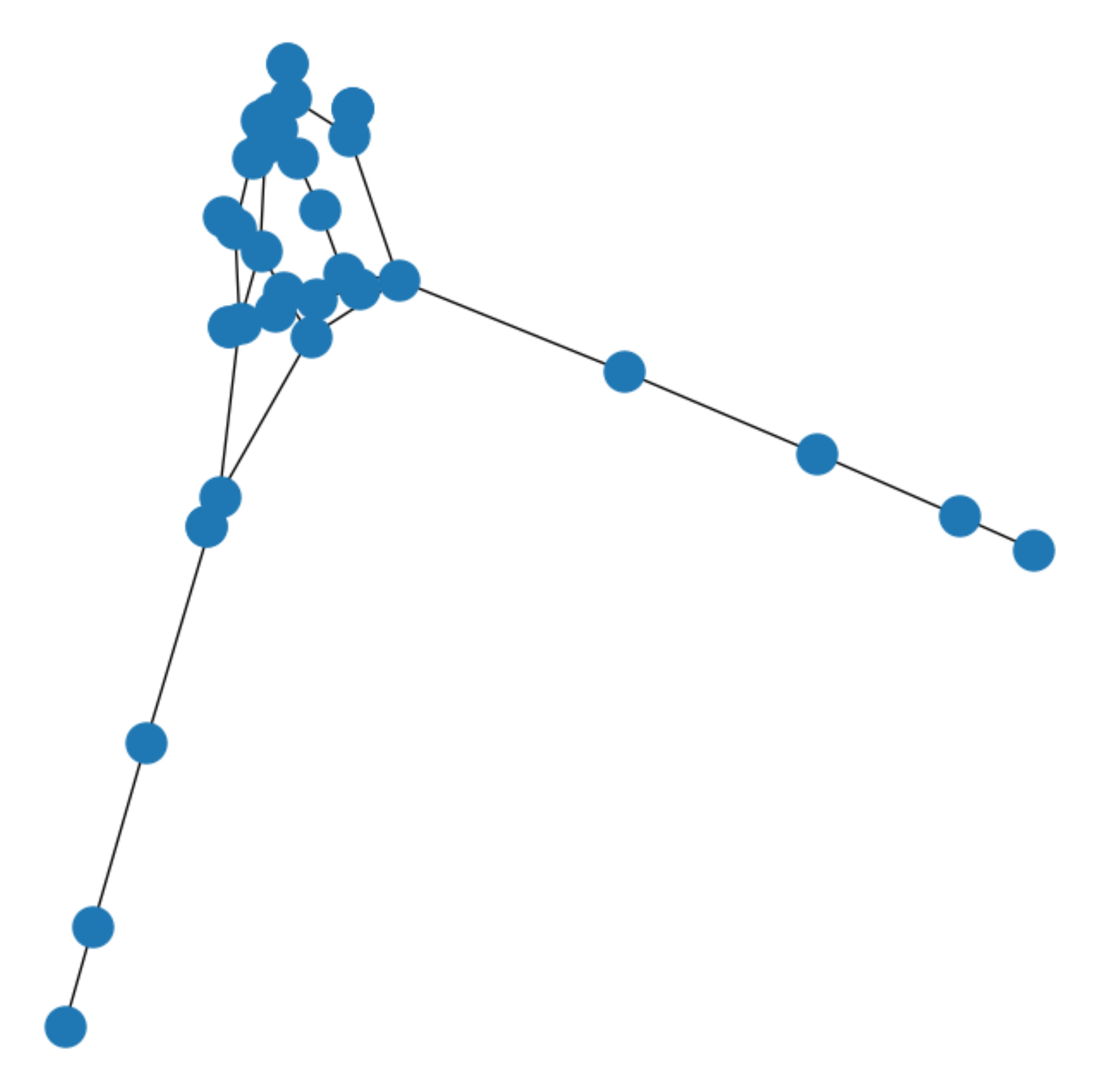}
         \includegraphics[width=0.095\paperwidth, trim= 20 0 20 0, clip]{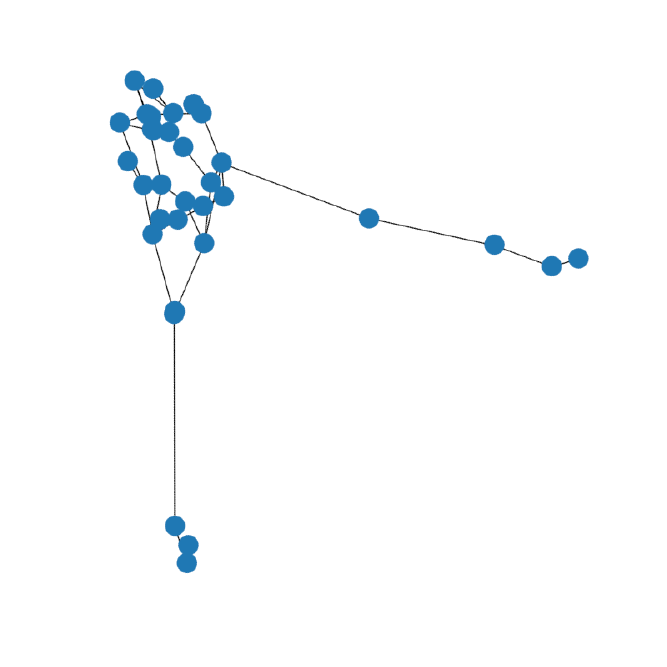}
         \includegraphics[width=0.095\paperwidth, trim= 20 0 20 0, clip]{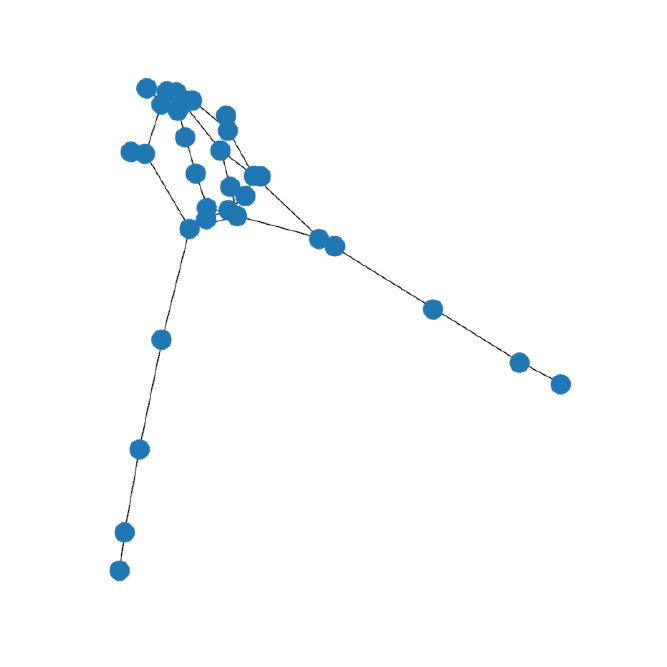}
         \includegraphics[width=0.095\paperwidth, trim= 20 0 20 0, clip]{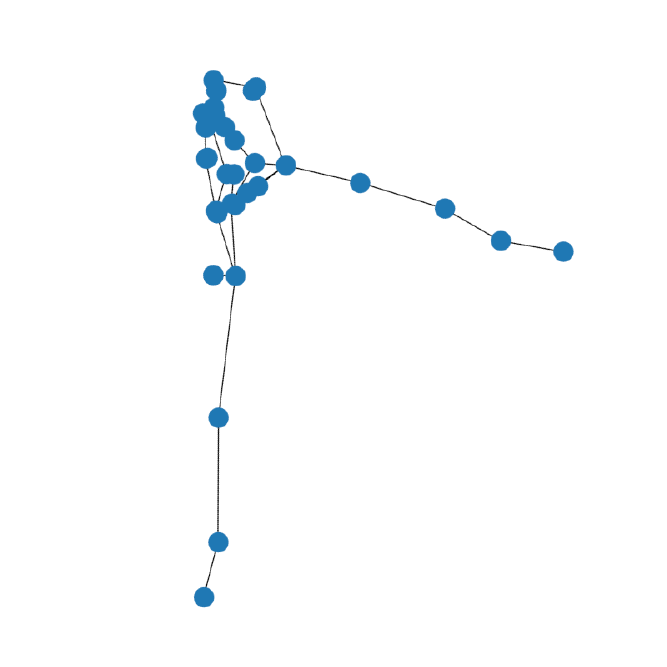} \hspace{0.2cm}
         \includegraphics[width=0.095\paperwidth, trim= 10 -20 10 0, clip]{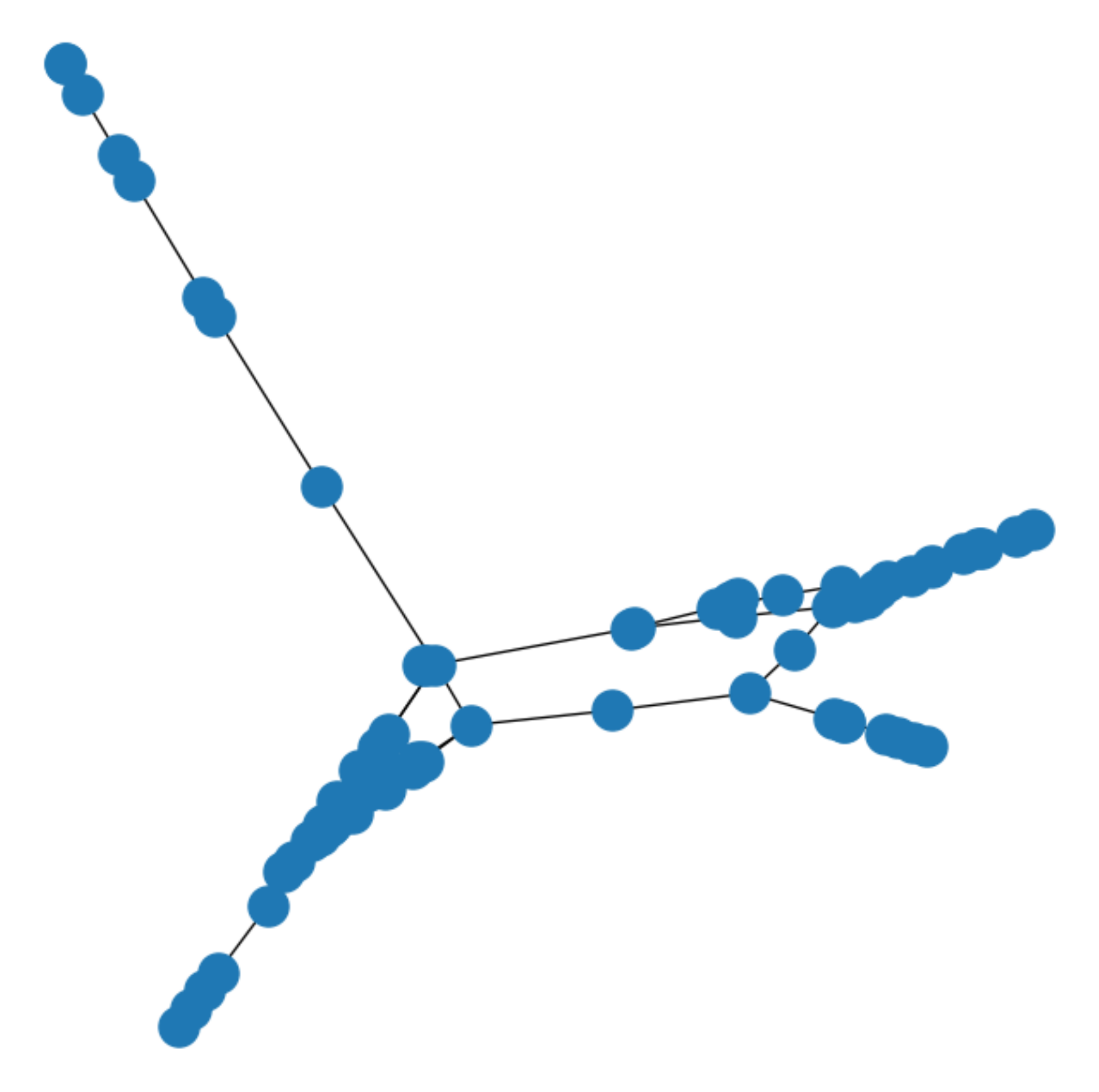}
         \includegraphics[width=0.095\paperwidth, trim= 30 0 0 0, clip]{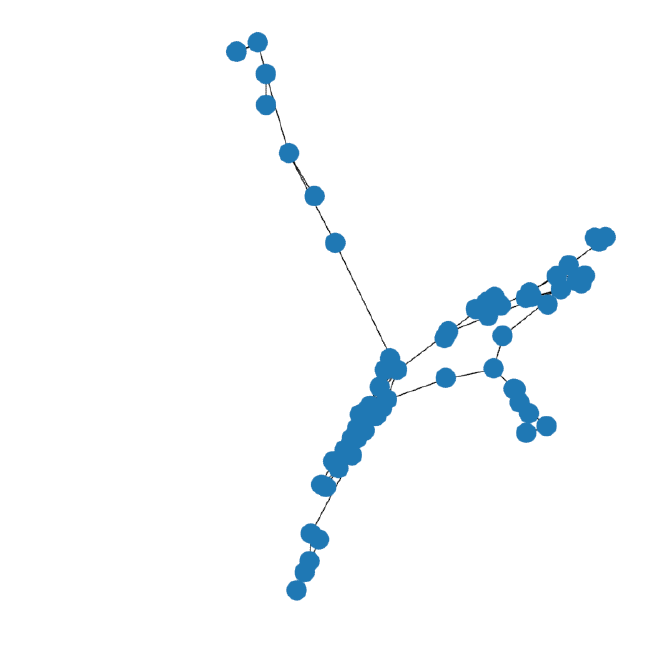}
         \includegraphics[width=0.095\paperwidth, trim= 20 0 20 0, clip]{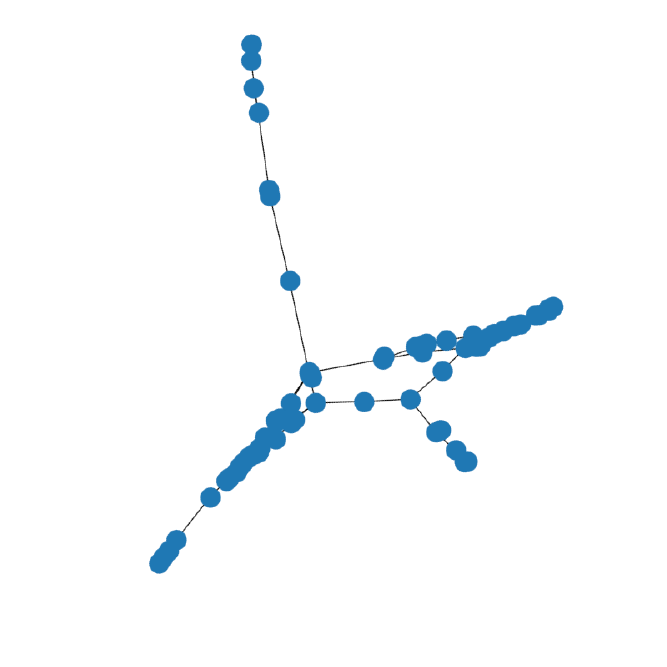}
         \includegraphics[width=0.095\paperwidth, trim= 20 0 20 0, clip]{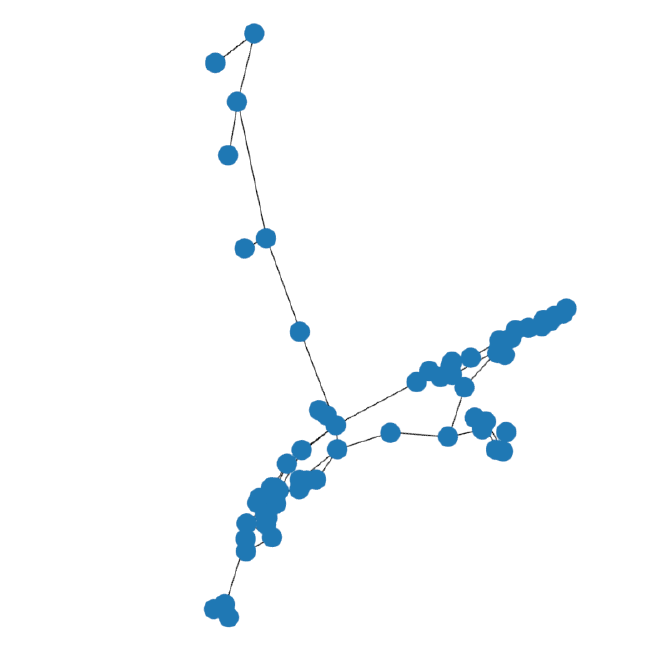}
         \includegraphics[width=0.095\paperwidth, trim= 20 0 20 0, clip]{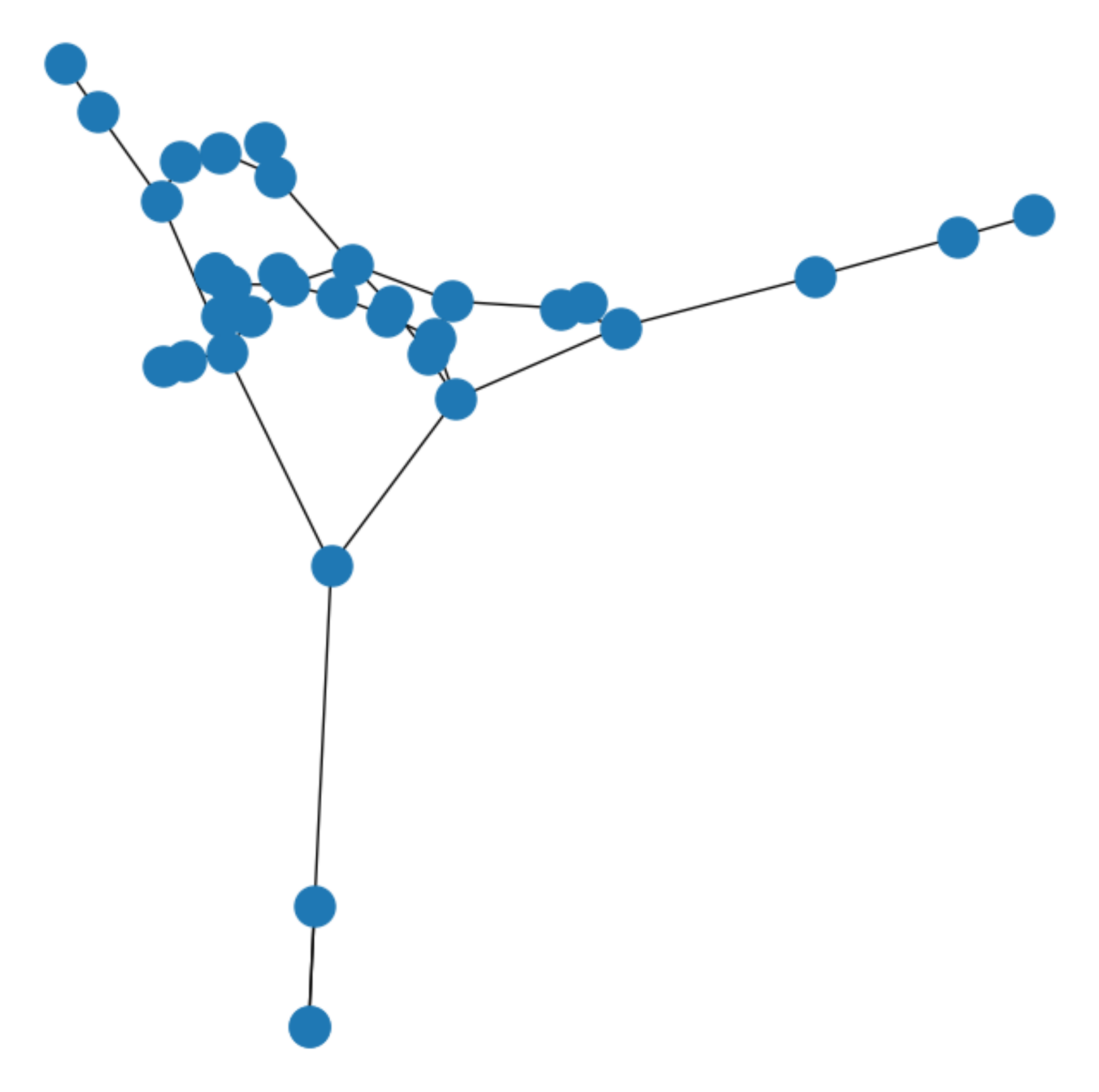}
         \includegraphics[width=0.095\paperwidth, trim= 10 0 20 0, clip]{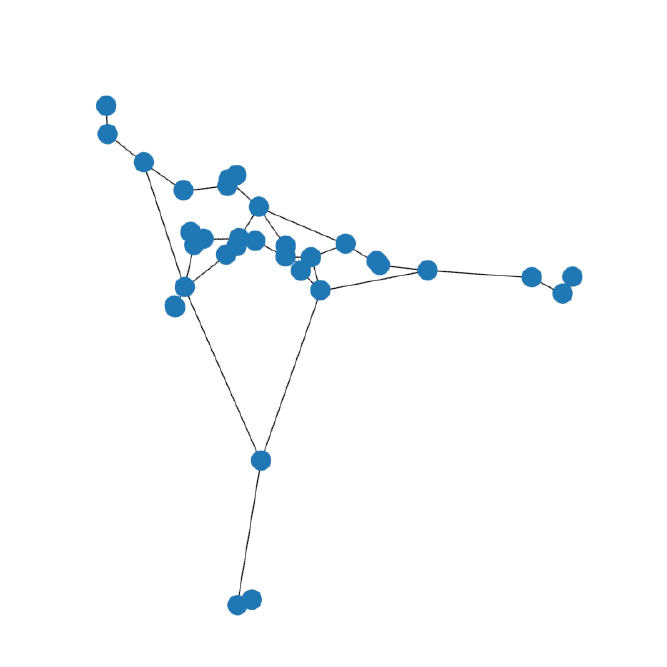}
         \includegraphics[width=0.095\paperwidth, trim= 10 0 25 0, clip]{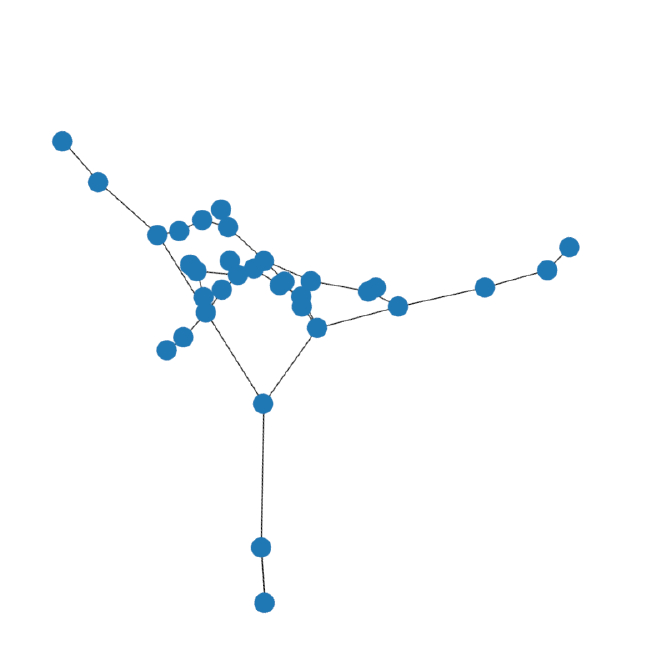}
         \includegraphics[width=0.095\paperwidth, trim= 20 0 20 0, clip]{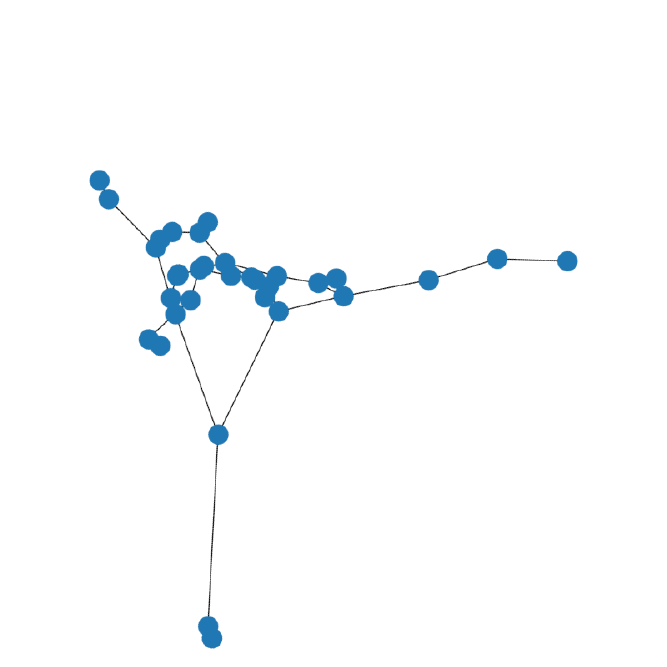} \hspace{0.2cm}
         \includegraphics[width=0.095\paperwidth, trim= 0 0 0 0, clip]{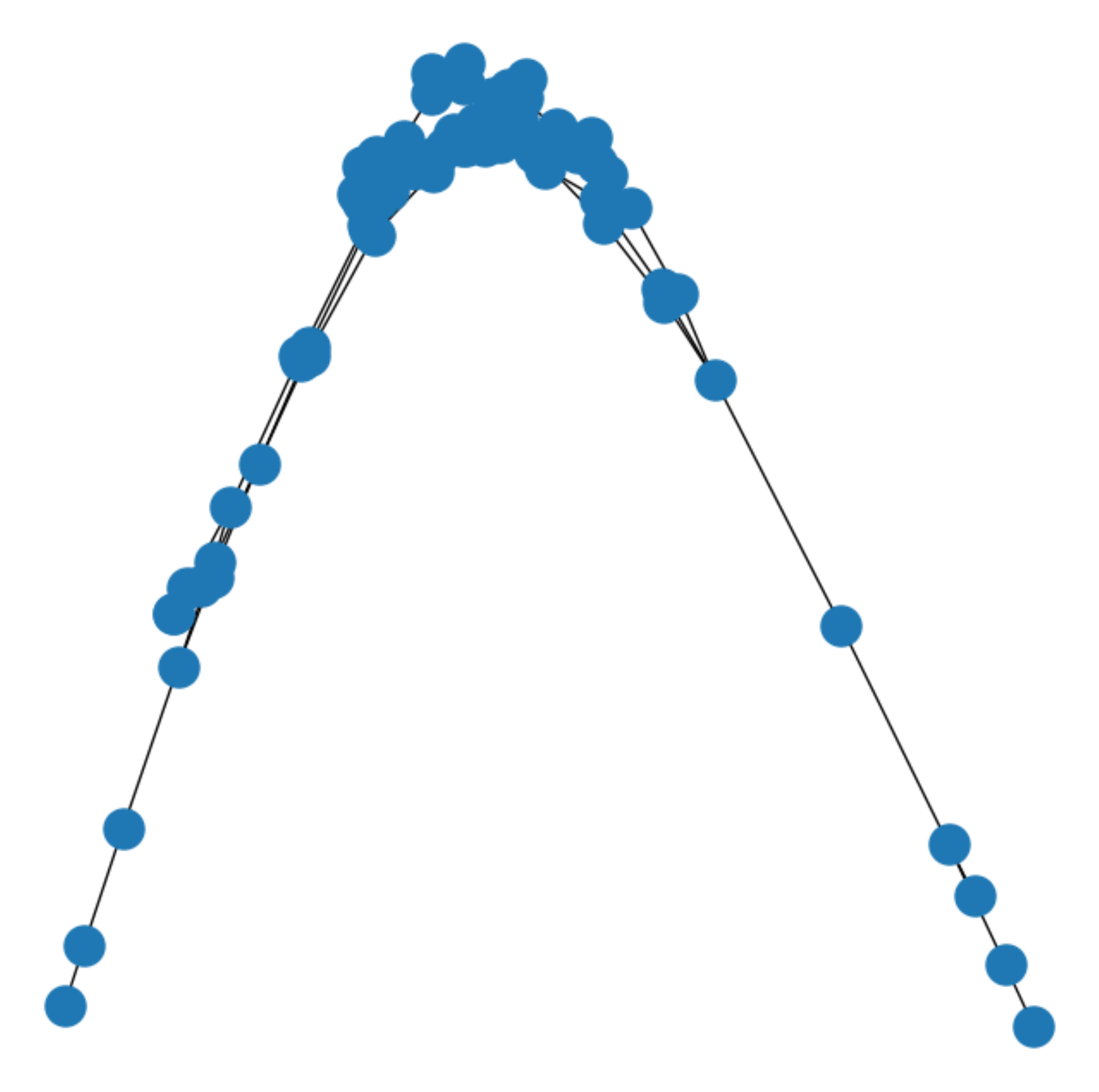}
         \includegraphics[width=0.095\paperwidth, trim= 20 30 20 0, clip]{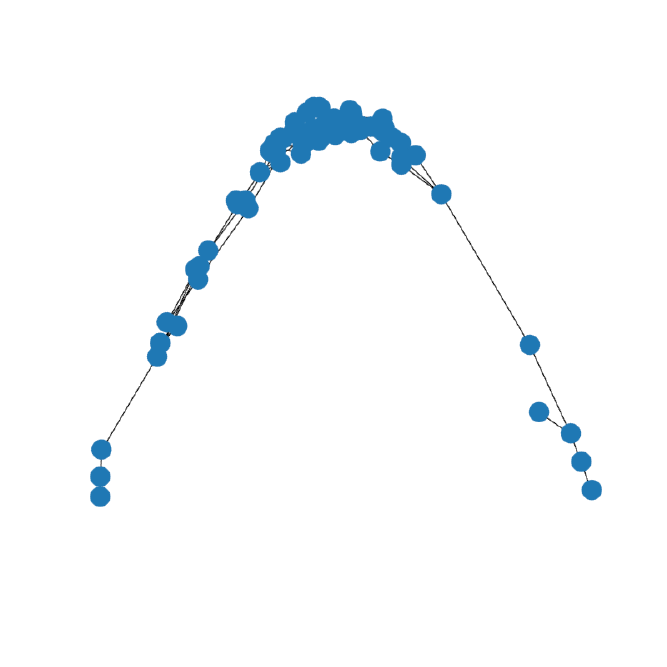}
         \includegraphics[width=0.095\paperwidth, trim= 20 30 10 0, clip]{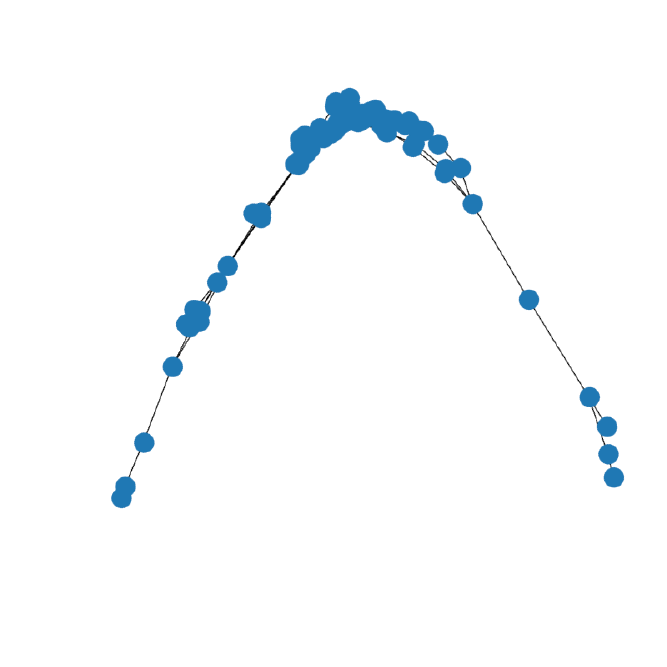}
         \includegraphics[width=0.095\paperwidth, trim= 10 30 20 0, clip]{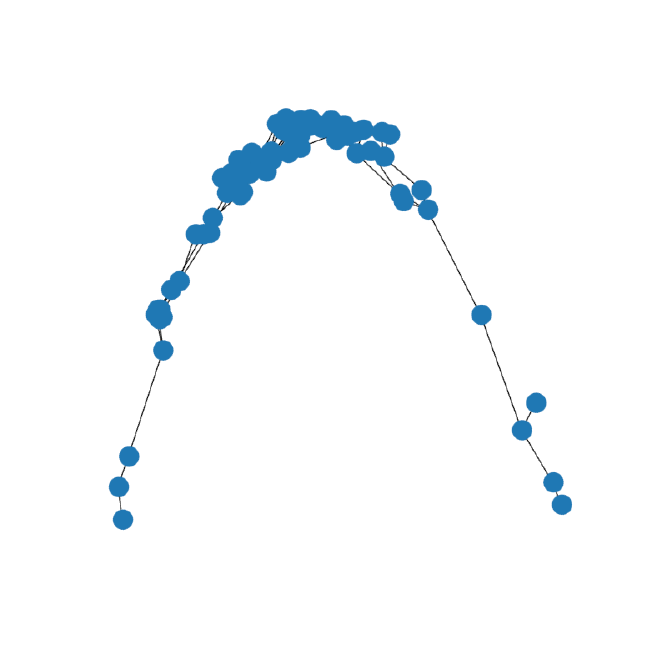}
         
            \caption{\textsc{Spectral layout}. \mt{Predicted node coordinates for the \textsc{Rome} dataset (first four column) and the \textsc{Sparse} dataset (subsequent four columns). Each row depicts the Ground-Truth positions (GT), the graph layout produced by GCN, GAT, GIN model, left-to-right. We report the predictions on three different test graphs (rows). }}

    \label{fig:sup_exp_spectral}
\end{figure*}

In Figures~\ref{fig:sup_exp_kamada} and \ref{fig:sup_exp_spectral}, we report a qualitative evaluation obtained by the best performing models for each different GNN architecture on three randomly picked graphs from the test set of each dataset. Figure~\ref{fig:sup_exp_kamada} shows the aforementioned evaluation in the case of the \textsc{Kamada-Kawai} layout supervision, both for the \textsc{Rome} dataset (first four columns, where the first one depicts the Ground Truth (GT)) and for the \textsc{Sparse} dataset. Figure~\ref{fig:sup_exp_spectral} shows the same analysis in the case of the \textsc{Spectral} layouts. 
The results show the good performances of the \mt{GND} framework in generating two heterogeneous styles of graph layouts, learning from different ground truth node coordinates. In order to give a more comprehensive analysis, we report in Table \ref{tab:sup_exp} a quantitative comparison among the global Procrustes Statistic similarity values  obtained on the test set by the best models, for both datasets.
We report the average score and its standard deviation over three runs with different seeds for the weights random number generator. 

\mt{The strength of the Laplacian PE is validated by the decent performances yielded by the MLP baseline. Conversely, the random features characterizing the rGNNs are not sufficient to solve this node regression task. Some additional structural information is required in order to jointly represent the node position and its sorroundings. Indeed, all the models exploiting the proposed solution outperform the competitors. The improved performances with respect to MLP are due to the fact that nodes states receive implicit feedback on their own position during the message passing steps.}
\mt{The proposed GAT model with Laplacian PE} achieves the best performances in all the settings. We believe that the attention mechanism plays a crucial role in the task of distinguishing the right propagation patterns, alongside the fact that the multi-head attention mechanism provides a bigger number of learnable parameters with respect to the competitors. 

In general, the \textsc{Spectral} layout is easier to be learned by the models. This can be due to the fact that the Laplacian PE represent an optimal feature for this task, given the common spectral approach. Even from a qualitative perspective generated layouts are almost identical to the Ground Truth.
Vice versa, the \textsc{Kamada-Kawai} layout represents an harder task to be learned from ground-truth positions, especially in the case of the \textsc{Sparse} dataset. As pointed out in \cite{dwivedi2020benchmarking}, Laplacian-based PE have still some limitations given by natural symmetries such as the arbitrary sign of eigenvectors, that several recent works are trying to solve \cite{cui2021positional}.  



\subsection{GNNs learn to draw minimizing aesthetic loss functions}
\label{sec:stress}

In Section \ref{sec:super}, GNDs explicitly minimize the distances with respect to certain ground-truth node positions, hence learning to draw directly from data according to certain layouts. In this second experimental setting, instead, we want to build GNNs capable to draw at inference time respecting certain aesthetic criteria which are implicitly learnt during training. 
We defined our framework in such a way that powerful PE features are mapped to 2D coordinates. Given a smooth and differentiable loss function defined on such output, we can leverage the BP algorithm in order to learn to minimize heterogeneous criteria. We investigate the case in which the GNN models minimize the \textit{Stress function} (see Eq. \ref{eq:stress}) on the predicted node coordinates. 
Only during the training phase, for each graph, we compute the shortest-path $d_{ij}$ among every node couple $(i,j)$. At inference time, the GND framework process the graph topology (the adjacency matrix) and the node features, directly predicting the node coordinates, without the need of any further information.

We use the same experimental setup, \mt{competitors} and hyper-parameters selection grids of Section \ref{sec:super}. However, according to a preliminary run of the models which achieved poor performances, we varied the hidden state dimension grid to $\{ 100, 200, 300\}$. This means that this task need a bigger representational capability with respect to the previous one, which is coherent with the complex implicit nature of the learning problem.
We set the Stress normalization factor to $w_{ij}= \frac{1}{d_{ij}}$ (hence, $\alpha=1$) and compute the averaged Stress function\footnote{We used the following \textit{average Stress} definition to avoid potential numerical issues: 

$\textsc{stress}(P) = \frac{1}{D}\sum_{i < j} w_{ij}\big(||p_i -p_j|| - d_{ij}\big)^2 $

where $D$ is the number of considered node couples.}. For this experiment, we use the stress value obtained on the validation split as the metric to select the best performing model. \mt{ For comparison,  we report the stress loss values obtained by three State-of-the-art Graph Drawing methods. Neato \footnote{Implementation available through Graphviz, \url{https://graphviz.com}} leverage the stress majorization \cite{gansner2004graph} algorithm to effectively minimize the stress. PivotMDS \cite{brandes2006eigensolver} is a deterministic dimension reduction based approach. Finally, ForceAtlas2 \cite{jacomy2014forceatlas2} generates graph layouts through a force-directed method.  
}

\begin{table}[th!]

\centering
 \caption{Average Stress loss value obtained on the Training set and Test set by the best selected models, for each dataset. We report the mean and standard deviation obtained over three runs initialized with different fixed seeds. \mt{We do not report standard deviations for Neato and PivotMDS, being deterministic algorithms.}}

\resizebox{\columnwidth}{!}{%
    \begin{tabular}{@{}p{.2cm}p{1.cm}p{1.45cm}p{1.6cm}p{1.45cm}p{1.6cm}@{}}\toprule 
    & \multirow{2}{*}{Model} & \multicolumn{2}{c}{\textsc{Rome}} & \multicolumn{2}{c} {\textsc{Sparse}} \\  
    \cmidrule(l){3-4}   \cmidrule(l){5-6}   
    & &  \textsc{train loss}  & \textsc{test loss}   &  \textsc{train loss}  & \textsc{test loss}  \\
     \midrule
     


    \multirow{3}{*}{\rotatebox{90}{\textsc{GD}}}& \mt{ForceAtl.2}      & 27.44 \scriptsize{$\pm$  0.01}        & 26.82 \scriptsize{$\pm$ 0.02}              & 23.31 \scriptsize{$\pm$ 0.01}            & 22.69 \scriptsize{$\pm$ 0.02} \\
    & \mt{Neato}    & 4.35                                  & 4.34                                       & 5.61                                     & 5.66 \\
     & \mt{Piv.MDS}     & 16.40                                 & 16.65                                      & 28.93                                    & 29.27 \\
    \midrule
  \multirow{4}{*}{\rotatebox{90}{\textsc{Compet.}}} & \mt{MLP}      & 1.07 \scriptsize{$\pm$ 0.01}          & 1.06 \scriptsize{$\pm$ 0.03}               & 1.16 \scriptsize{$\pm$ 0.02}             & 1.18 \scriptsize{$\pm$ 0.00} \\
  &  \mt{rGCN}     & 1.25 \scriptsize{$\pm$ 0.01}          & 1.24 \scriptsize{$\pm$ 0.04}               & 1.70 \scriptsize{$\pm 0.02$}             &  1.72 \scriptsize{$\pm 0.01$}  \\ 
   & \mt{rGAT}     &  0.98 \scriptsize{$\pm$ 0.02}    &  0.97 \scriptsize{$\pm$  0.00}  &  1.34 \scriptsize{$\pm$  0.01} &   1.36 \scriptsize{$\pm$  0.01} \\ 
    &\mt{rGIN}     & 1.11 \scriptsize{$\pm$ 0.03}          & 1.49 \scriptsize{$\pm$ 0.04}               & 1.49 \scriptsize{$\pm 0.07$}             &  1.89 \scriptsize{$\pm$ 0.02} \\ 
    \midrule
   \multirow{3}{*}{\rotatebox{90}{\textsc{\textbf{GND}}}}& GCN     & 0.51 \scriptsize{$\pm$ 0.02}          & 0.53 \scriptsize{$\pm$ 0.03}               & 0.56 \scriptsize{$\pm 0.05$}             &  0.61 \scriptsize{$\pm 0.02$}  \\ 
   & GAT      & {\bf 0.33} \scriptsize{$\pm$ 0.02}    & {\bf 0.34} \scriptsize{$\pm$ {\bf 0.00}}   & {\bf 0.30} \scriptsize{$\pm$ {\bf 0.02}} &  {\bf 0.33} \scriptsize{$\pm$ {\bf 0.01}} \\ 
   & GIN      & 0.49 \scriptsize{$\pm$ 0.04}          & 0.88 \scriptsize{$\pm$ 0.05}               & 0.28 \scriptsize{$\pm 0.05$}             &  0.81 \scriptsize{$\pm$ 0.01} \\ 

    \bottomrule
    \end{tabular}
}

 \label{tab:stress}
\end{table}

\begin{figure}[ht!]
\noindent\rule{\columnwidth}{0.4pt} \vspace{-0.85cm}\\

\hspace{3.85cm} \textsc{Rome} \vspace{-.2cm}\\
\noindent\rule{\columnwidth}{0.4pt} \vspace{-.6cm}\\

\hspace{1.4cm}  \textsc{GCN}  \hspace{1.1cm} $\quad $ \textsc{GAT} $\quad $   \hspace{1.3cm} \textsc{GIN}  \vspace{-.2cm}  \\

    \centering
         \includegraphics[width=0.25\columnwidth, trim=20 20 20 20, clip]{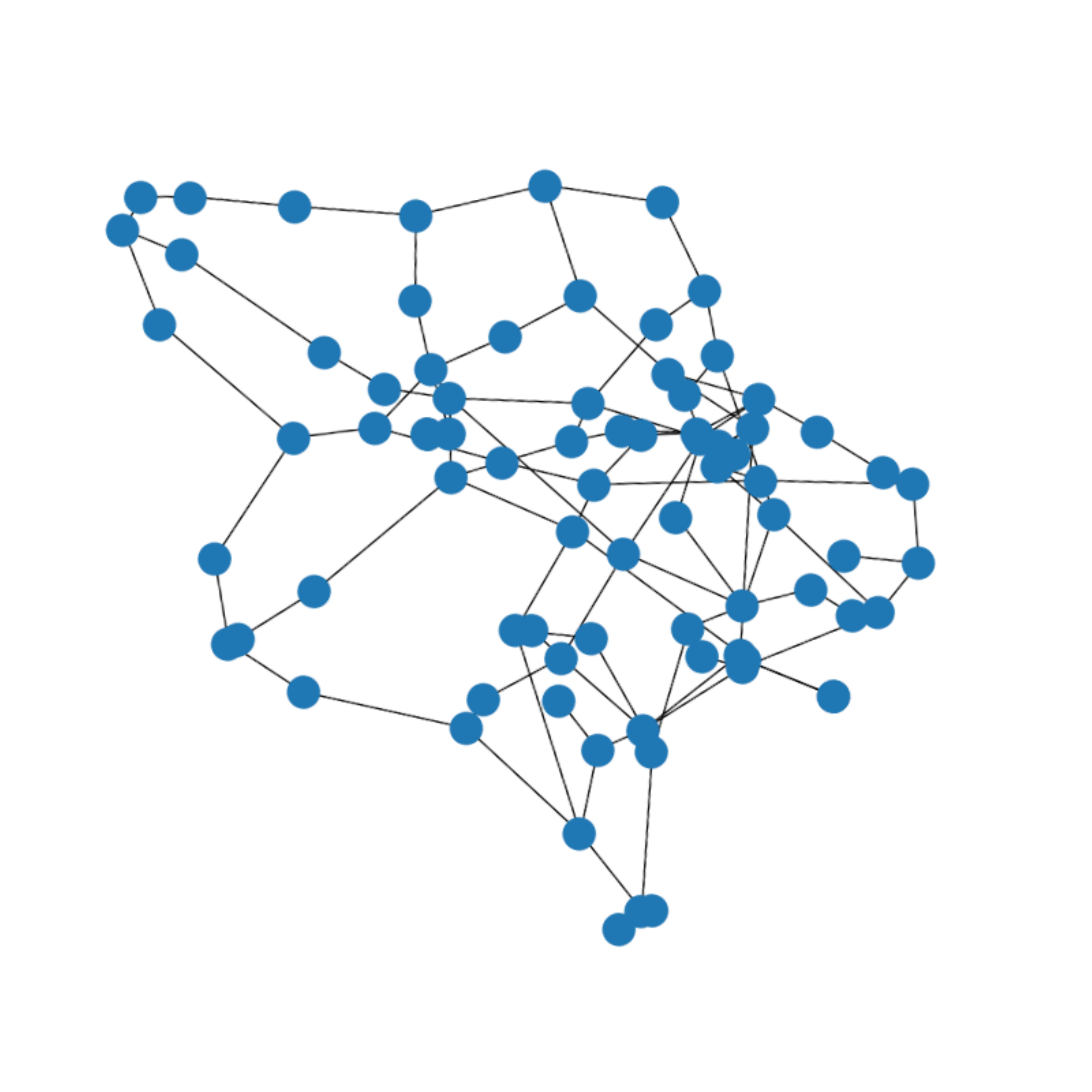}
         \includegraphics[width=0.25\columnwidth, trim=20 20 20 20, clip]{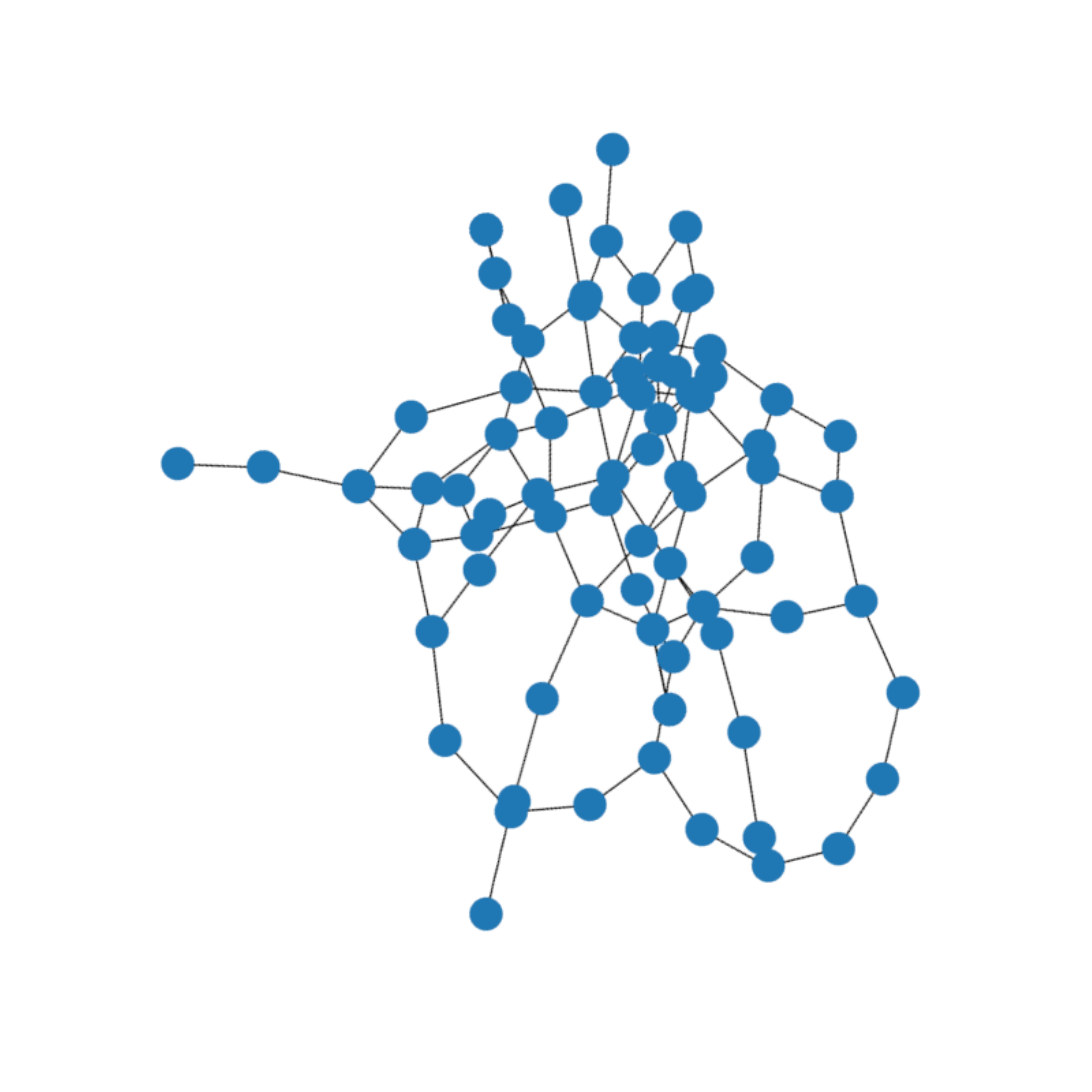}
         \includegraphics[width=0.25\columnwidth, trim=20 20 20 20, clip]{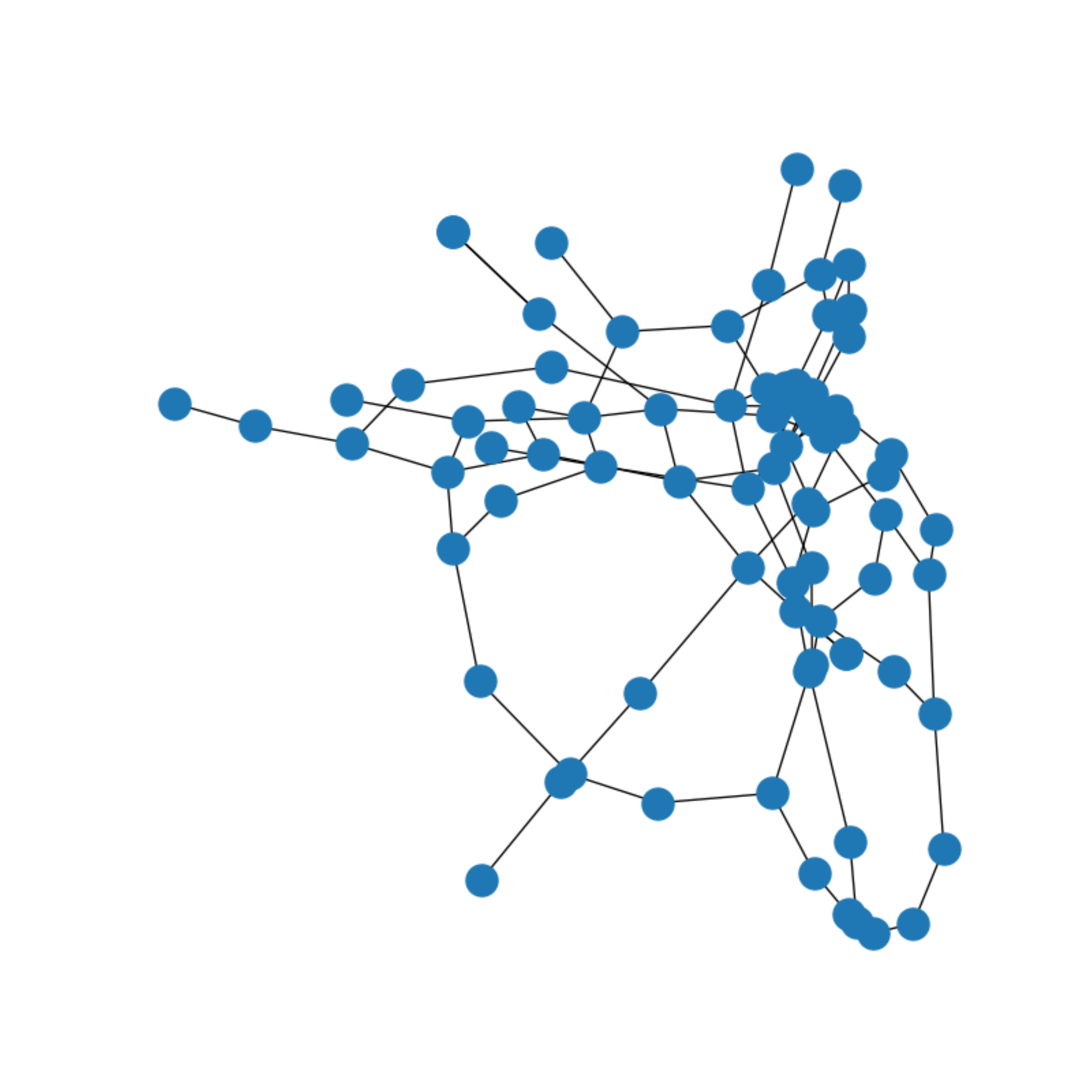}
         \includegraphics[width=0.25\columnwidth, trim=20 20 20 20, clip]{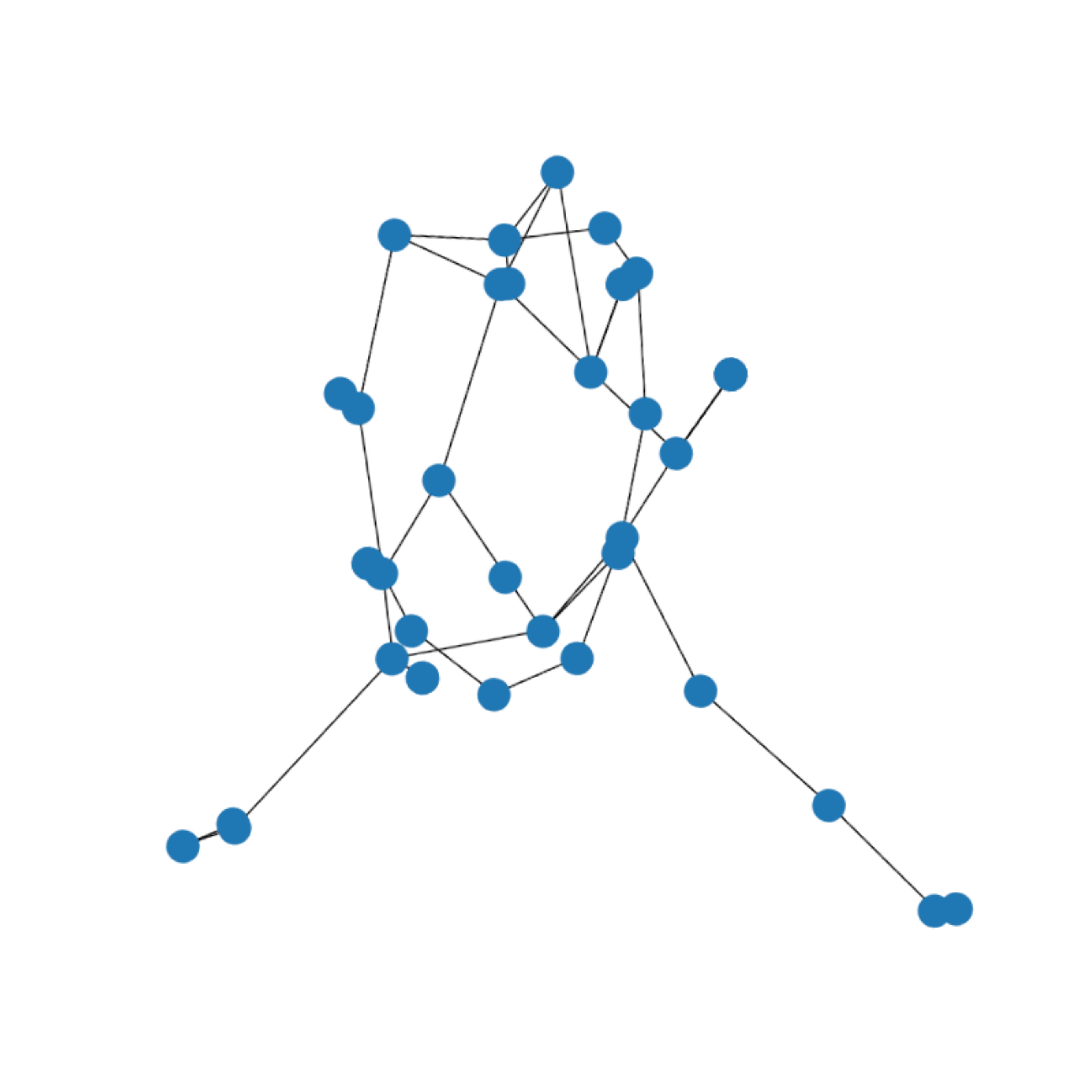}
         \includegraphics[width=0.25\columnwidth, trim=20 20 20 20, clip]{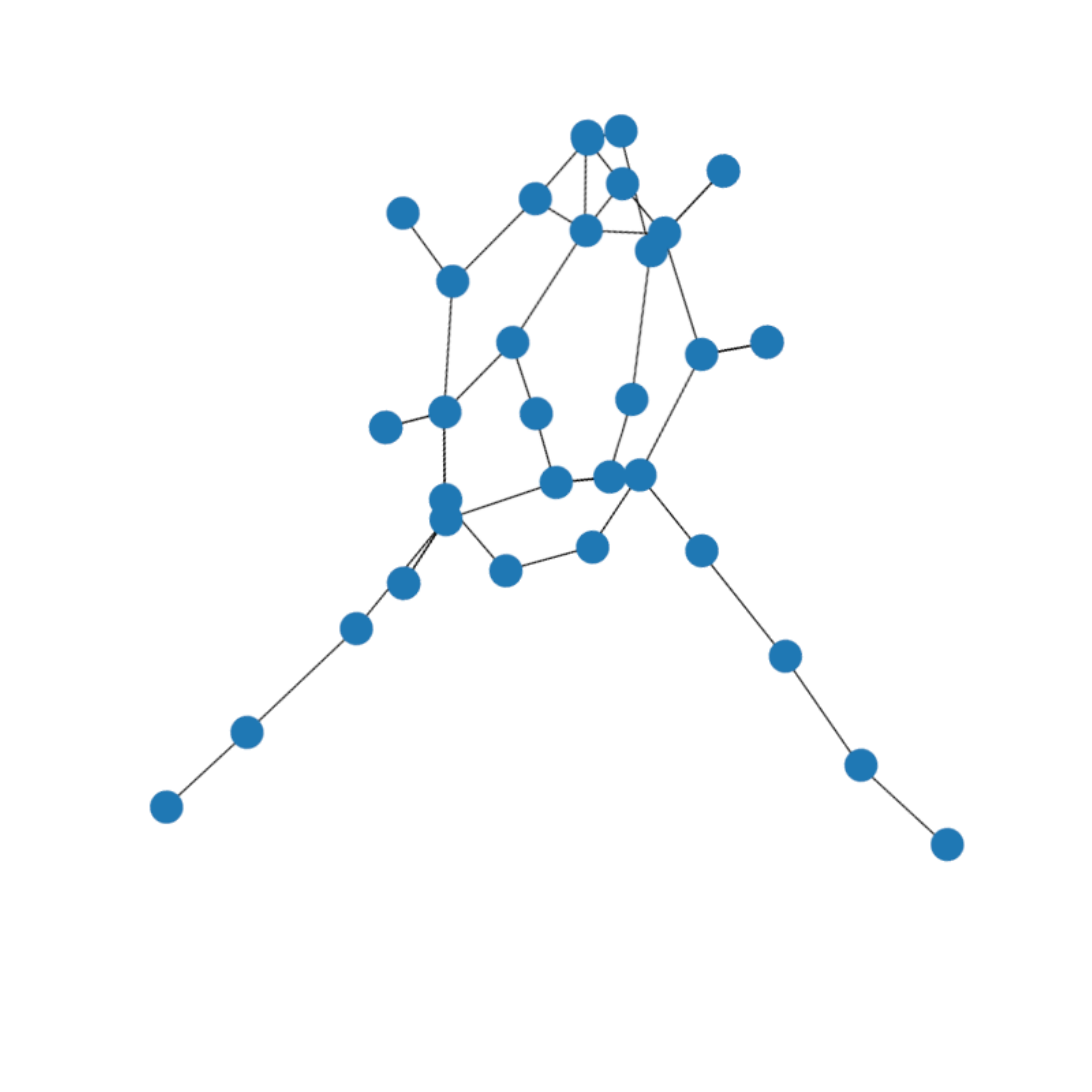}
         \includegraphics[width=0.25\columnwidth, trim=20 20 20 20, clip]{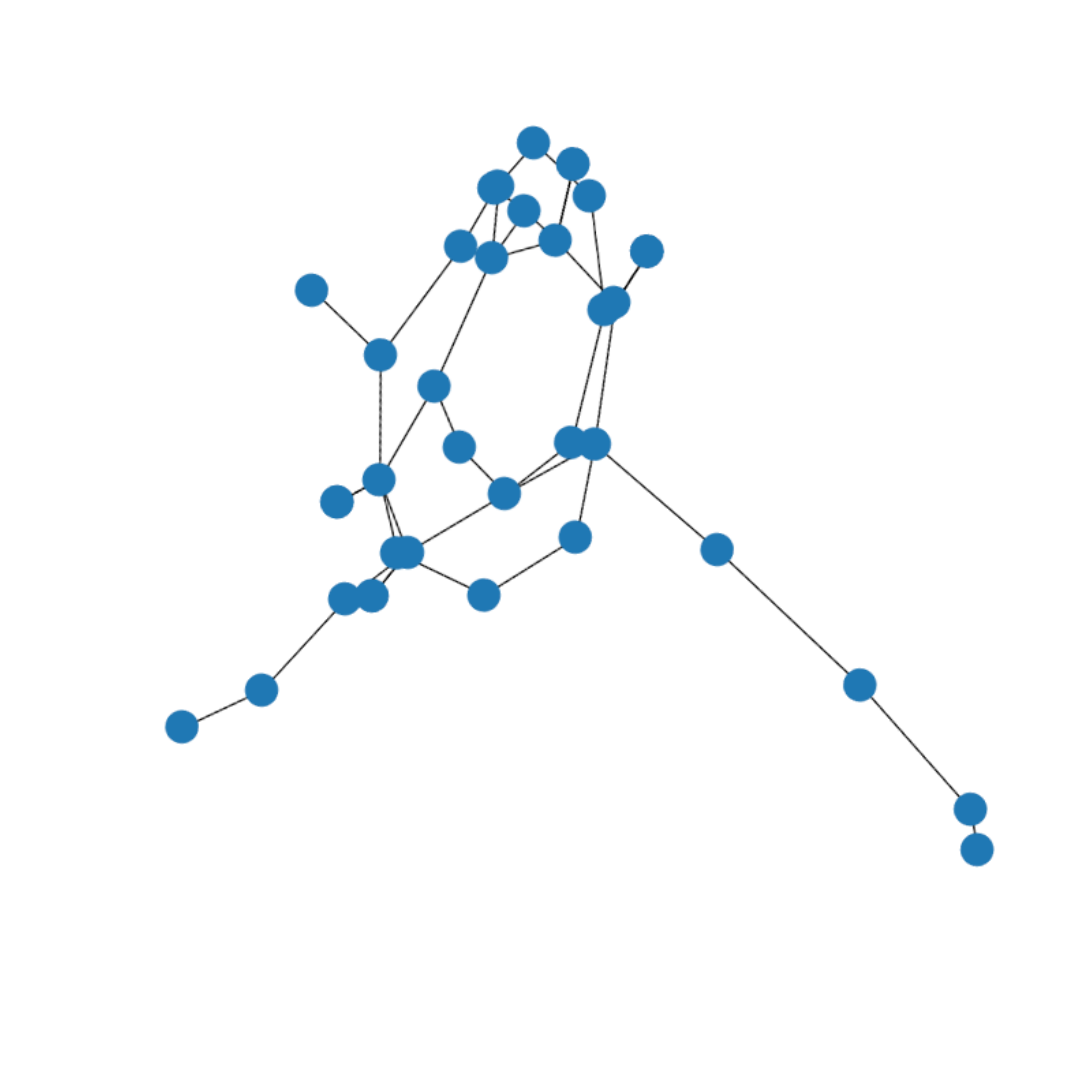}
         \includegraphics[width=0.25\columnwidth, trim=20 20 20 20, clip]{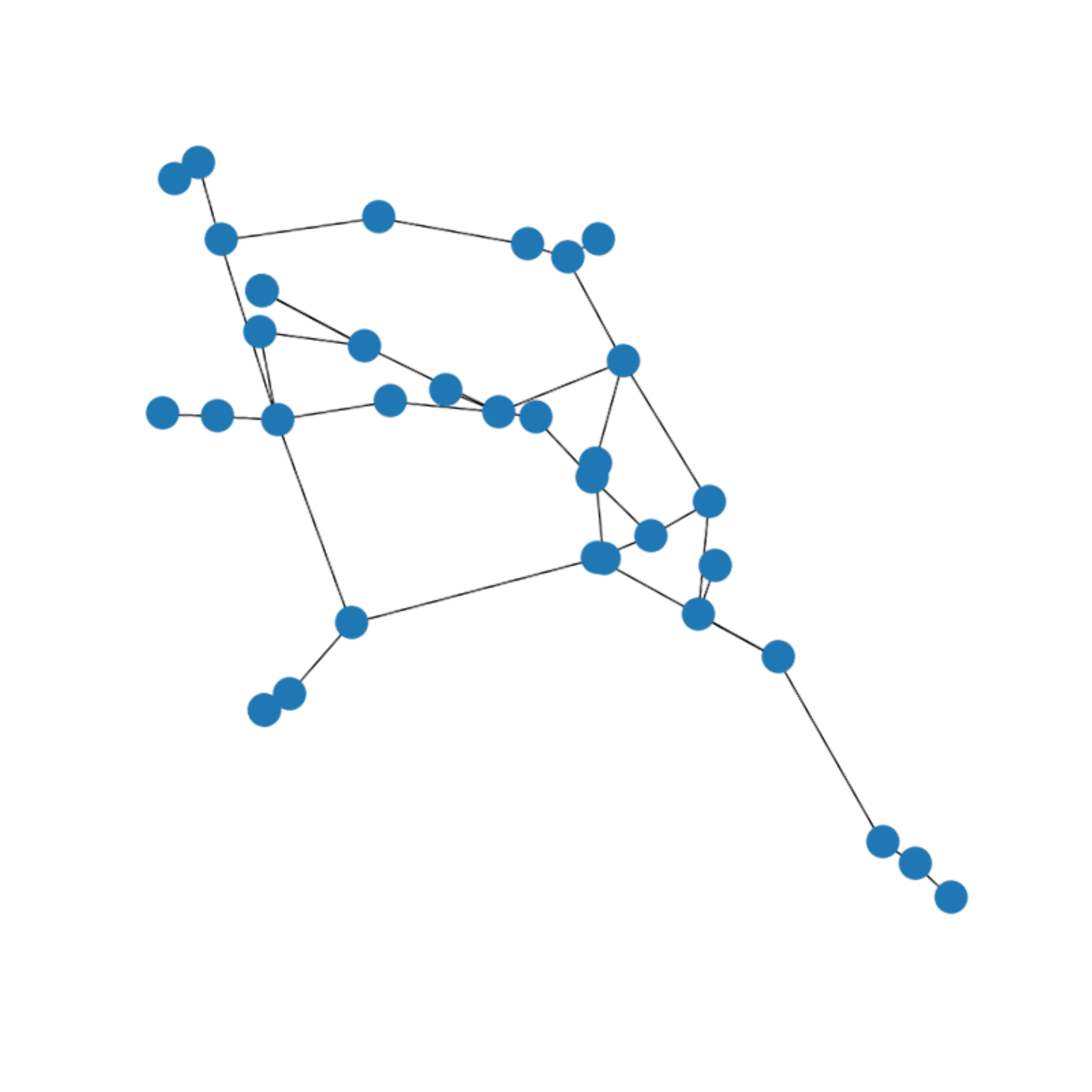}
         \includegraphics[width=0.25\columnwidth, trim=20 20 20 20, clip]{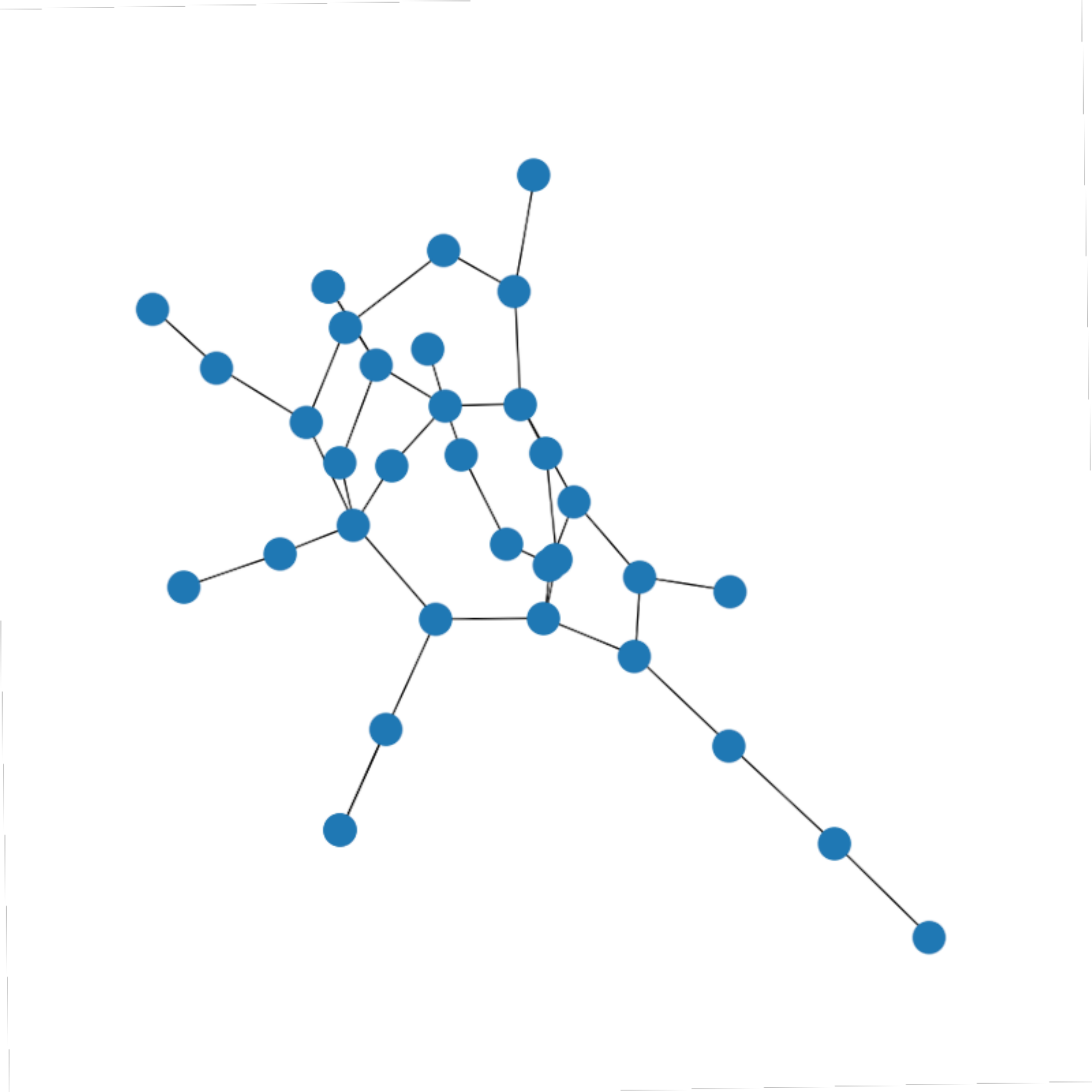}
         \includegraphics[width=0.25\columnwidth, trim=20 20 20 20, clip]{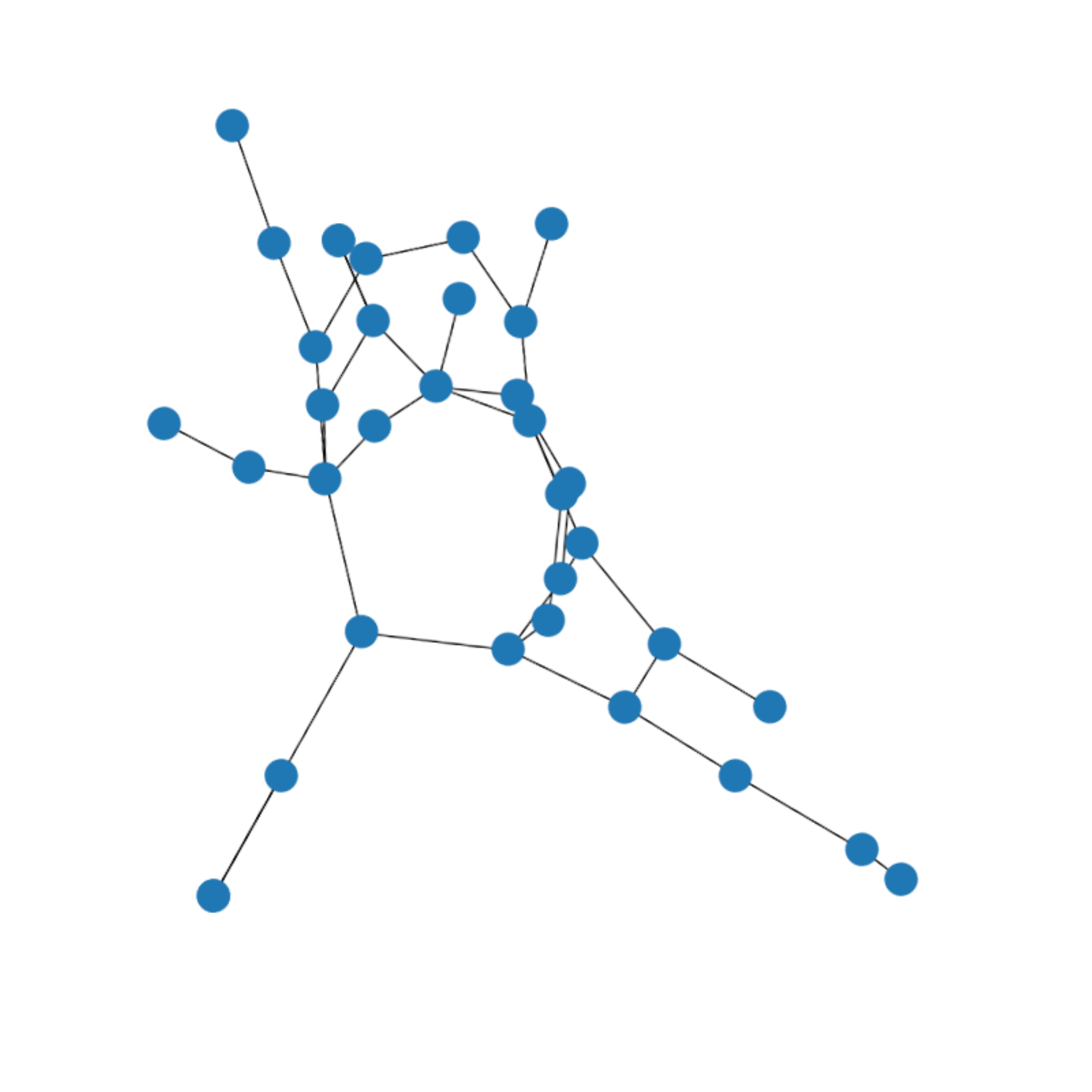}


    \caption{\textsc{Stress minimization on Rome}. Qualitative example of the graph layout produced by three GNN models on the test graphs of the \textsc{Rome} dataset. Each row contains one of the same three graphs depicted in the first column of Figure \ref{fig:sup_exp_kamada} for comparison with the layout produced by Kamada-Kawai \cite{kamada1989algorithm}. }
    \label{fig:stress_rome}
\end{figure}

\begin{figure}[h]
\noindent\rule{\columnwidth}{0.4pt} \vspace{-0.85cm}\\

\hspace{3.8cm} \textsc{Sparse} \vspace{-.2cm}\\
\noindent\rule{\columnwidth}{0.4pt} \vspace{-.6cm}\\

\hspace{1.2cm}  \textsc{GCN}  \hspace{1.3cm} $\quad $ \textsc{GAT} $\quad $   \hspace{1.5cm} \textsc{GIN}    \vspace{-.3cm} \\

    \centering

         \includegraphics[width=0.25\columnwidth, trim = 20 20 20 20, clip]{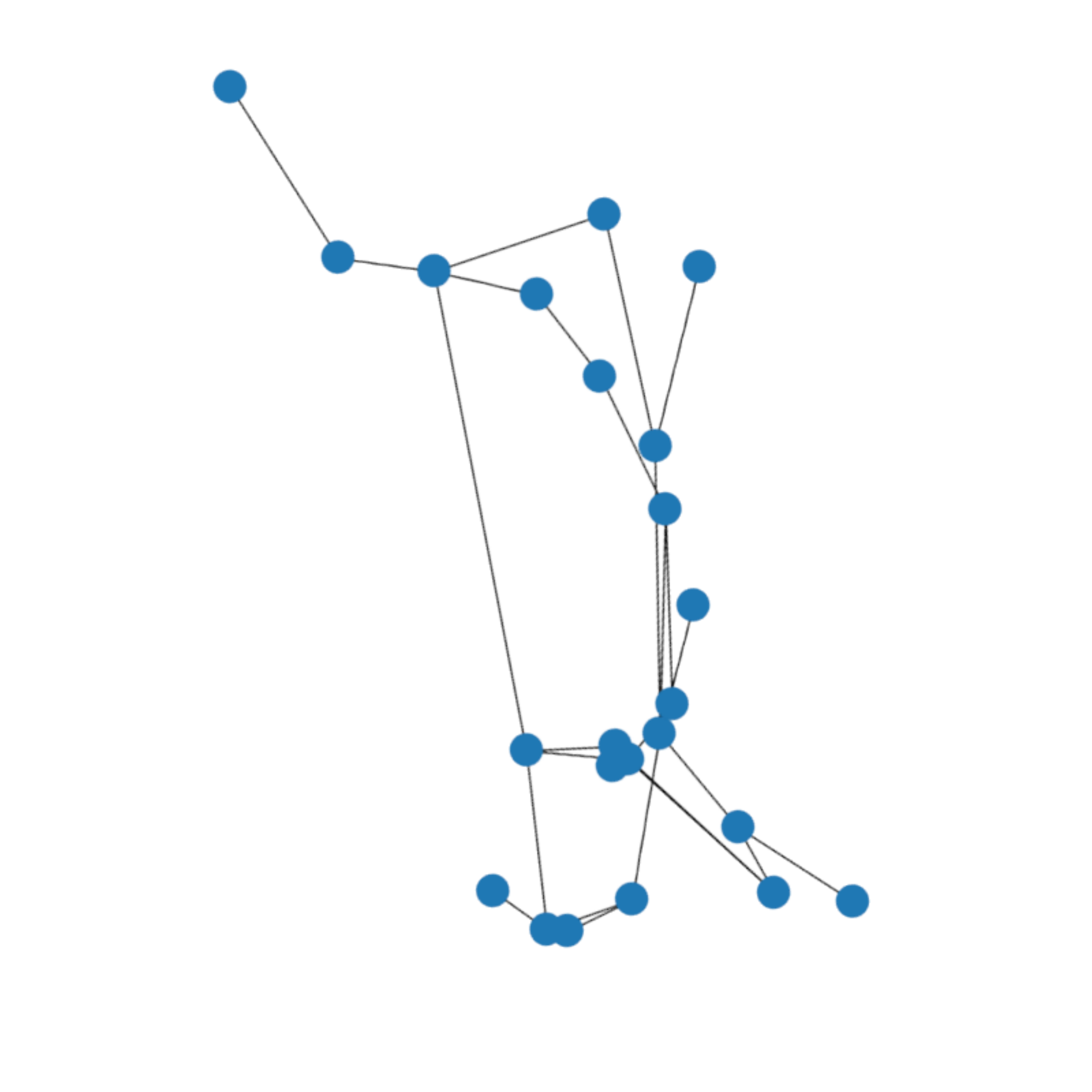}
         \includegraphics[width=0.25\columnwidth, trim = 20 20 20 20, clip]{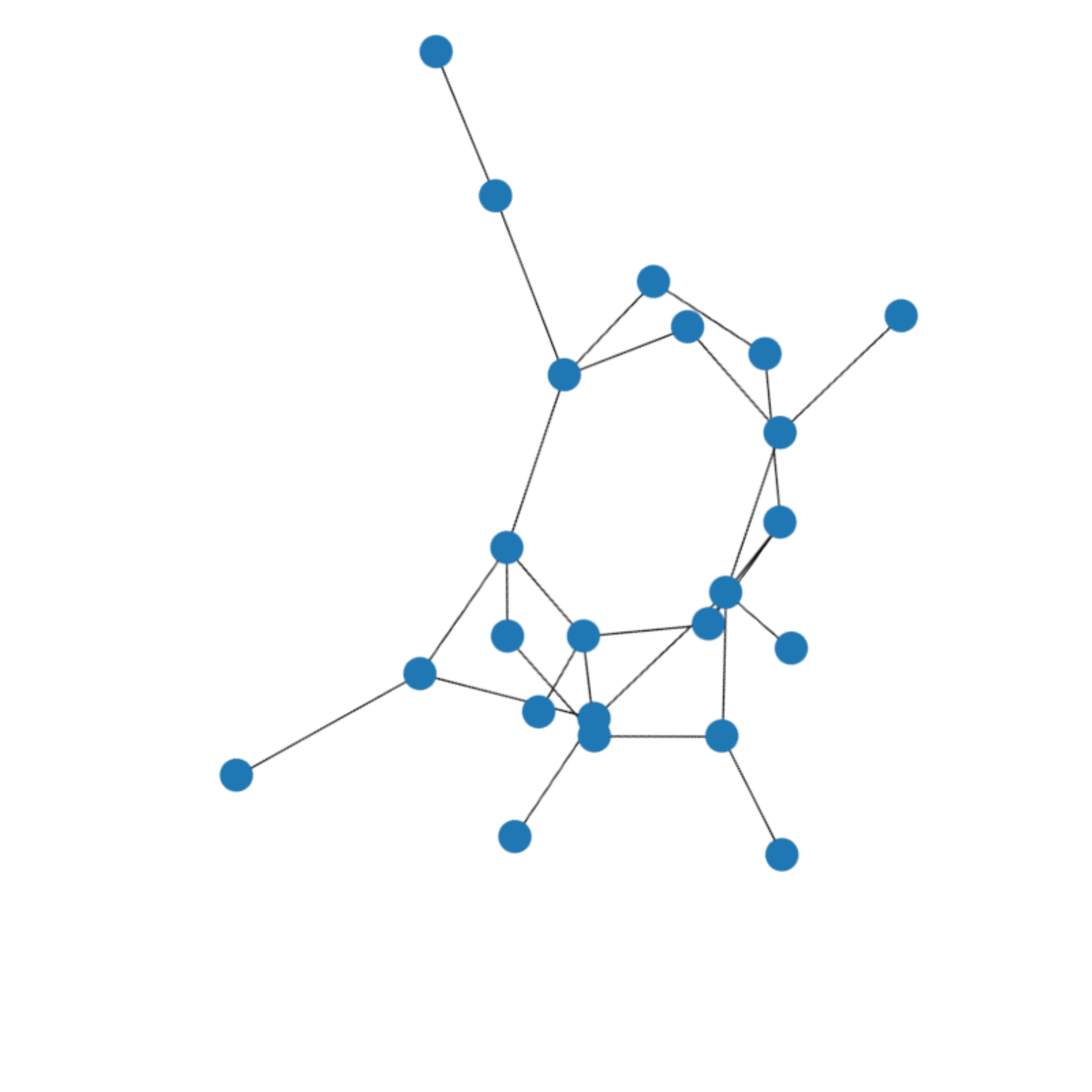}
         \includegraphics[width=0.25\columnwidth, trim = 20 20 20 20, clip]{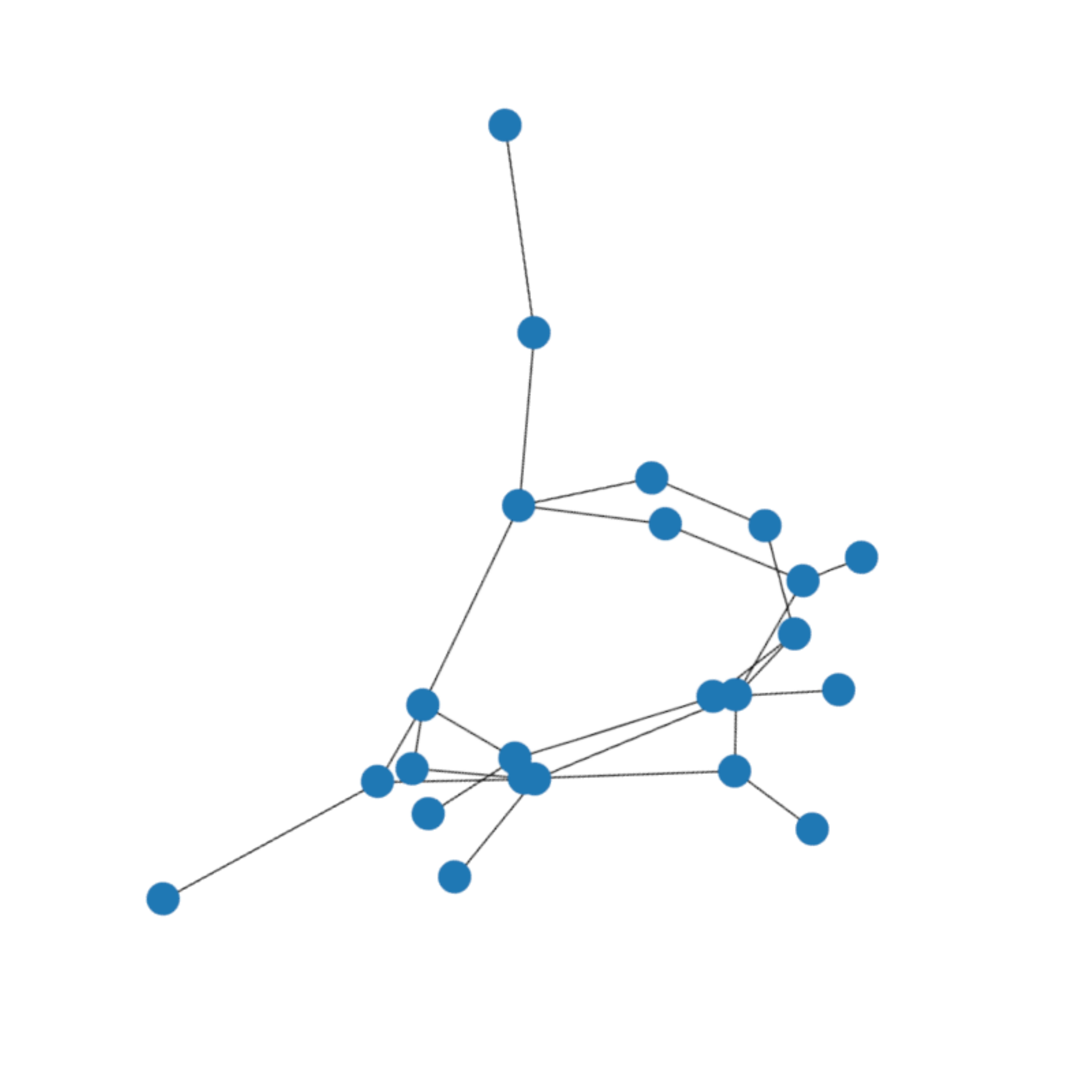}
         \includegraphics[width=0.25\columnwidth, trim = 20 20 20 20, clip]{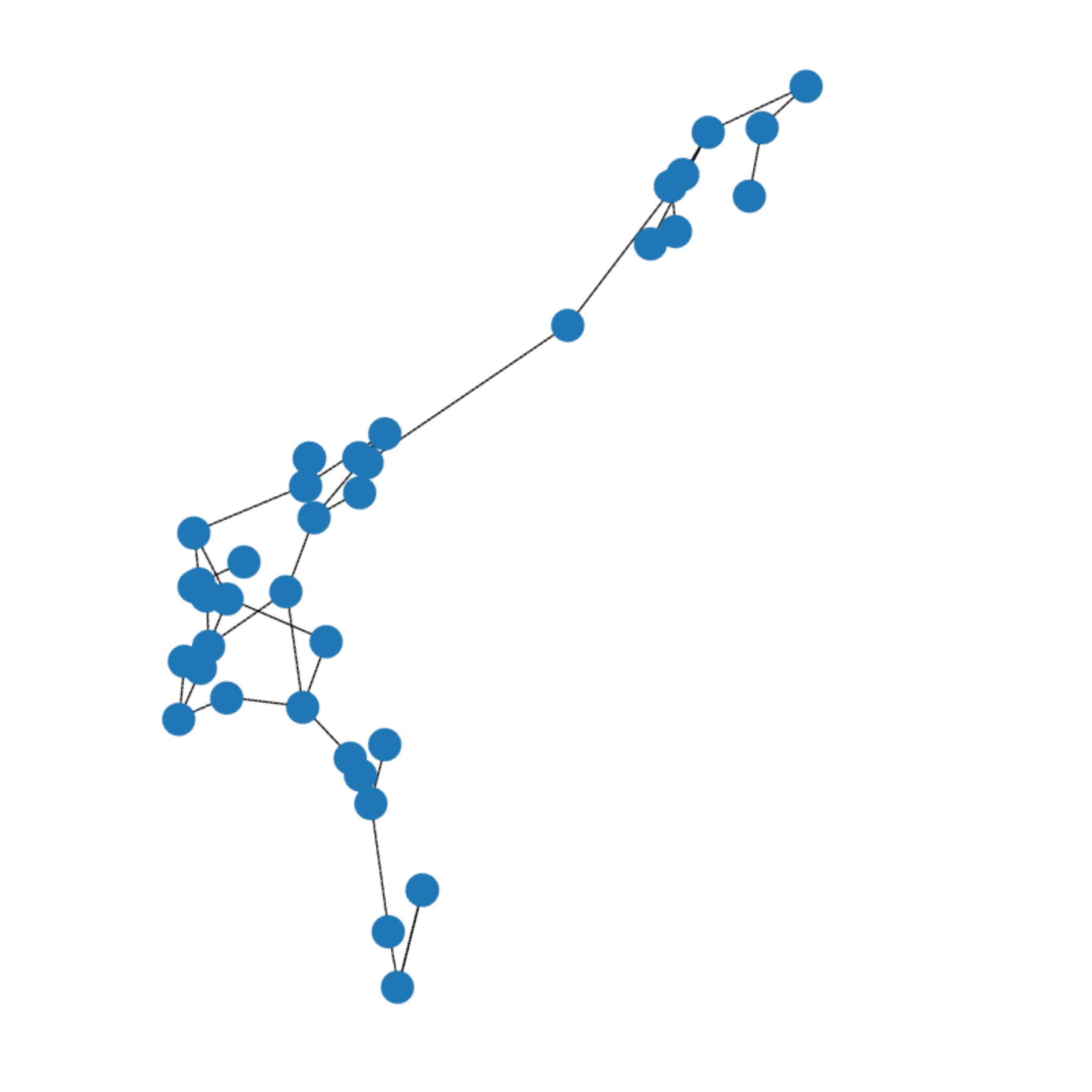}
         \includegraphics[width=0.25\columnwidth, trim = 20 20 20 20, clip]{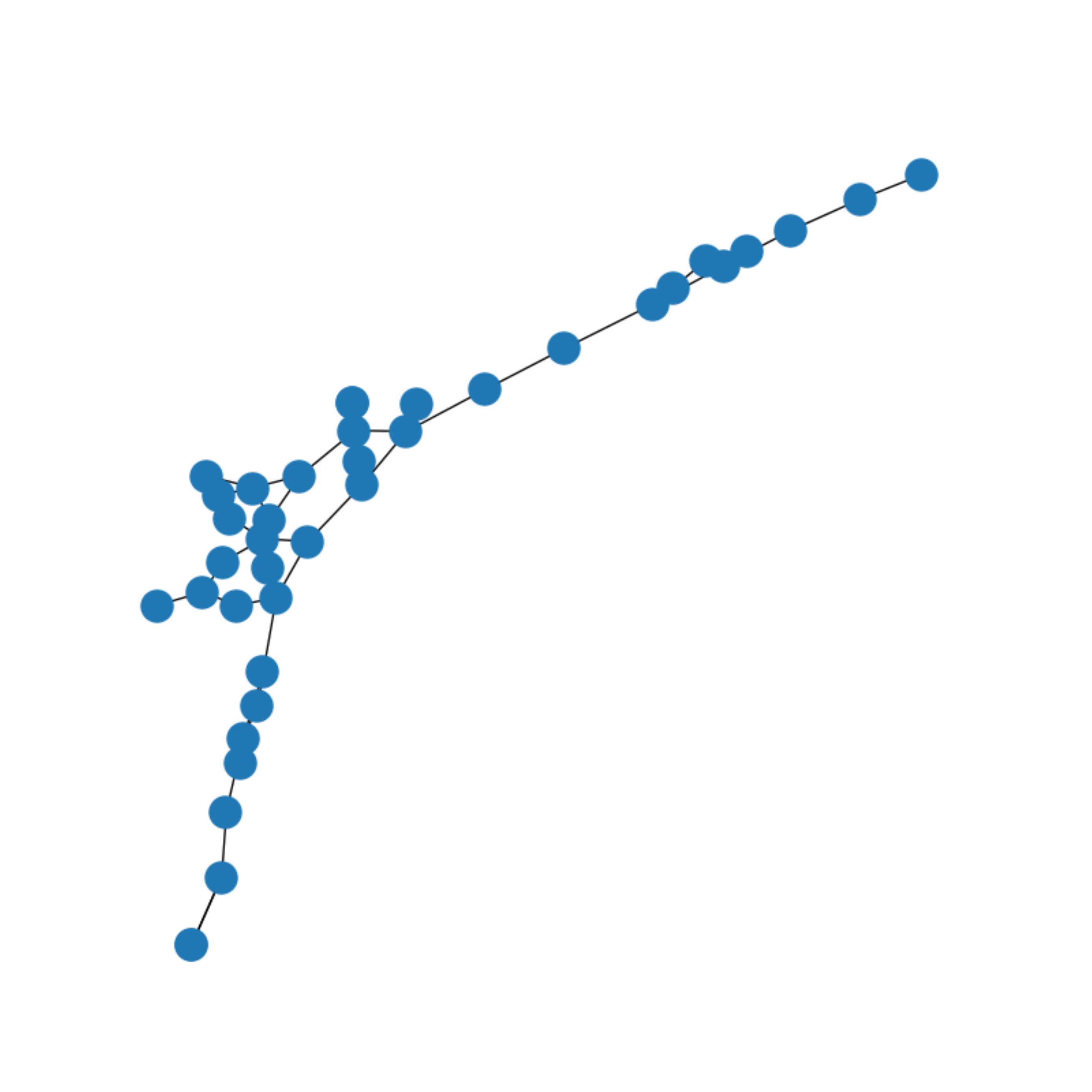}
         \includegraphics[width=0.25\columnwidth, trim = 20 20 20 20, clip]{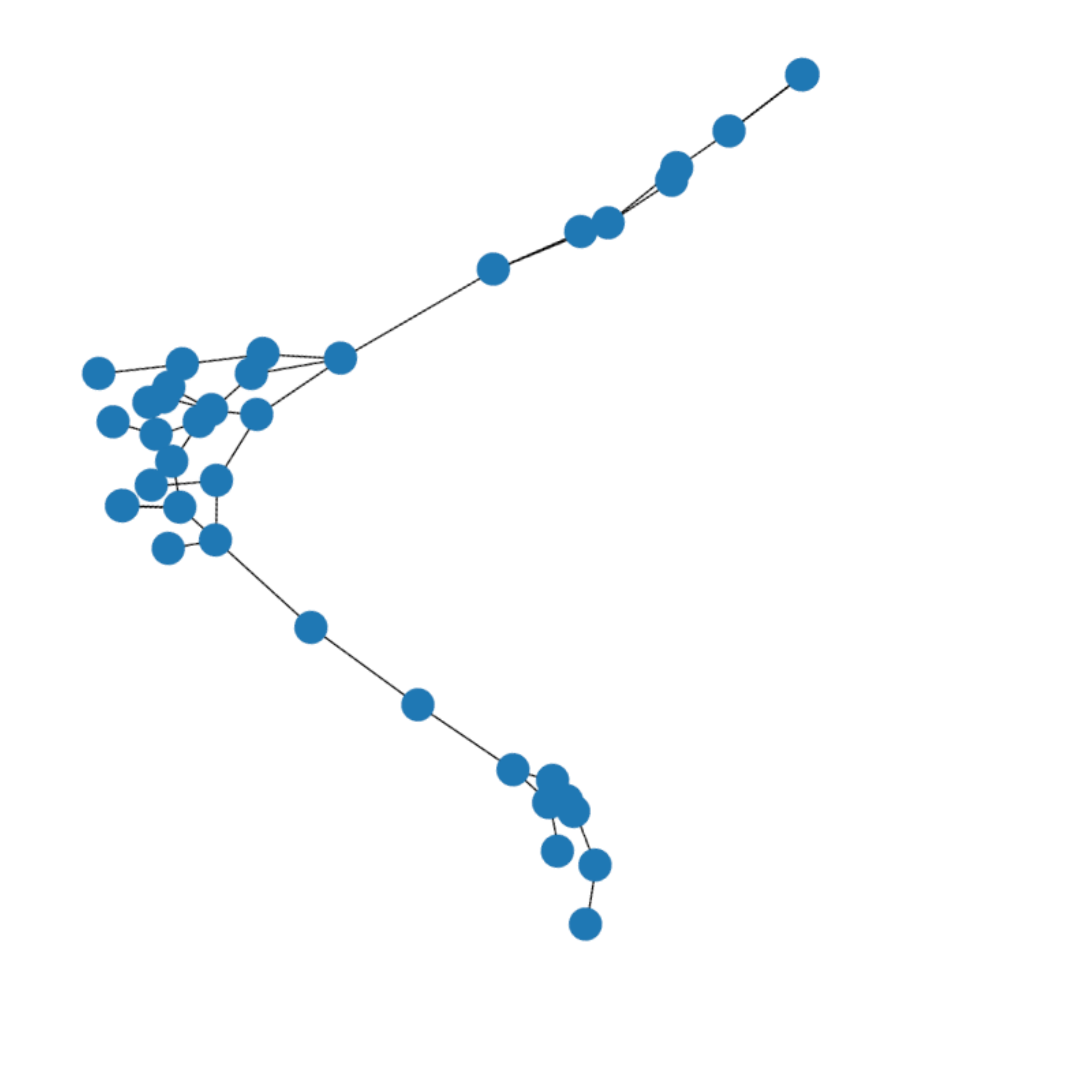}
         \includegraphics[width=0.25\columnwidth, trim = 20 20 20 20, clip]{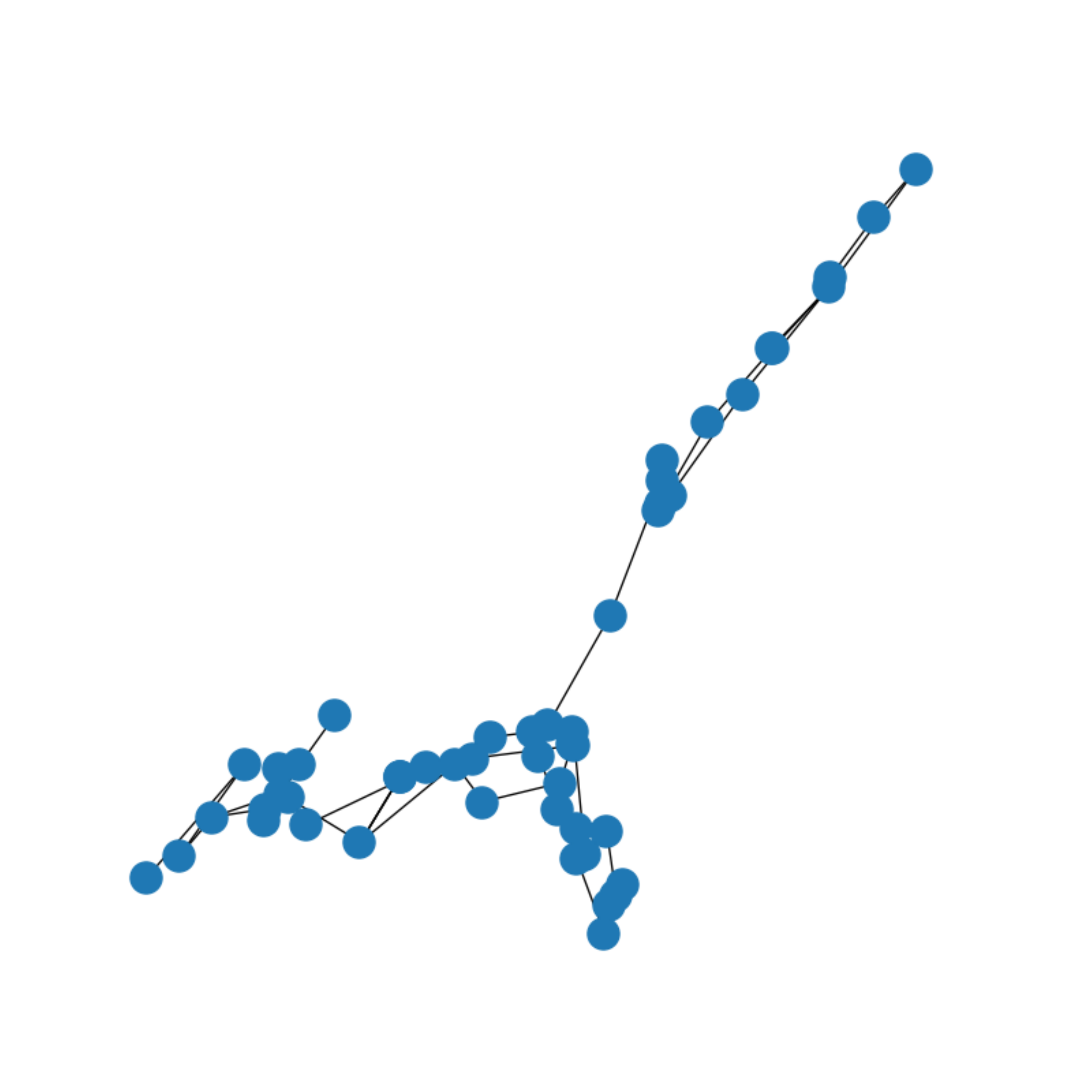}
         \includegraphics[width=0.25\columnwidth, trim = 20 20 20 20, clip]{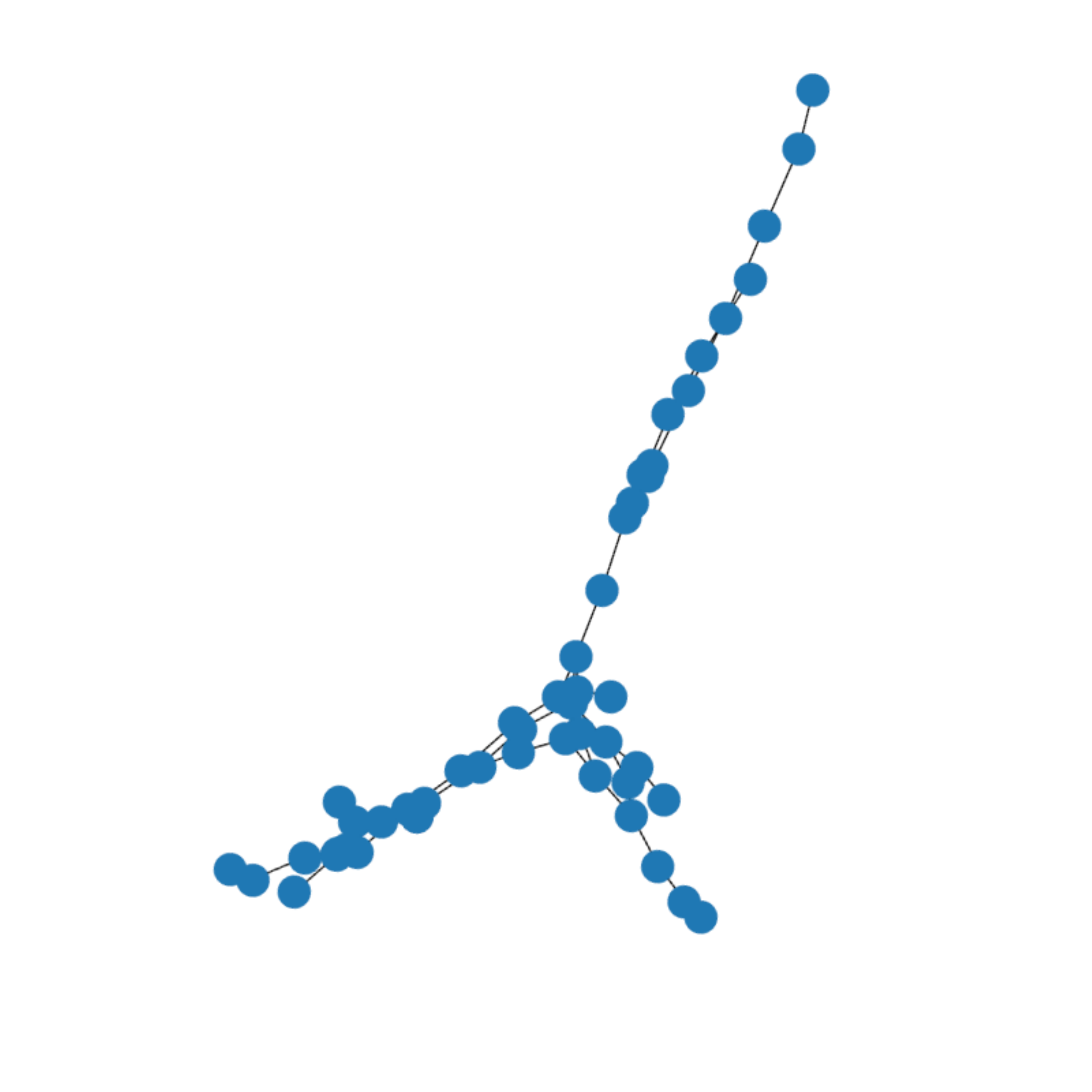}
         \includegraphics[width=0.25\columnwidth, trim = 20 20 20 20, clip]{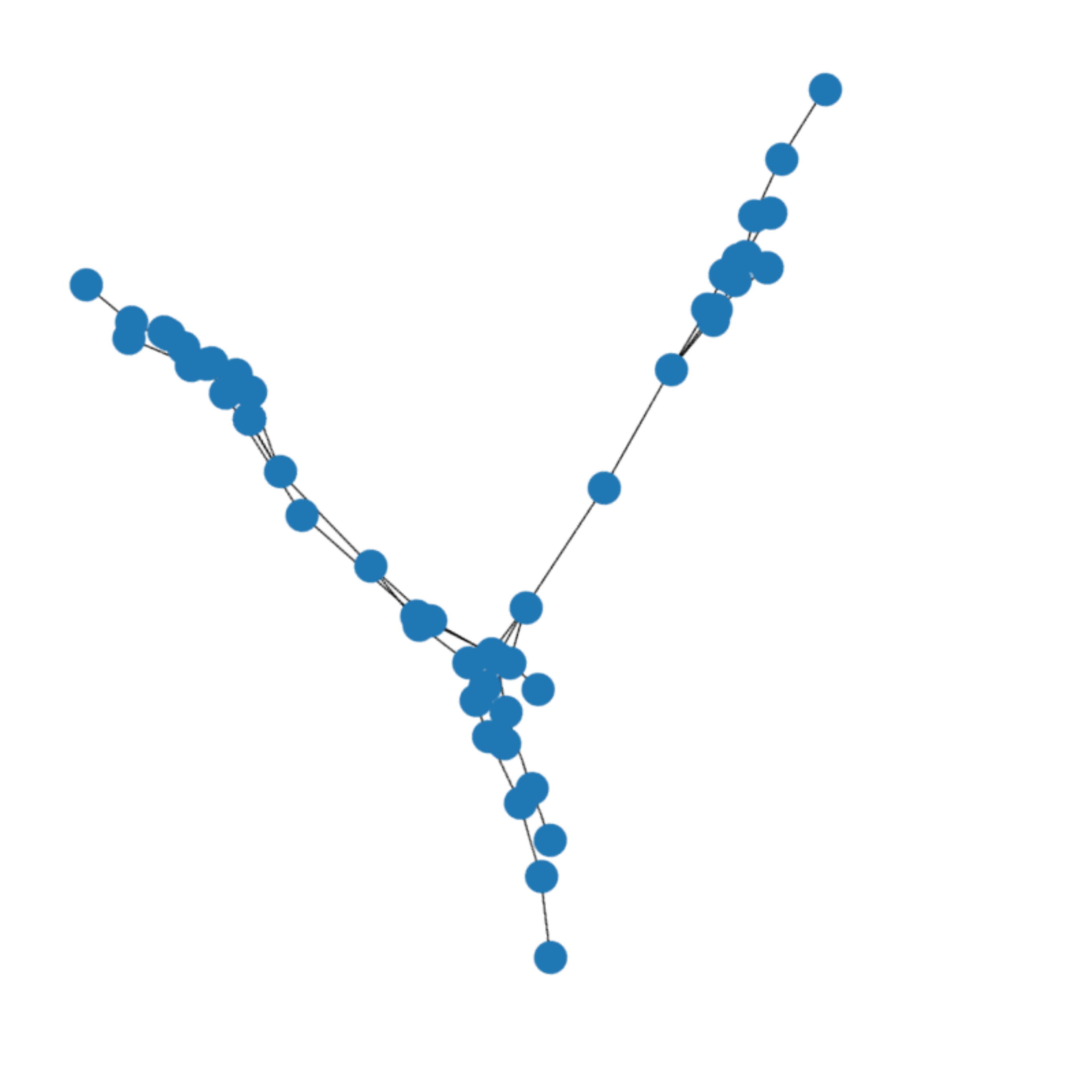}


         
    \caption{\textsc{Stress minimization on Sparse}. Same setting of Figure \ref{fig:stress_rome}.}
    \label{fig:stress_random}
\end{figure}

We report in Figure \ref{fig:stress_rome} and  \ref{fig:stress_random} some qualitative examples of the graph layouts produced by the best selected GNN models on test samples (the same graphs selected for Figure \ref{fig:sup_exp_kamada}) of the two datasets,  following the aforementioned setting. 
Noticeably, all three models succeed in producing a layout that adheres to the typical characteristics of graphs obtained via Stress minimization. \mt{In particular, for reference on the drawing style, the layouts of these same graphs generated via the Kamada-Kawai algorithm (that also minimize stress) are depicted in the first and fifth columns of Figure~\ref{fig:sup_exp_kamada}, for \textsc{Rome} and \textsc{Sparse} dataset respectively. Comparing the graph layout produced by the various GNN models and the aforementioned ones from Kamada-Kawai, also in this case it is easy to see from a qualitative analysis that the GAT model is the best performing one. }
The peculiar characteristics of the \textsc{Sparse} dataset (sparse connection patterns, causing many symmetries and isomorphic nodes) entail a hardship in minimizing the stress loss function in some of the reported examples.

A quantitative comparison is reported in Table \ref{tab:stress}, with the stress values obtained by the best models for each competitor and dataset, both at training time and test time, averaged over three runs initialized with different seeds. Once again, GAT performs the best. The metrics obtained by the GIN model highlight an overfit of the training data, given the selected grid parameters. \mt{GND models obtain better stress than all the SOTA Graph Drawing packages, with Neato being the best performing one in terms of stress minimization, as expected. Similar conclusion with respect to the previous experiment can be drawn regarding the results obtained by rGNNs and MLP. Indeed, these result show how learning to minimize stress requires both positional and structural knowledge, and that the message passing process foster the discriminative capability of the learned node states, with respect to solely exploiting local information. }

Summing up, the experimental campaign showed the generalization capabilities of the proposed framework even in the task of minimizing common aesthetic criteria imposed on the GNNs node-wise predictions, such as the \textit{stress function}, on unseen graphs. The GND framework is capable to predict node positions on unseen graphs respecting typical stress minimization layouts,  without providing at inference time any explicit graph-theoretic/shortest-path information.

\subsection{GNNs learn to Draw from Neural Aesthetes}
\label{sec:neu}

In the previous Section, we showed that GNDs are capable to learn to minimize a differentiable smooth function that implicitly guides the node coordinates positioning.
In a similar way, the Neural Aesthetes presented in Section \ref{sec:proposal_1} provide a smooth differentiable function that can be leveraged to find a good gradient descent direction for the learning parameters.
In this Section, we mix the two proposals in order to build a Graph Neural Drawer that learns to generate graph layouts thanks to the gradients provided by the edge-crossing Neural Aesthete, and, eventually, to optimize the combination of several aesthetic losses.

At each learning epoch, GND minimizes the loss function $H(P)$ defined in Eq. \ref{eq:total_loss_neu},  over the whole edge list $\mathcal{E}$. 
The loss function can be computed as follows: the GNN model process the graph and predicts node-wise coordinates. Given such predicted node positions and the input graph adjacency matrix, the Neural Aesthete (which was trained beforehand as explained in Section \ref{sec:prop1_ex}) processes couples of arcs and output their degree-of-intersection. The overall loss function can then be composed by the contribution given by each of the considered arc-couples, as in Eq. \ref{eq:dir_no_target}.


We restrict our analysis \mt{to the Rome dataset, exploiting } a GAT model with 2 hidden layers, an hidden size of node state of 25, PE dimension $k=10$, learning rate $ \eta= 10^{-2}$. We compare the graph layout generated by this model in three randomly picked test graphs, comparing three different loss function definitions: (\textit{i}) stress loss, (\textit{ii}) Neural-Aesthete edge-crossing based loss $H(P)$, (\textit{iii}) a combination of the two losses  with a weighing factor  $\lambda=0.5$ acting on the Neural Aesthete loss, in particular:
\begin{equation}
    \textsc{Loss(P)} = \textsc{Stress}(P) + \lambda H(P) .
\end{equation}

\begin{figure}[h]
\noindent\rule{\columnwidth}{0.4pt} \vspace{-0.85cm}\\

\hspace{3.9cm} \textsc{Rome} \vspace{-.2cm}\\
\noindent\rule{\columnwidth}{0.4pt} \vspace{-.6cm}\\

\hspace{1.3cm}  \textsc{stress}  \hspace{.2cm} $\quad $ \textsc{Na-Crossing} $\quad $   \hspace{.1cm} \textsc{Combined}    \vspace{-.3cm} \\

    \centering

         \includegraphics[width=0.25\columnwidth, trim = 20 20 20 20, clip]{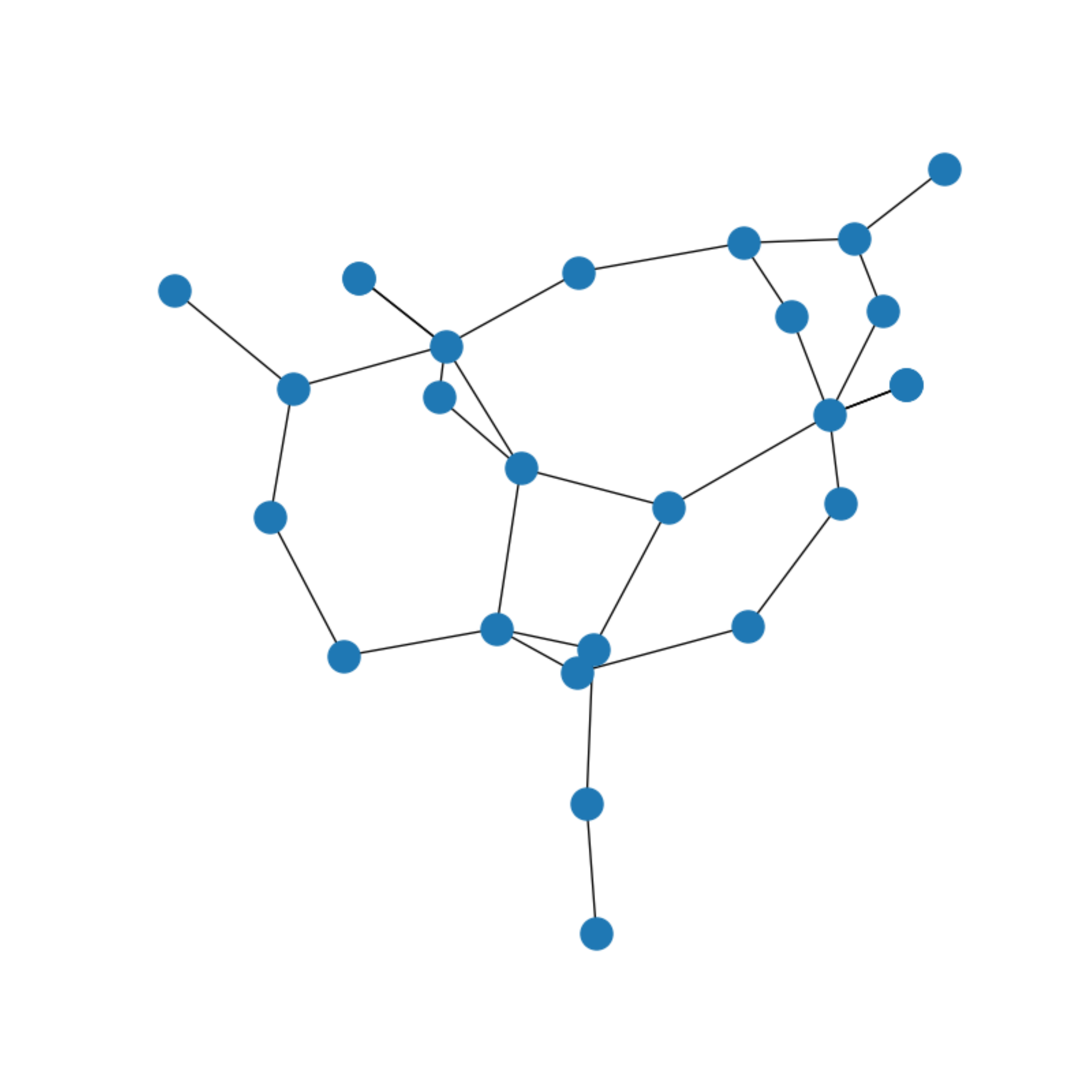}
         \includegraphics[width=0.25\columnwidth, trim = 20 20 20 20, clip]{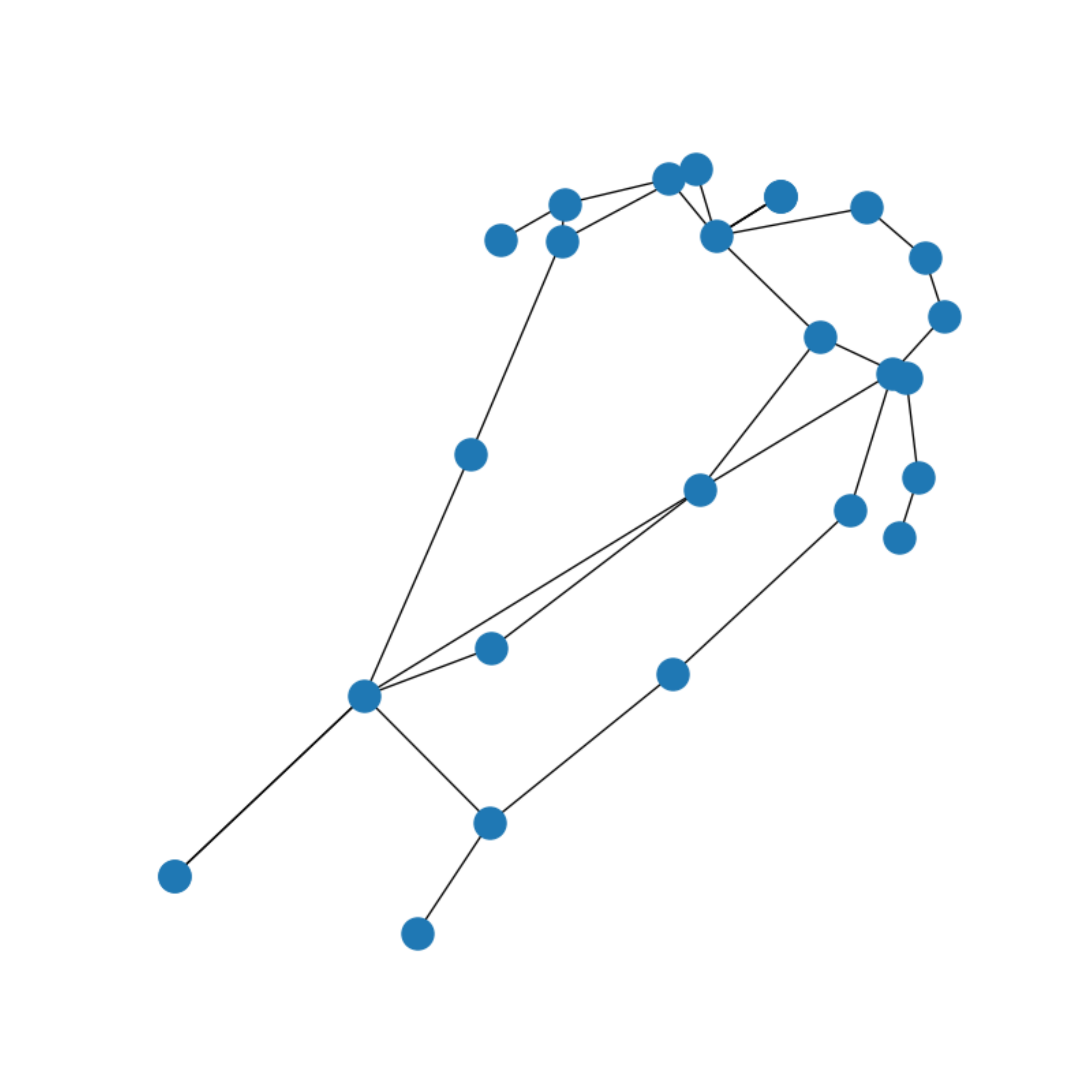}
         \includegraphics[width=0.25\columnwidth, trim = 20 20 20 20, clip]{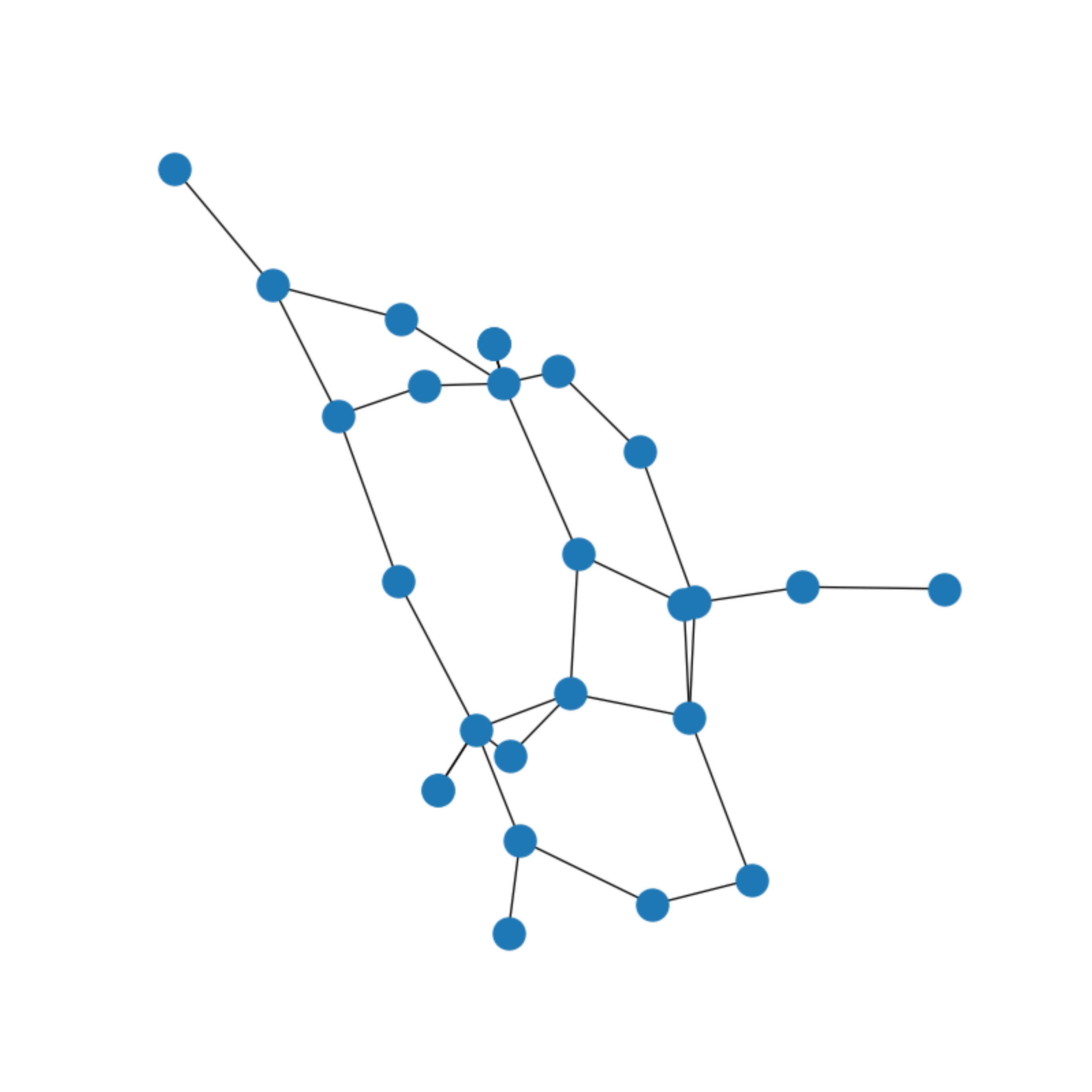}
         \includegraphics[width=0.25\columnwidth, trim = 20 20 20 20, clip]{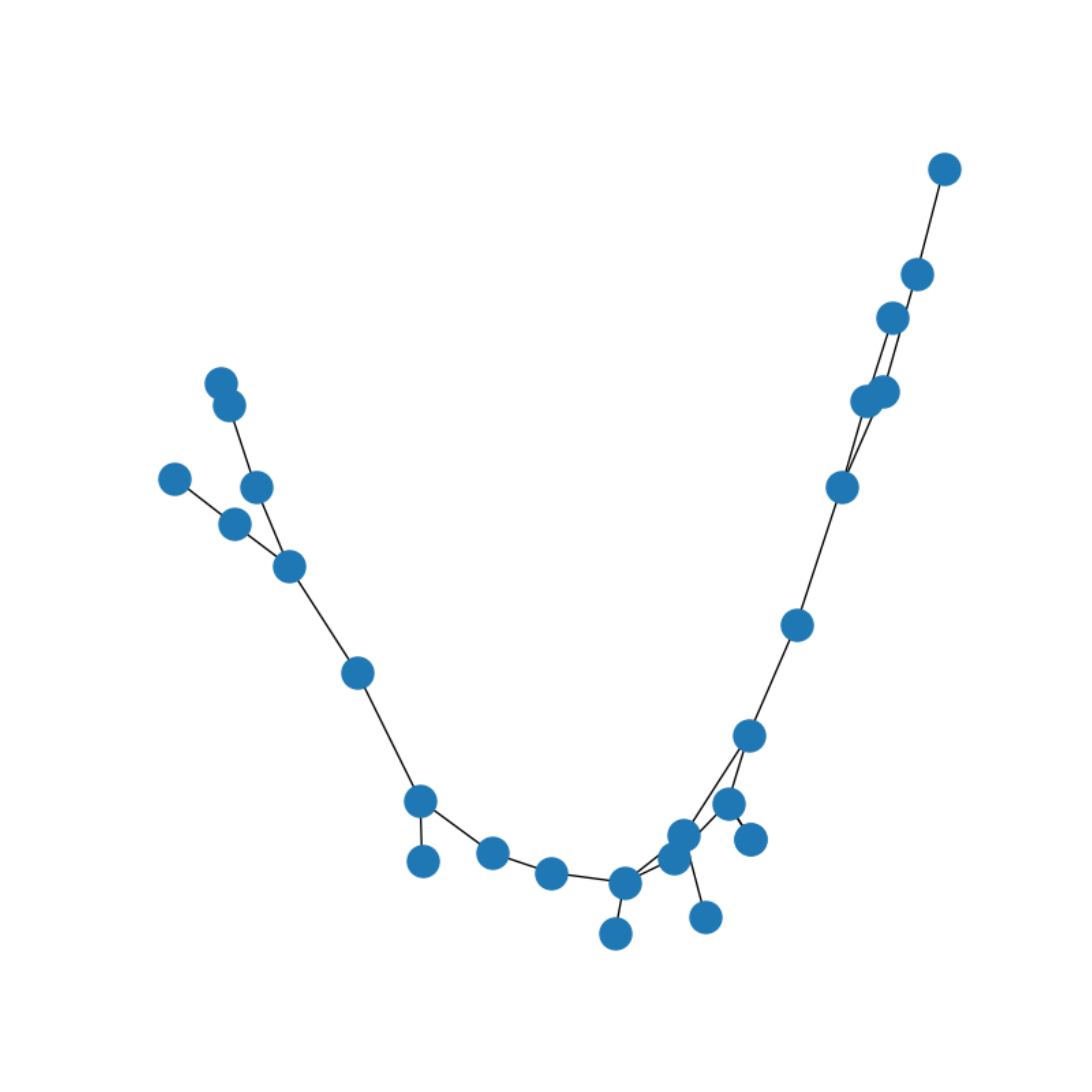}
        \includegraphics[width=0.25\columnwidth, trim = 20 20 20 20, clip]{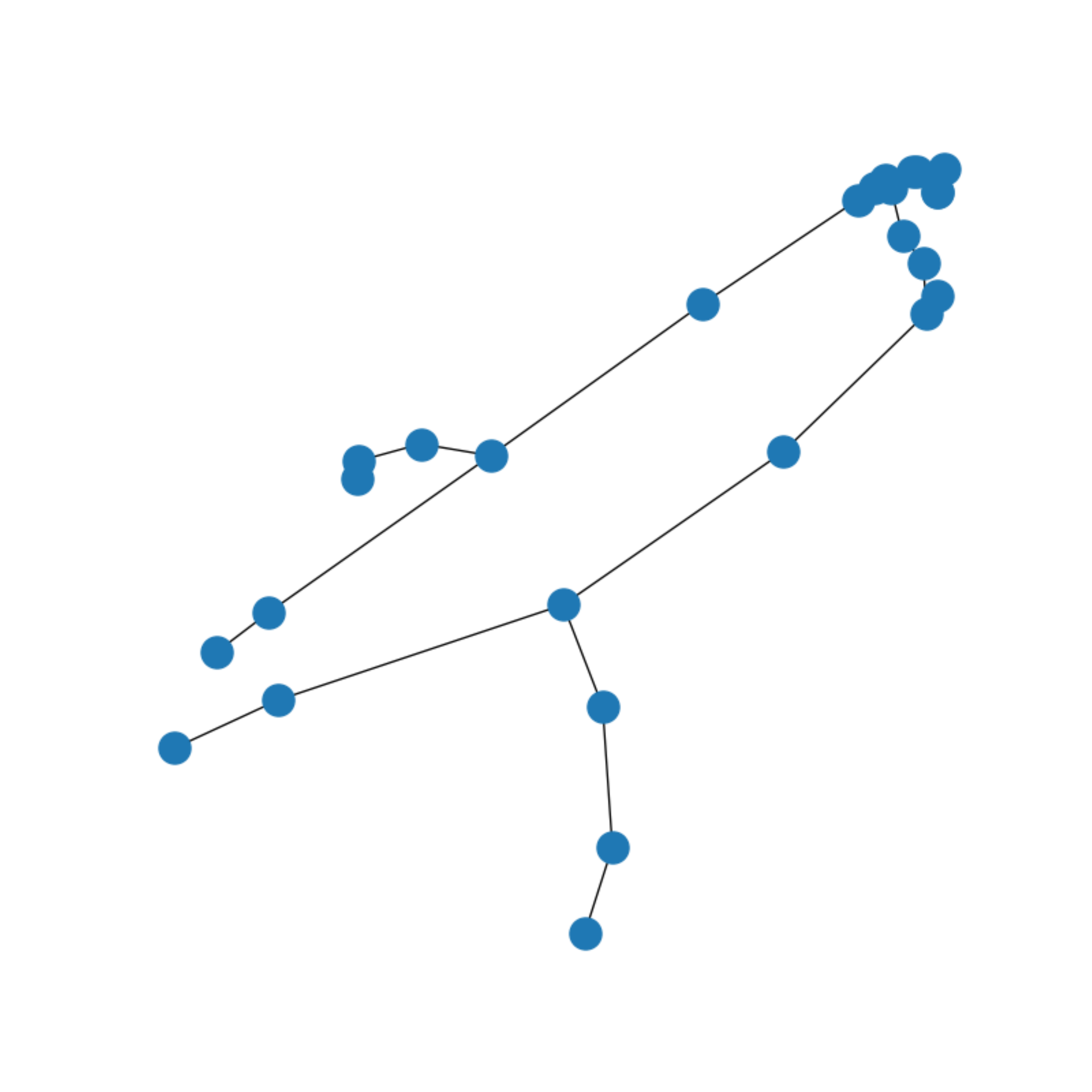}
         \includegraphics[width=0.25\columnwidth, trim = 20 20 20 20, clip]{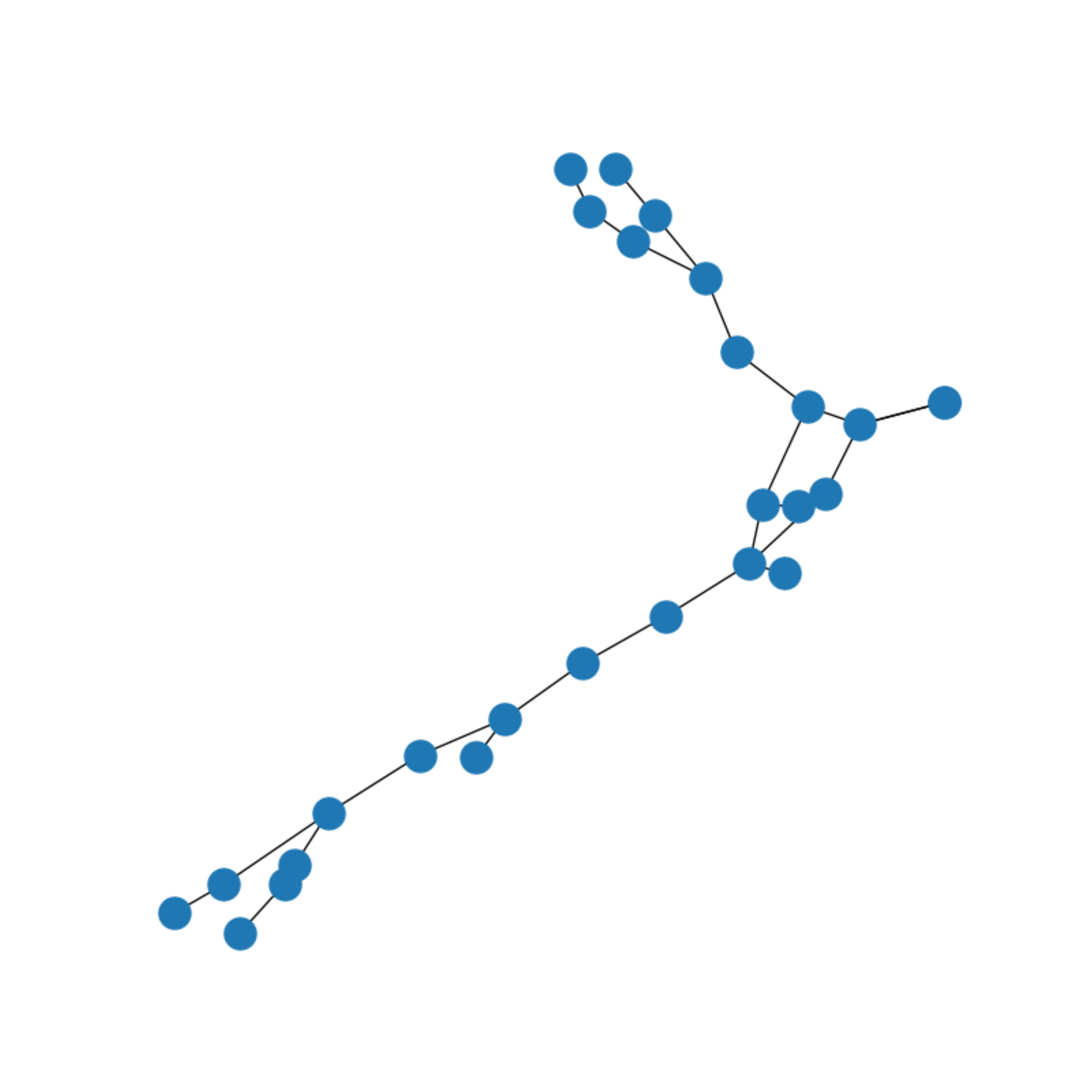}
         \includegraphics[width=0.25\columnwidth, trim = 20 20 20 20, clip]{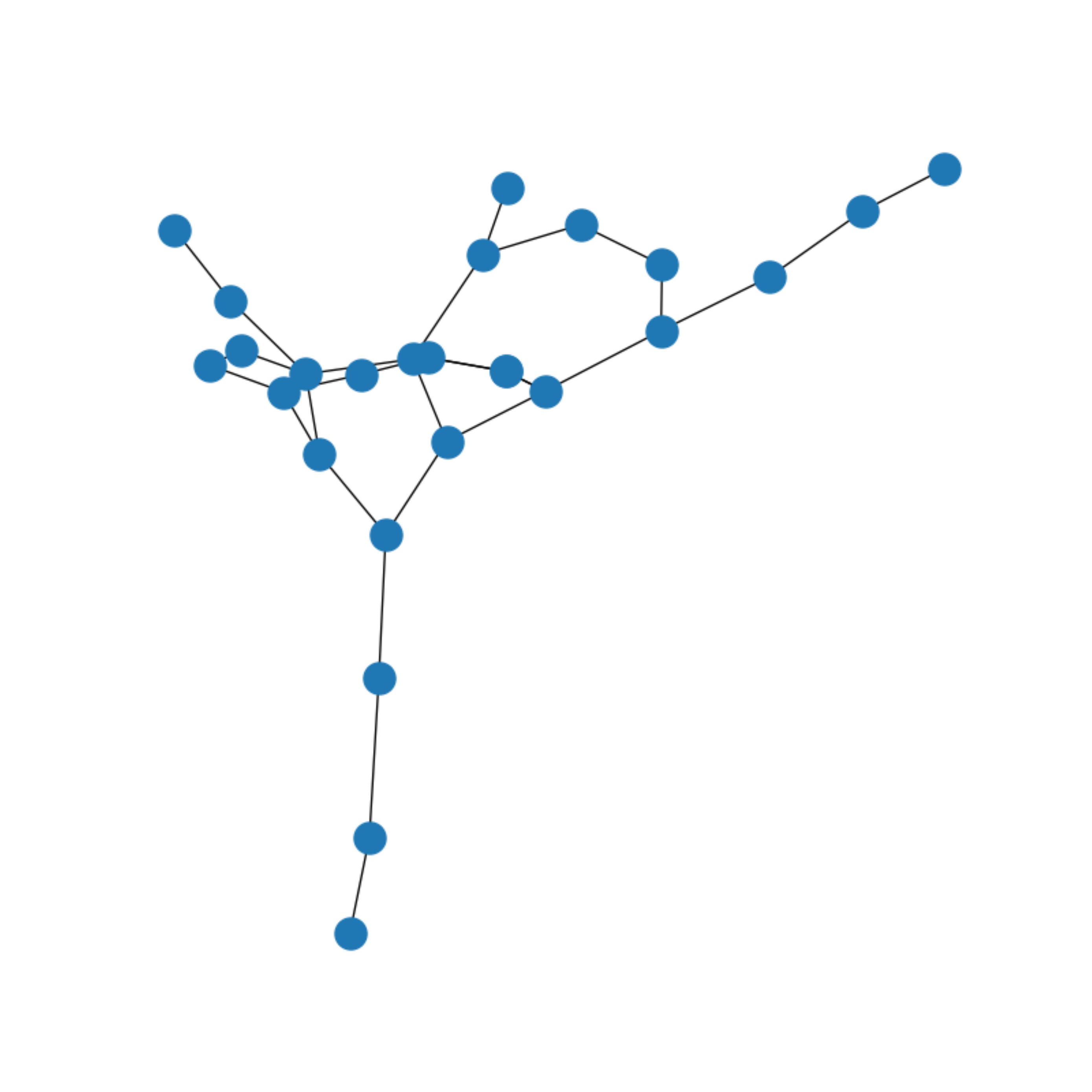}
         \includegraphics[width=0.25\columnwidth, trim = 20 20 20 20, clip]{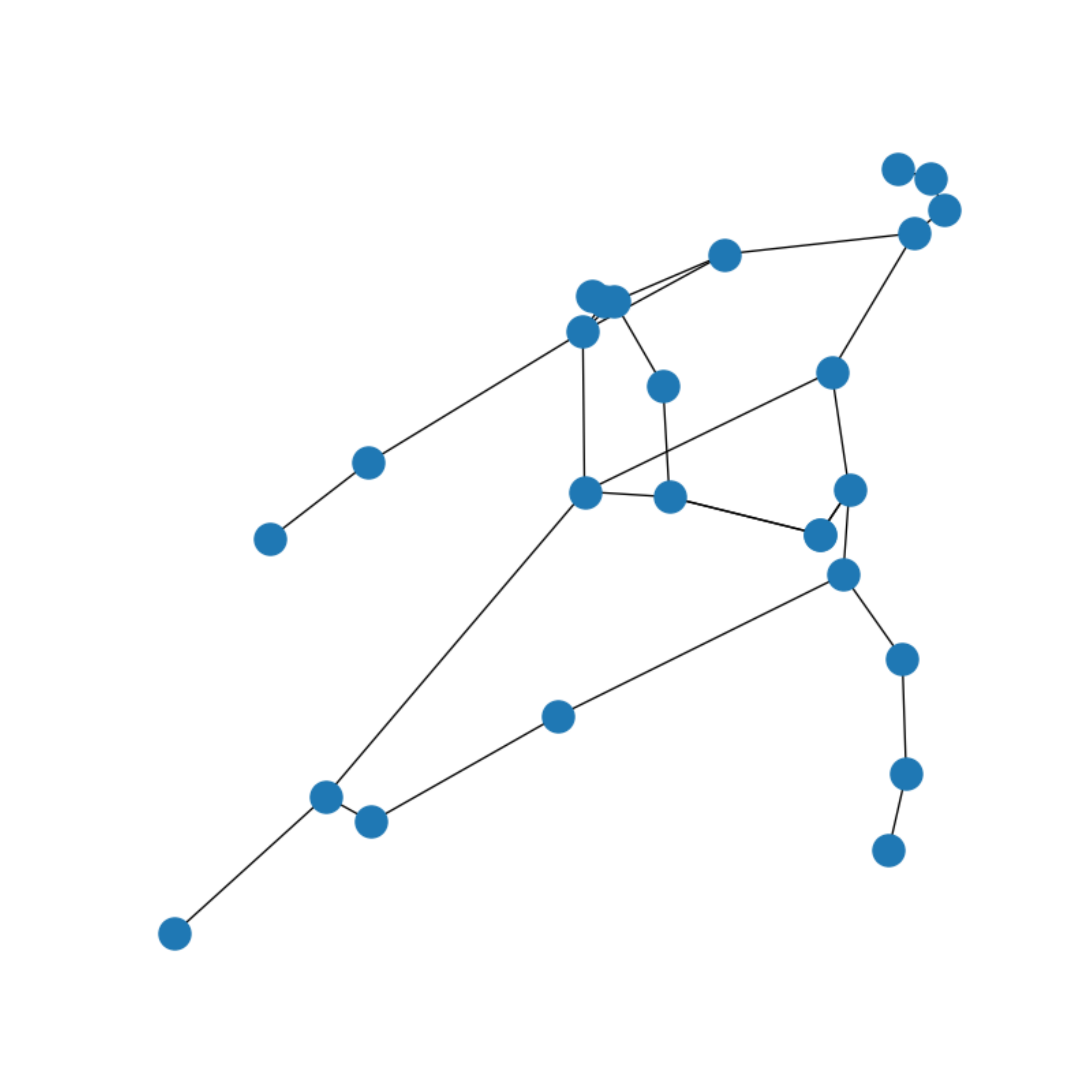}
         \includegraphics[width=0.25\columnwidth, trim = 20 20 20 20, clip]{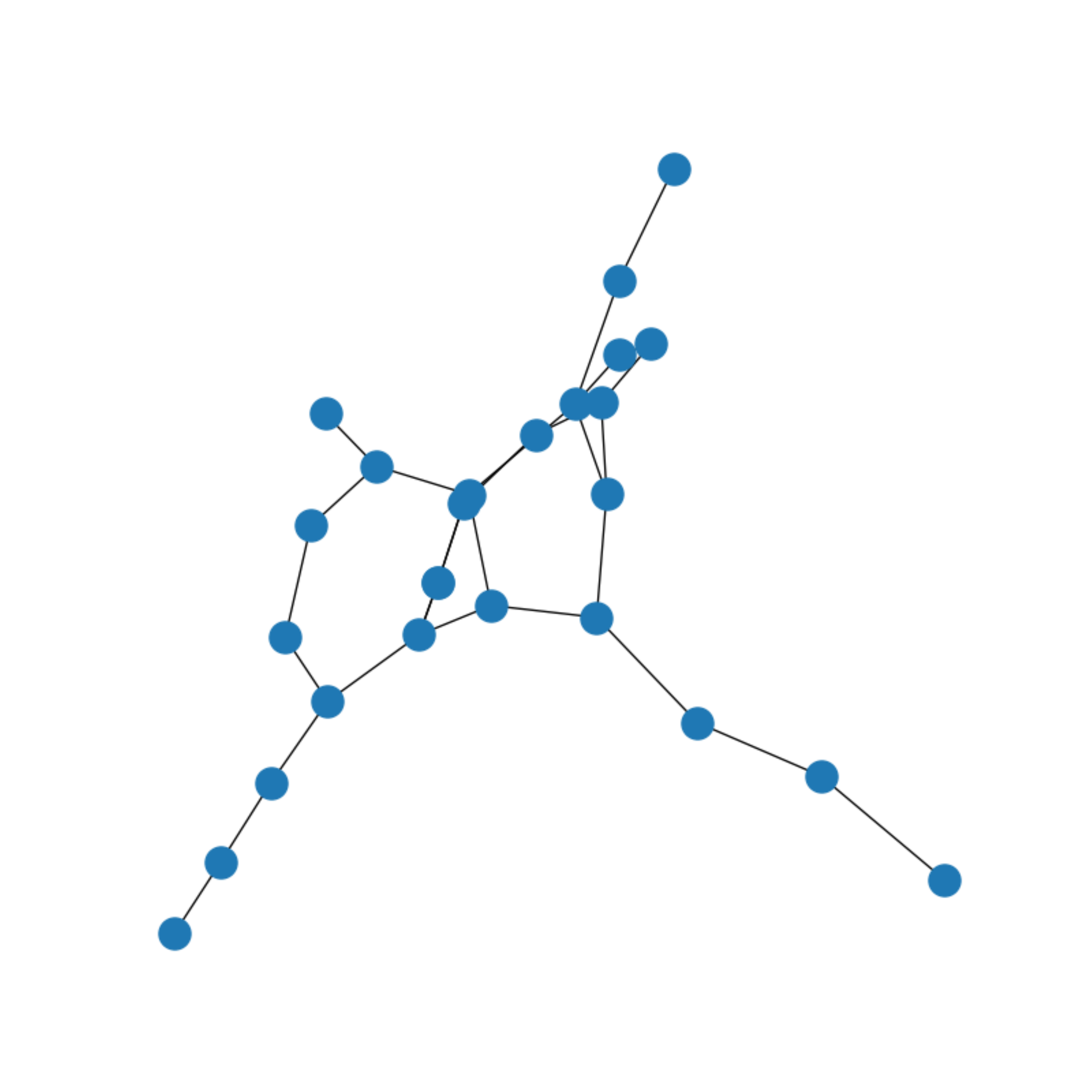}



    \caption{\textsc{Learning from the Neural Aesthete}. We report the layouts obtained on three randomly picked test graphs \mt{from the Rome dataset}, one for each row. Left-to-right: Graph layout generated by optimizing the stress loss function, the edge-crossing Neural Aesthete based loss (denoted with NA-Crossing), the combination of the two losses with a weighing factor $\lambda=0.5$. }
    \label{fig:edge_cross_gnd}
\end{figure}

We report in Figure \ref{fig:edge_cross_gnd} some qualitative results on three test graphs (one for each row). We compare the layout obtained optimizing the stress function (first column, see Section \ref{sec:stress}), the edge-crossing Neural Aesthete (second column) and the combination of the two losses. 

The styles of the generated layout are recognizable with respect to the plain optimization of the Neural Aesthete with Gradient Descent (see Figure \ref{fig:edge_cross}), meaning that the GND framework is able to fit the loss provided by the Neural Aesthete and to generalize it to unseen graphs.
Noticeable, the introduction of the combined loss functions (third column in Figure \ref{fig:edge_cross_gnd}) helps in better differentiating the nodes in the graph with respect to the case of solely optimizing stress. \mt{The Neural Aesthete guided layouts (second and third column) tend to avoid edge intersections, as expected.}
This opens the road to further studies in this direction, leveraging the generality of the Neural Aesthetes approach and the representation capability of GNNs.

\subsection{Computational Complexity}
\label{sec:comp}

The proposed framework leverages the same computational structure of the underlying GNN model, which we can generally describe, for each parameter update, as linear with respect to the edge number  $\mathcal{O}\big(T(|\mathcal{V}|+ |\mathcal{E}|)\big)$, where $T$ is the number of iterations/layers, $|\mathcal{V}|$ the number of nodes and $|\mathcal{E }|$ the number of edges. Through our approach, there is not any increase in the computation related to the graph topology or the edge connection patterns.
At inference time, the only additional requirement is the computation of the Laplacian PEs, requiring $\mathcal{O}(\mathcal{E}^{3/2})$, with $\mathcal{E}$ being the number of edges, that however can be improved with the Nystrom method \cite{fowlkes2004spectral, dwivedi2020benchmarking}. 

\begin{figure}[h]
\noindent\rule{\columnwidth}{0.4pt} \vspace{-0.85cm}\\

\hspace{1.4cm} \textsc{\mt{SuiteSparse Matrix Collection}} \vspace{-.2cm}\\
\noindent\rule{\columnwidth}{0.4pt} \vspace{-.6cm}\\

 \hspace{2.cm}  \textsc{GND}  \hspace{1.cm} $\quad $ \texttt{sfdp} $\quad $   \hspace{.2cm} PivotMDS    \vspace{-.3cm} \\


       \raisebox{1.2cm}{\tiny  
       \parbox{1.5cm}{ {  \footnotesize{\texttt{dwt\_162}} \\
         $|\mathcal{V}|=162$\\
         $|\mathcal{E}|=672$}\\
         }}
        \includegraphics[width=0.24\columnwidth, trim = 20 20 20 20, clip]{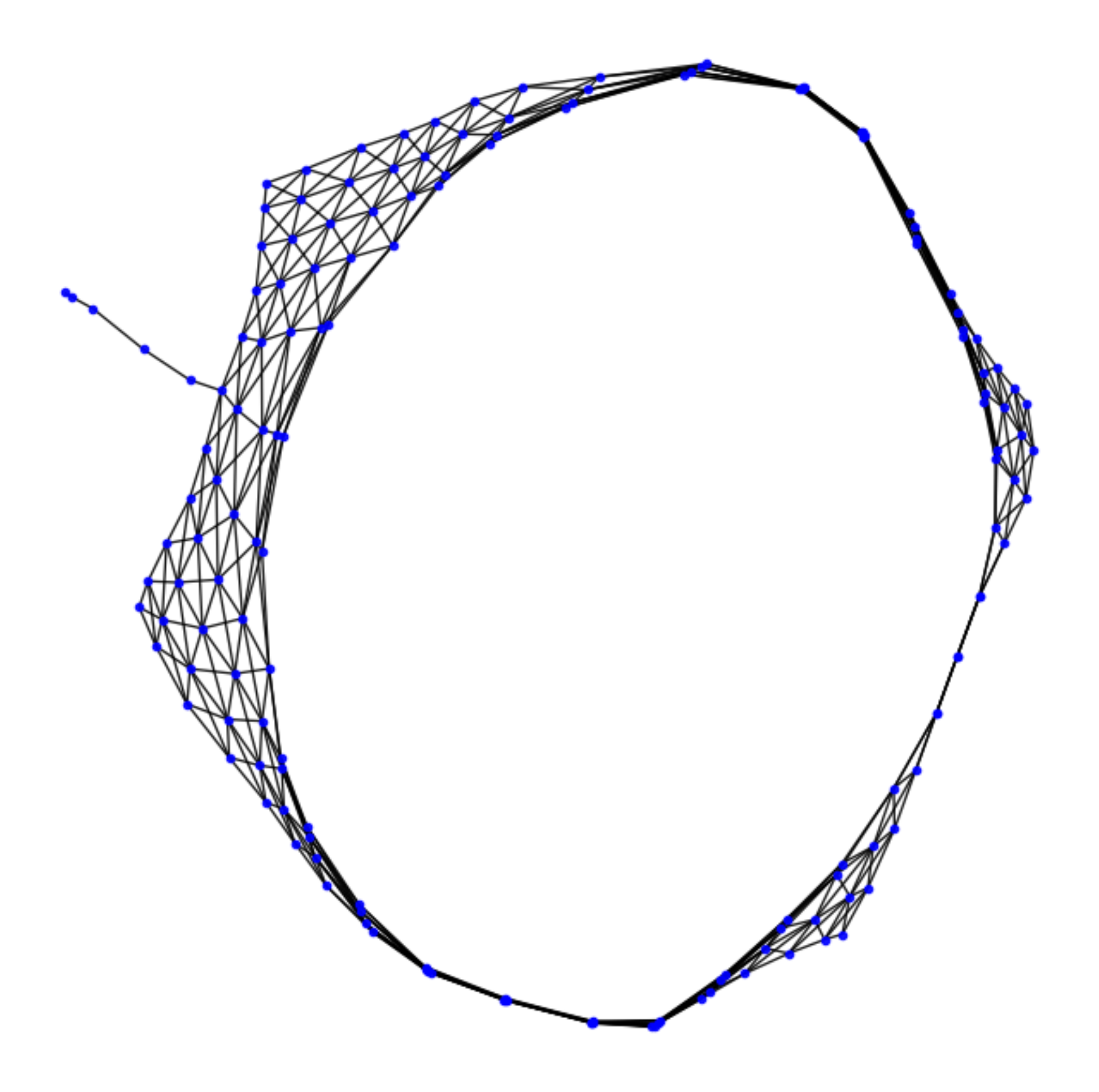}
         \includegraphics[width=0.24\columnwidth, trim = 20 20 20 20, clip]{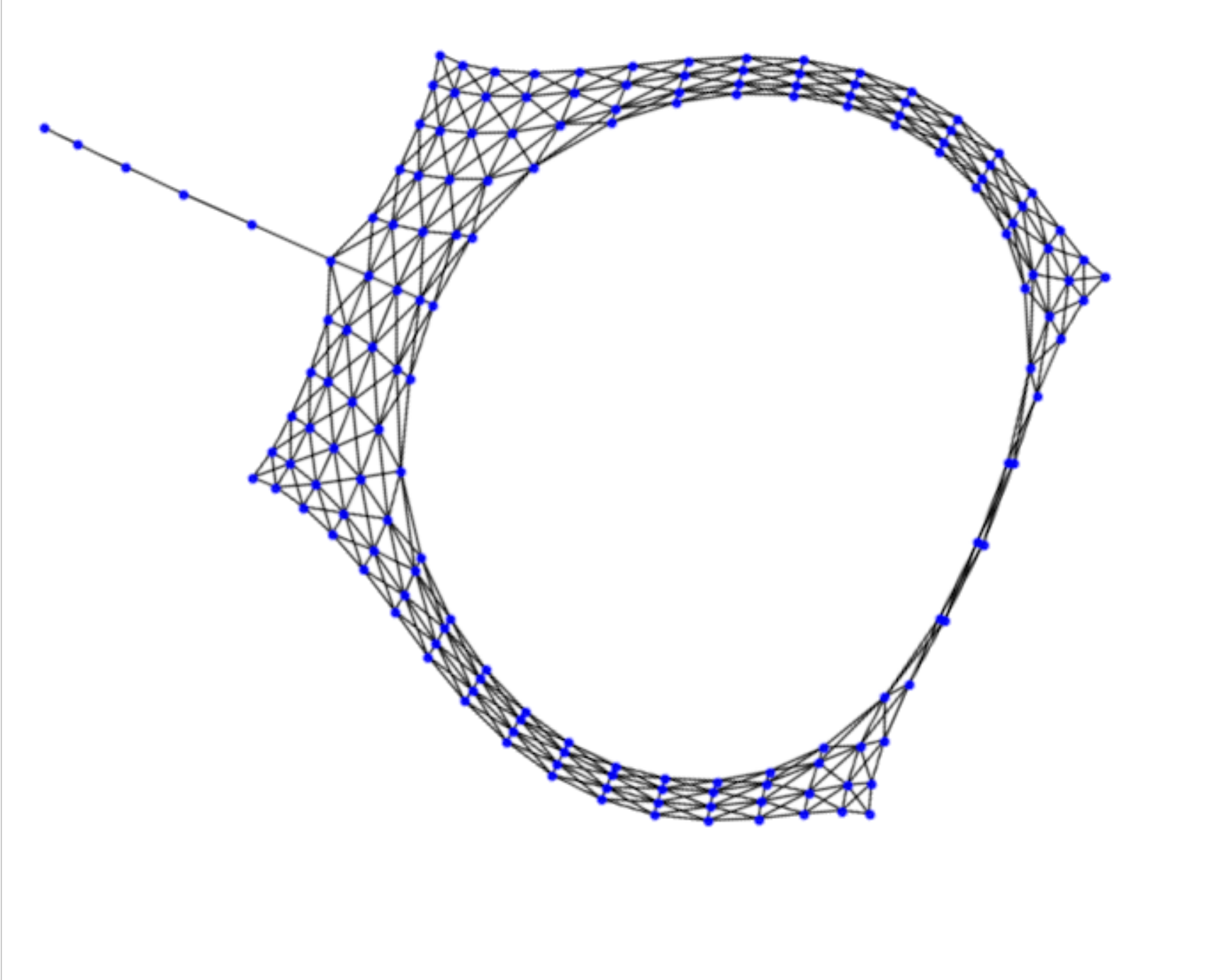}
         \includegraphics[width=0.24\columnwidth, trim = 20 20 20 20, clip]{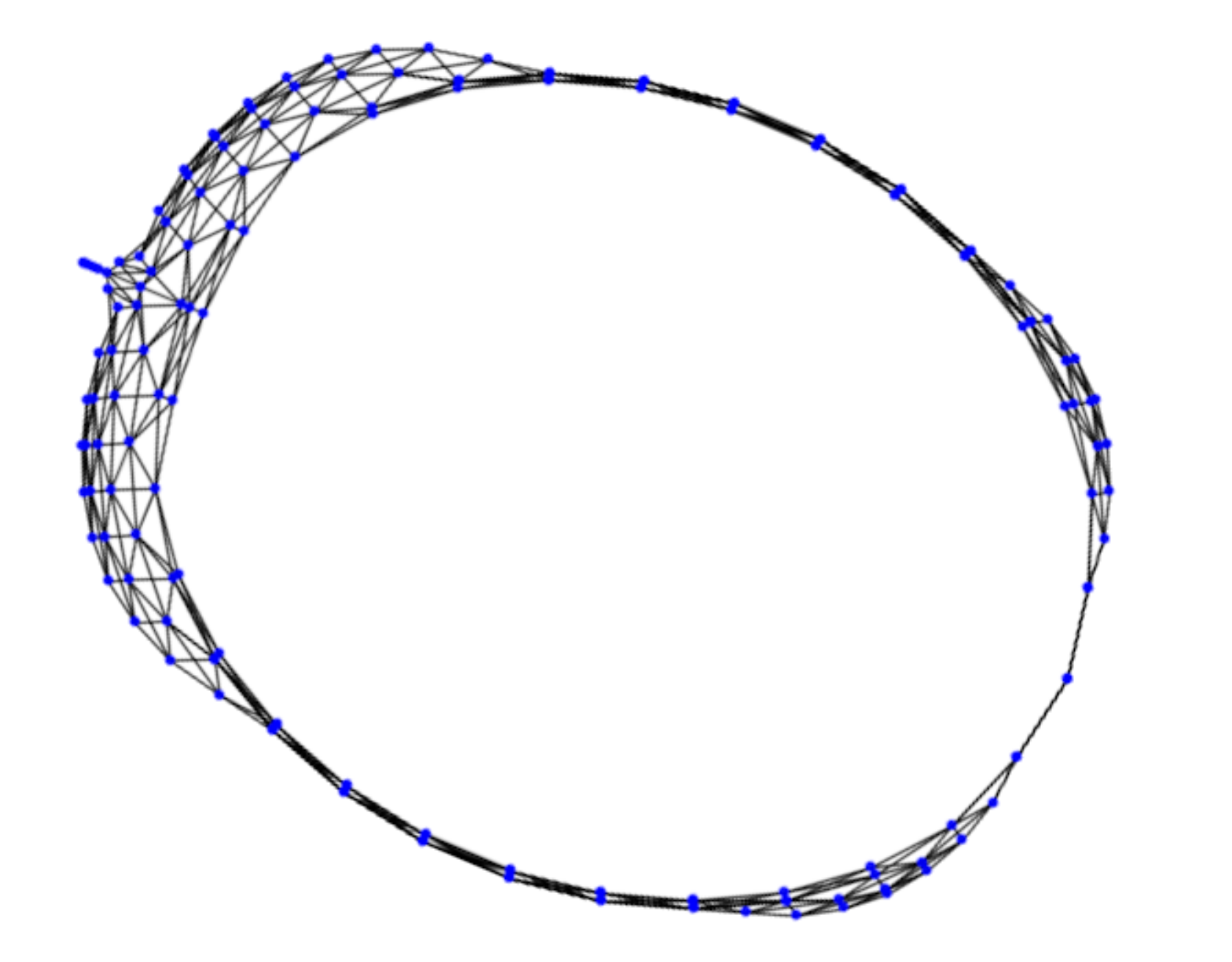}

        \raisebox{1.cm}{\tiny  
       \centering
       \parbox{1.5cm}{\footnotesize{\texttt{dwt\_307}} \\
         $|\mathcal{V}|=307$ \\$|\mathcal{E}|=1415$\\
         }}
        \includegraphics[width=0.24\columnwidth, trim = 20 20 20 20, clip]{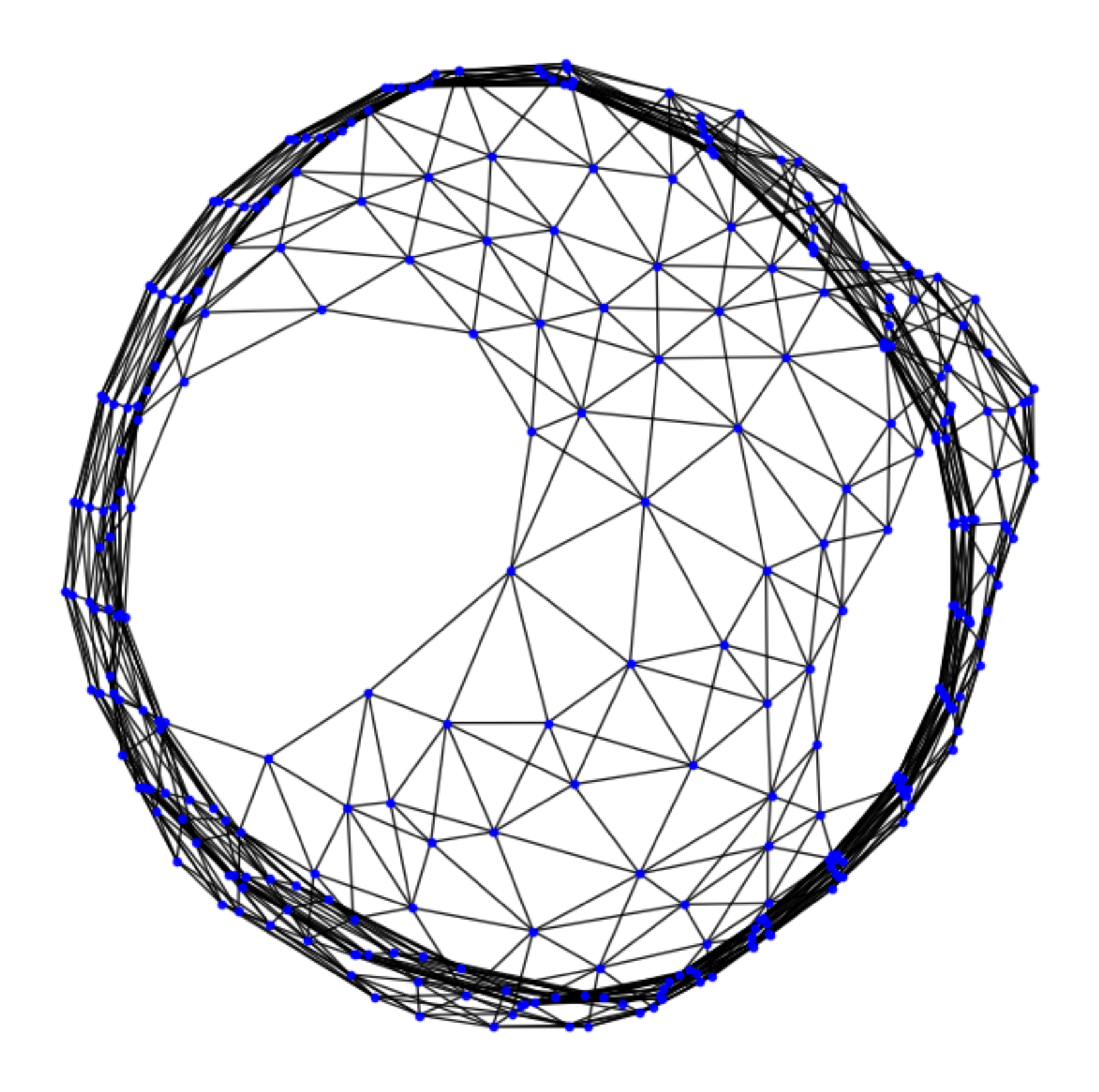}
         \includegraphics[width=0.24\columnwidth, trim = 20 20 20 20, clip]{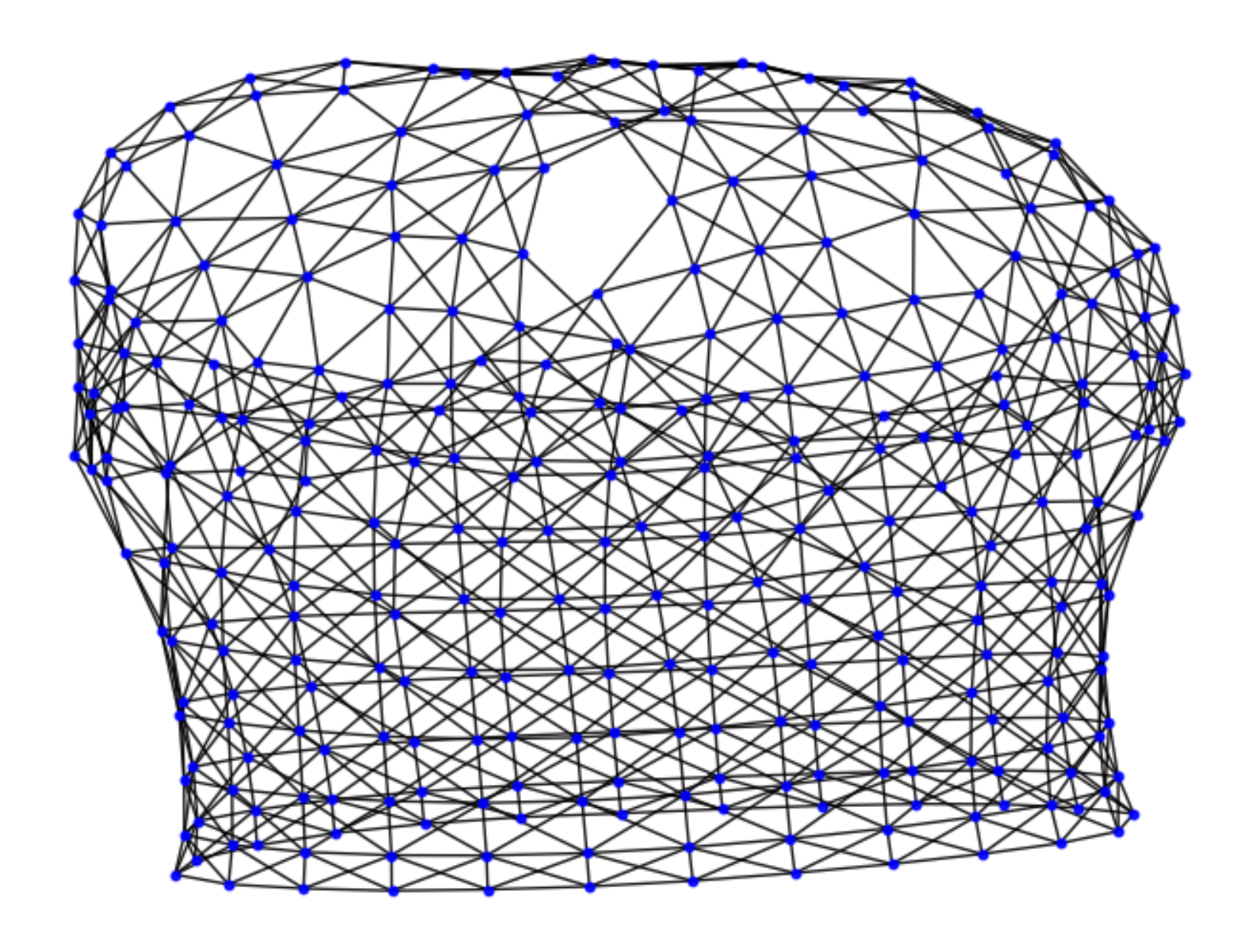}
         \includegraphics[width=0.24\columnwidth, trim = 20 20 20 20, clip]{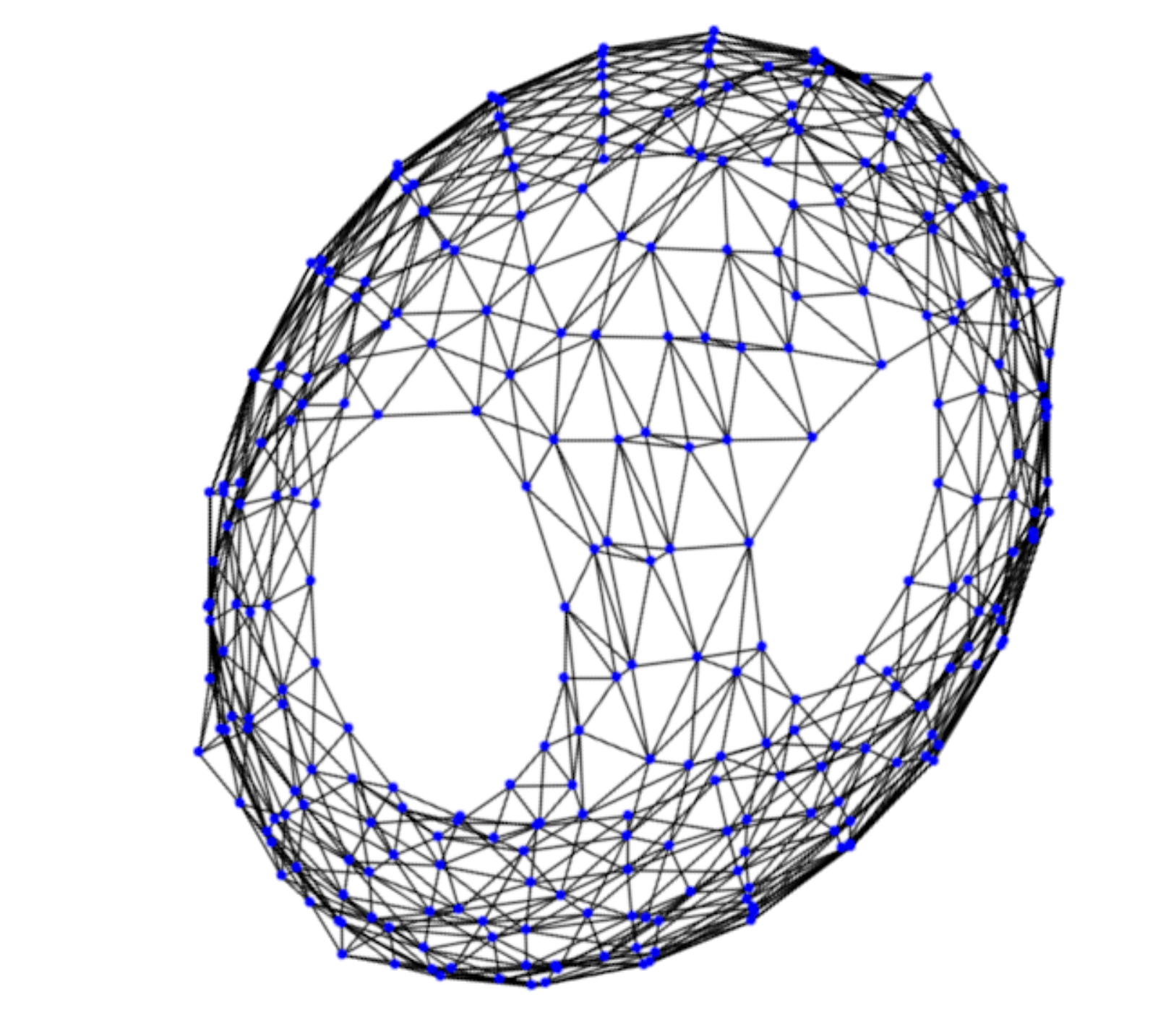}

         \raisebox{1.cm}{\tiny  
       \centering \parbox{1.5cm}{\footnotesize{\texttt{dwt\_503}} \\
         $|\mathcal{V}|=503$ \\$|\mathcal{E}|=3k$\\
         }}
         \includegraphics[width=0.24\columnwidth, trim = 20 20 20 20, clip]{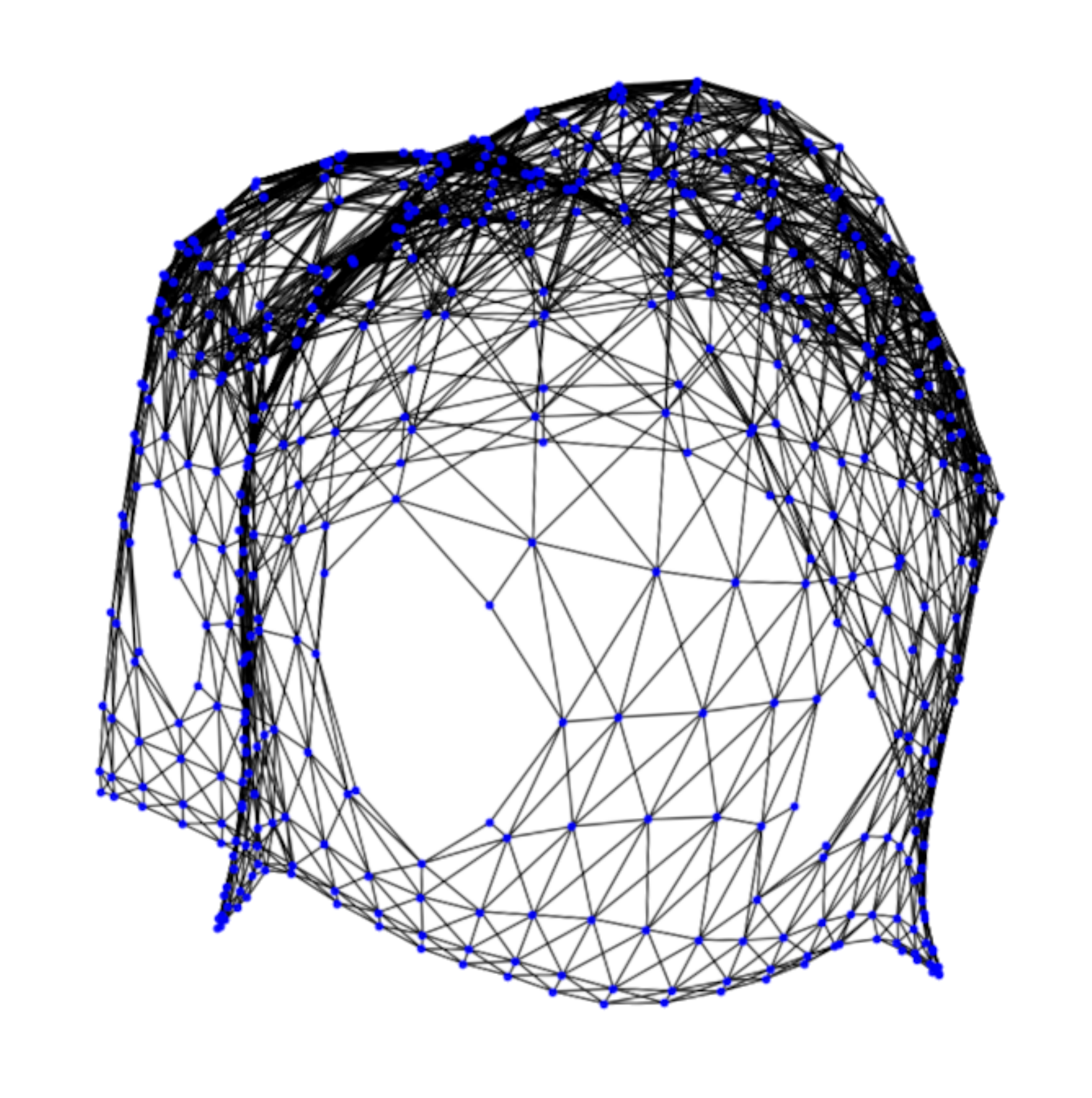}
         \includegraphics[width=0.24\columnwidth, trim = 20 20 20 20, clip]{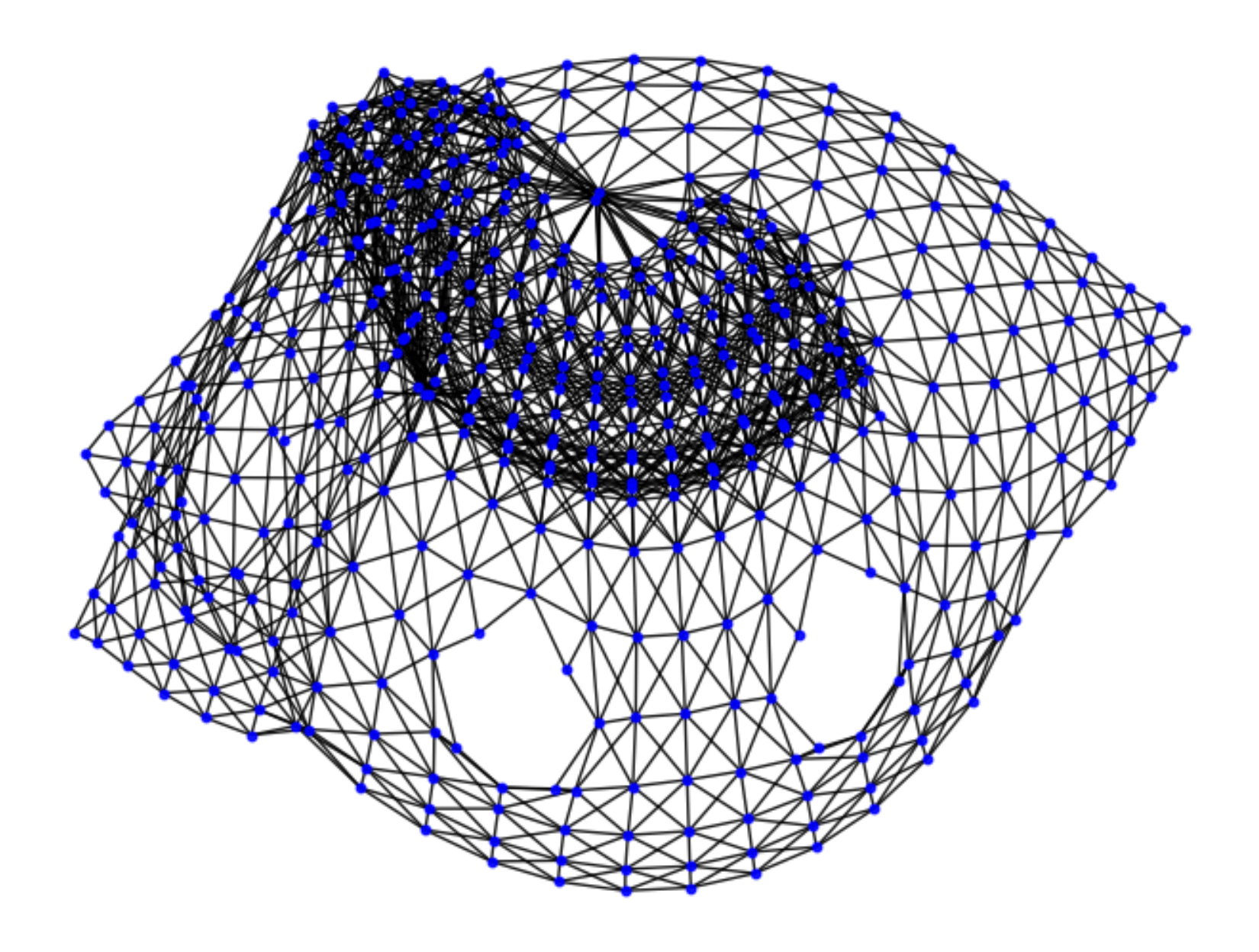}
         \includegraphics[width=0.24\columnwidth, trim = 20 20 20 20, clip]{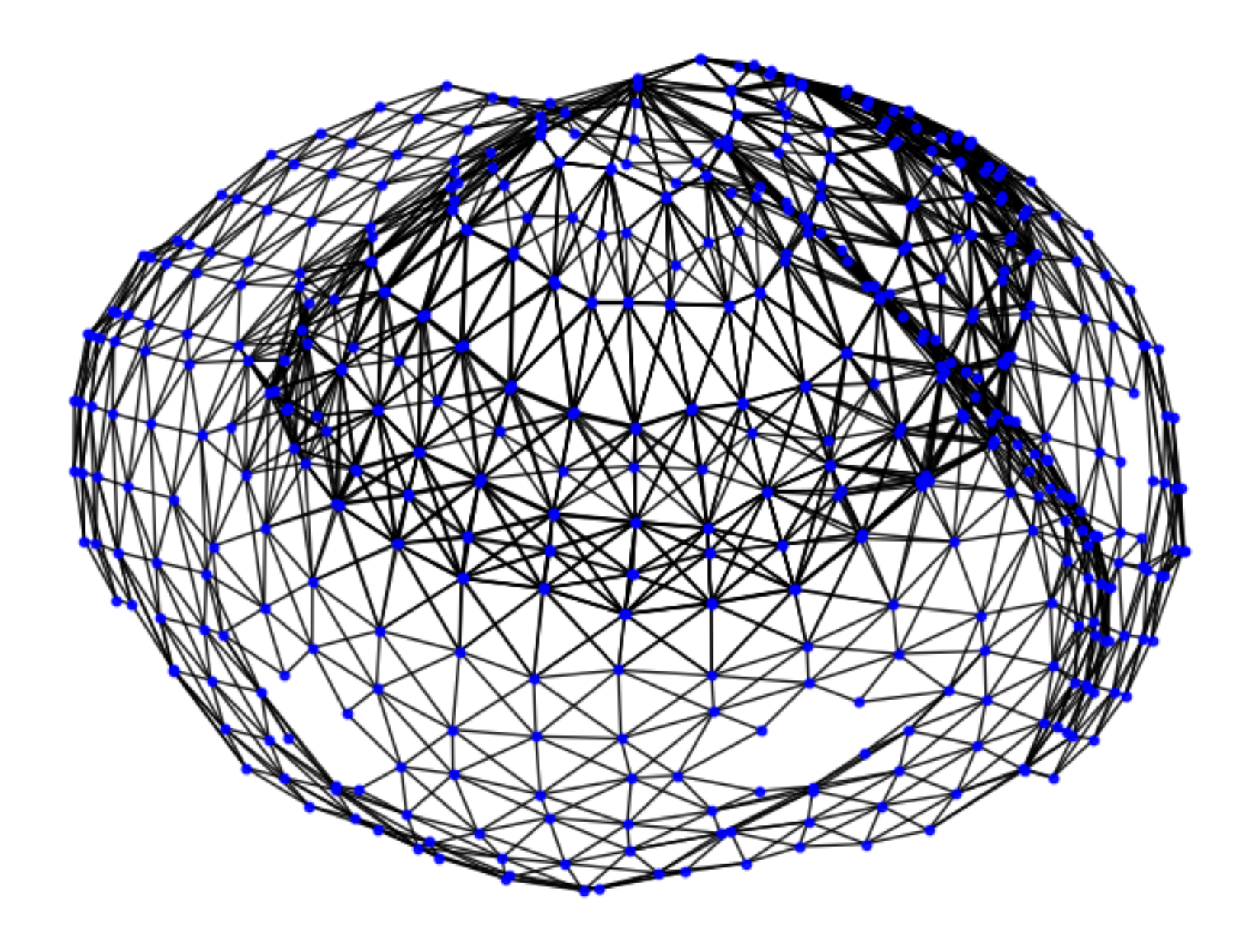}

        \raisebox{1.cm} {\tiny  
       \centering \parbox{1.5cm}{\footnotesize{\texttt{dwt\_992}} \\
         $|\mathcal{V}|=992$ \\$|\mathcal{E}|=9k$\\
         }}
         \includegraphics[width=0.24\columnwidth, trim = 20 20 20 20, clip]{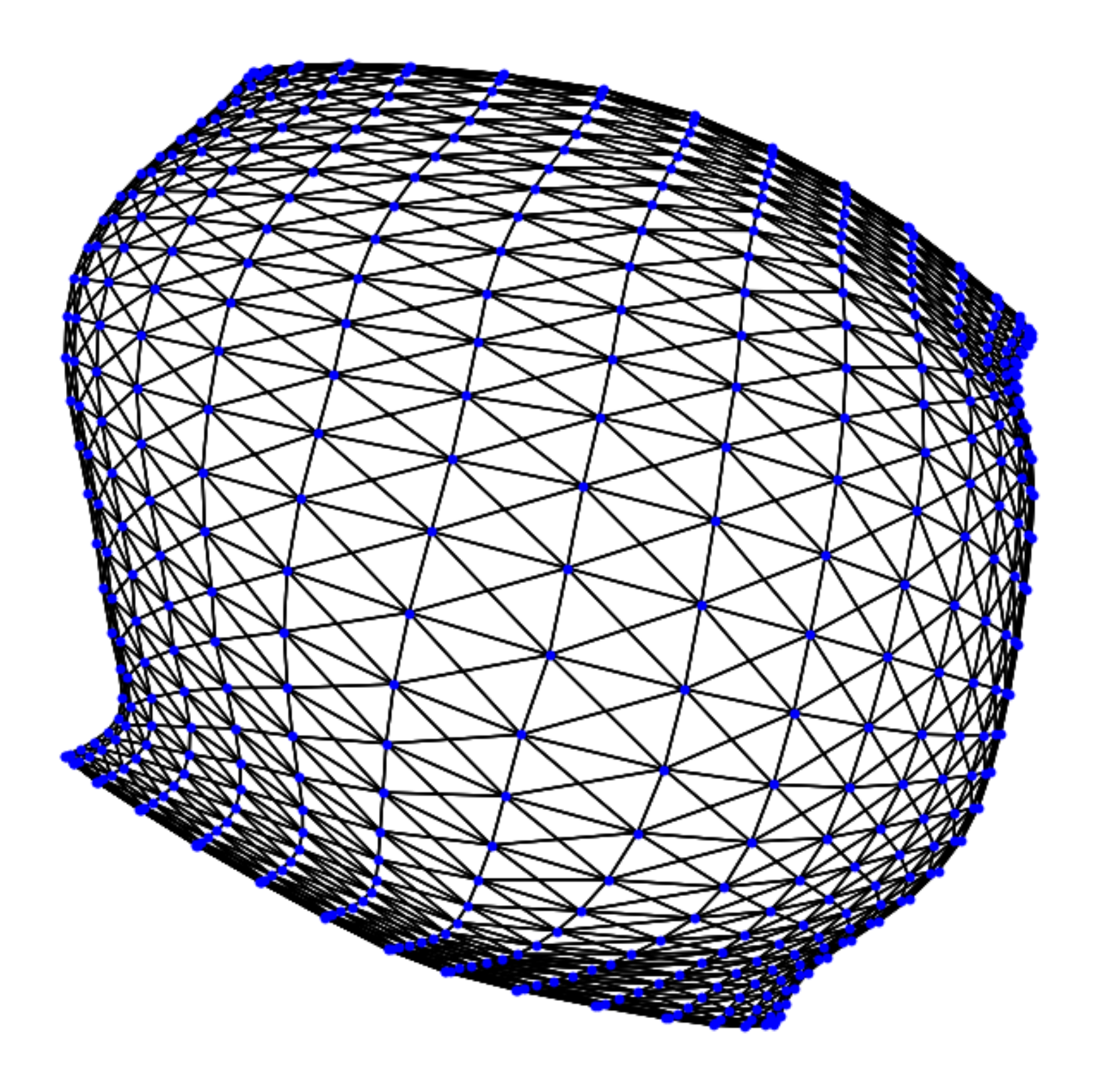}
         \includegraphics[width=0.24\columnwidth, trim = 20 20 20 20, clip]{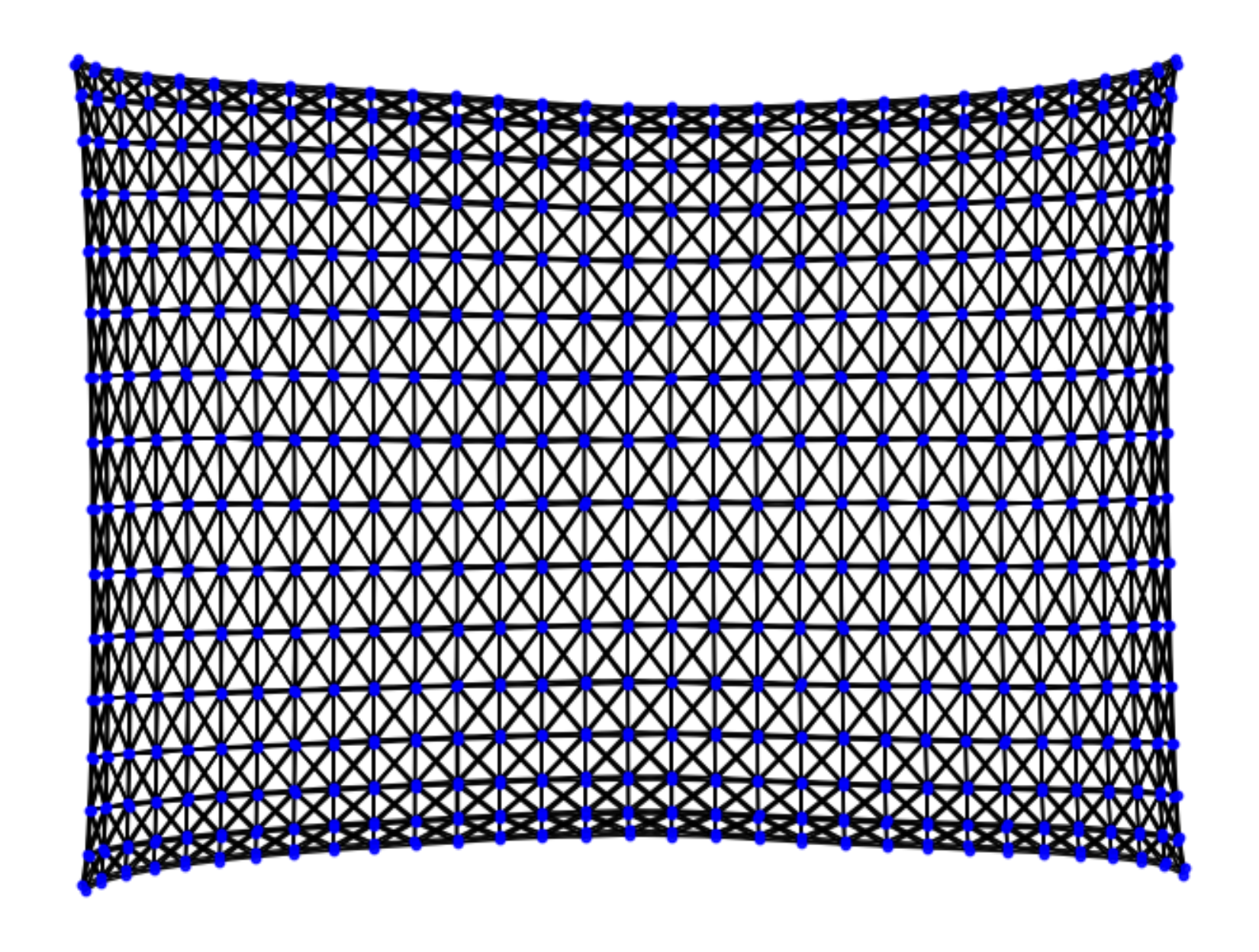}
         \includegraphics[width=0.24\columnwidth, trim = 20 20 20 20, clip]{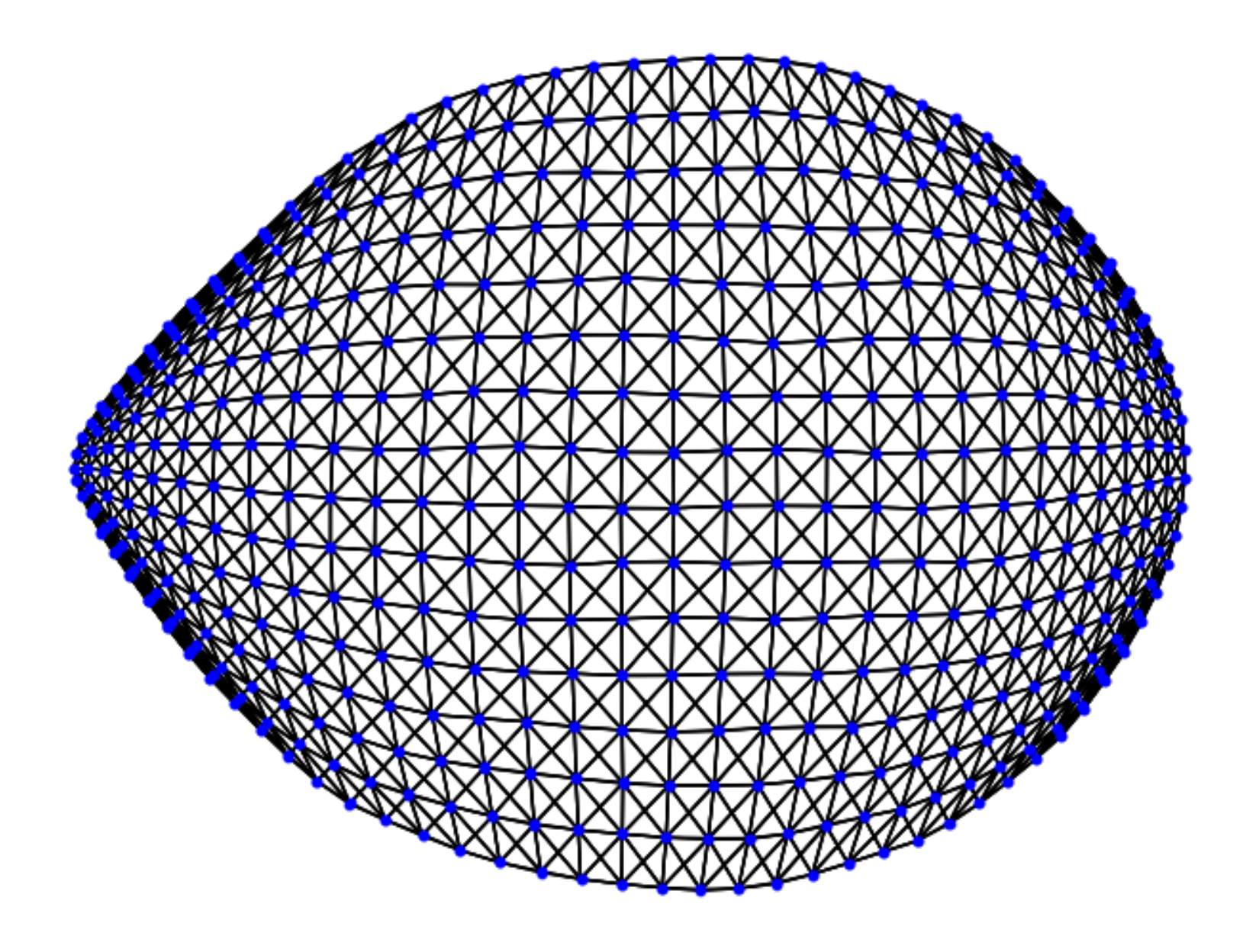}

         \raisebox{.7cm}{\tiny  
       \centering \parbox{1.5cm}{\footnotesize{\texttt{dwt\_1005}} \\
         $|\mathcal{V}|=1k$ \\$|\mathcal{E}|=5k$\\
         }}
         \includegraphics[width=0.24\columnwidth, trim = 20 20 20 20, clip]{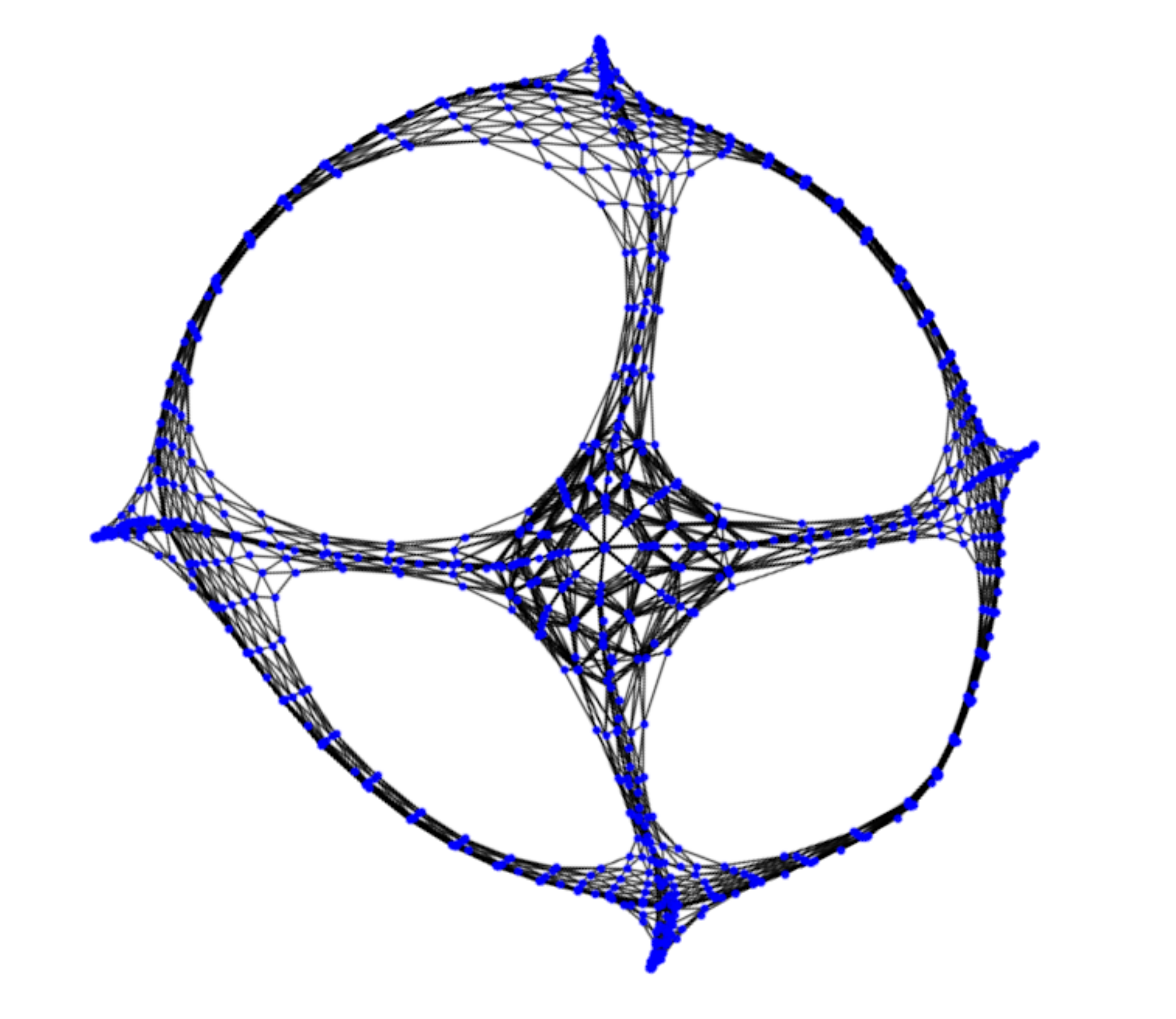}
         \includegraphics[width=0.24\columnwidth, trim = 20 20 20 20, clip]{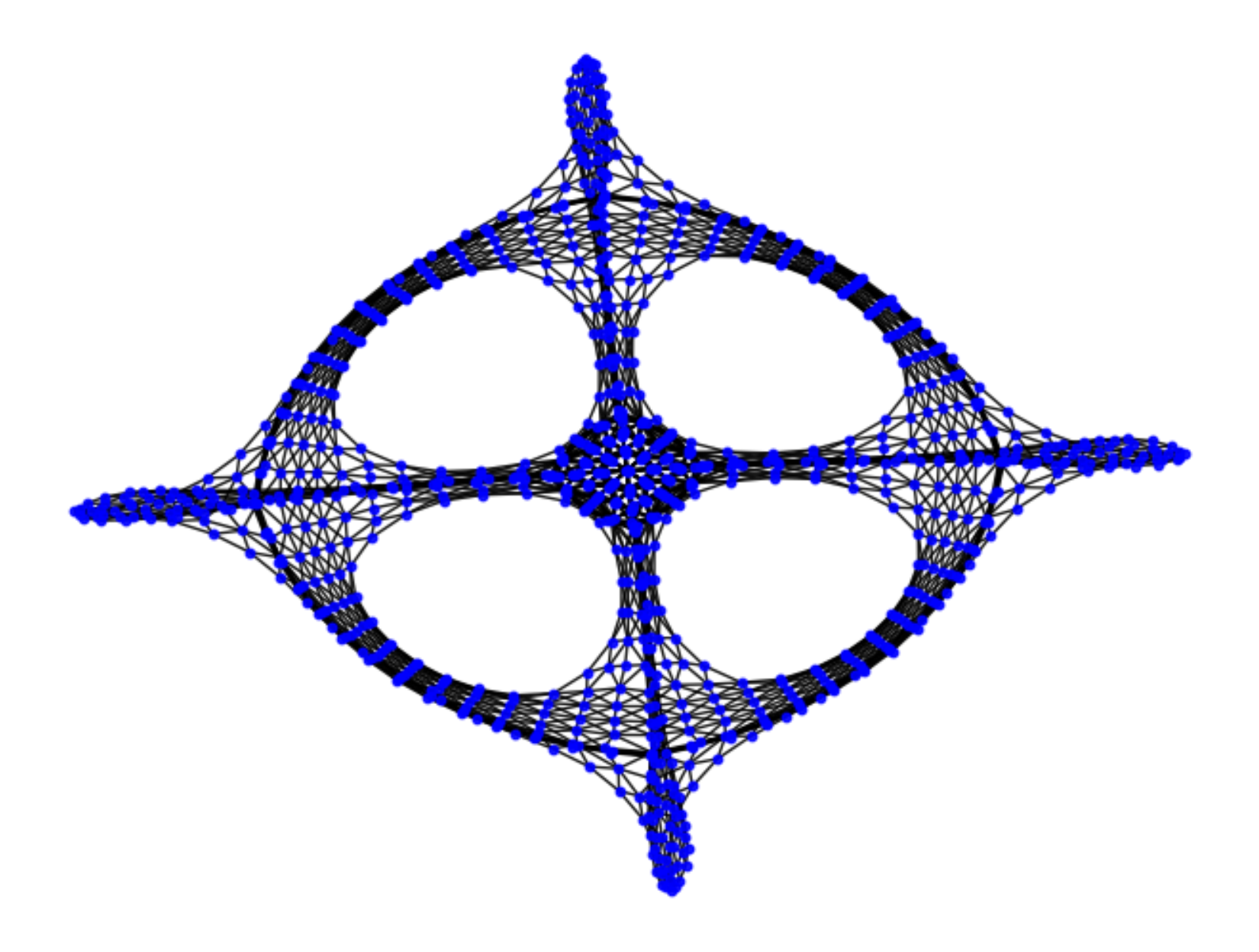}
         \includegraphics[width=0.24\columnwidth, trim = 20 20 20 20, clip]{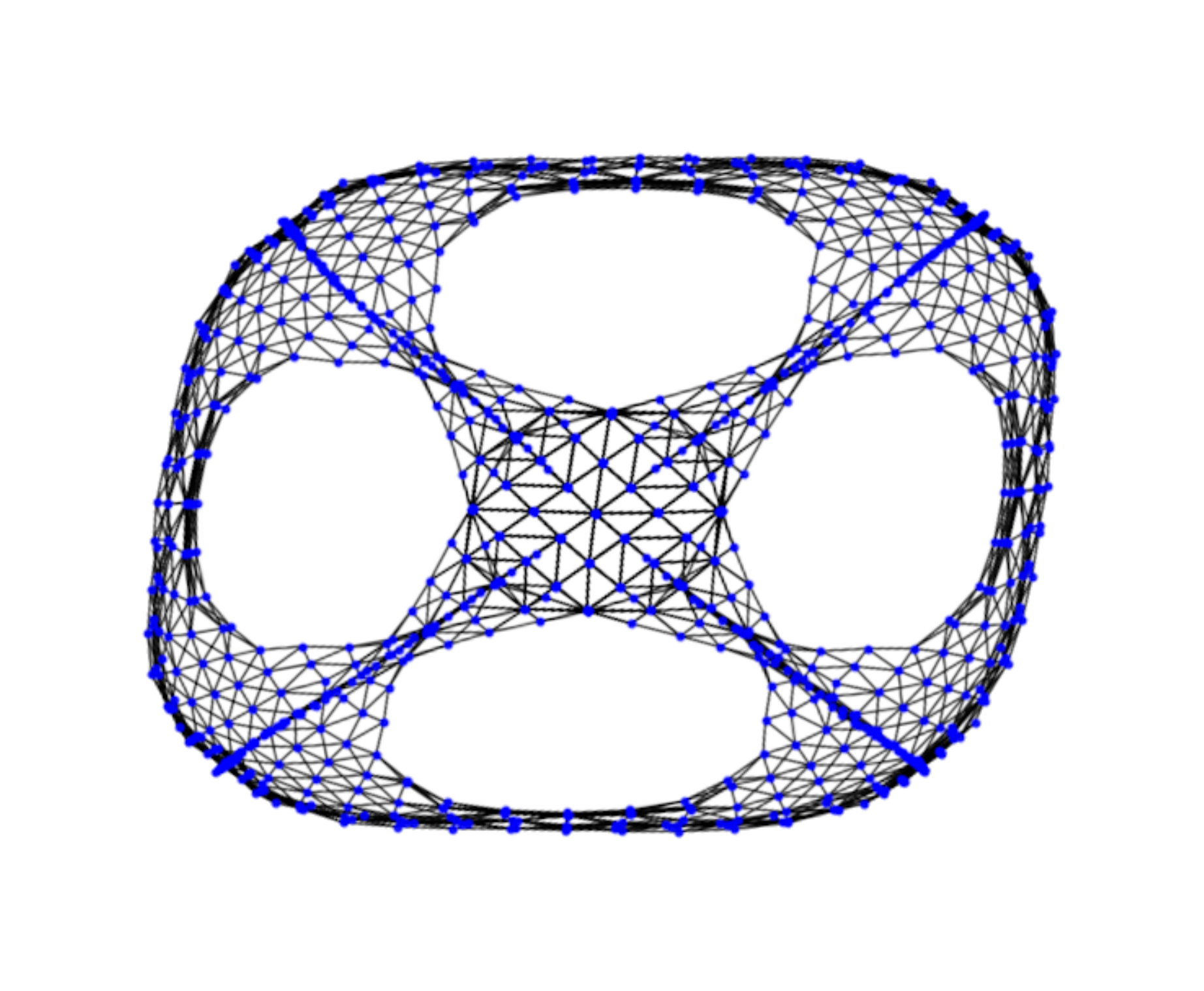}

        \raisebox{.5cm} {\tiny  
       \centering \parbox{1.5cm}{\footnotesize{\texttt{rdist3a}} \\
         $|\mathcal{V}|=2k$ \\$|\mathcal{E}|=57k$\\
         }}
         \includegraphics[width=0.24\columnwidth, trim = 20 20 20 20, clip]{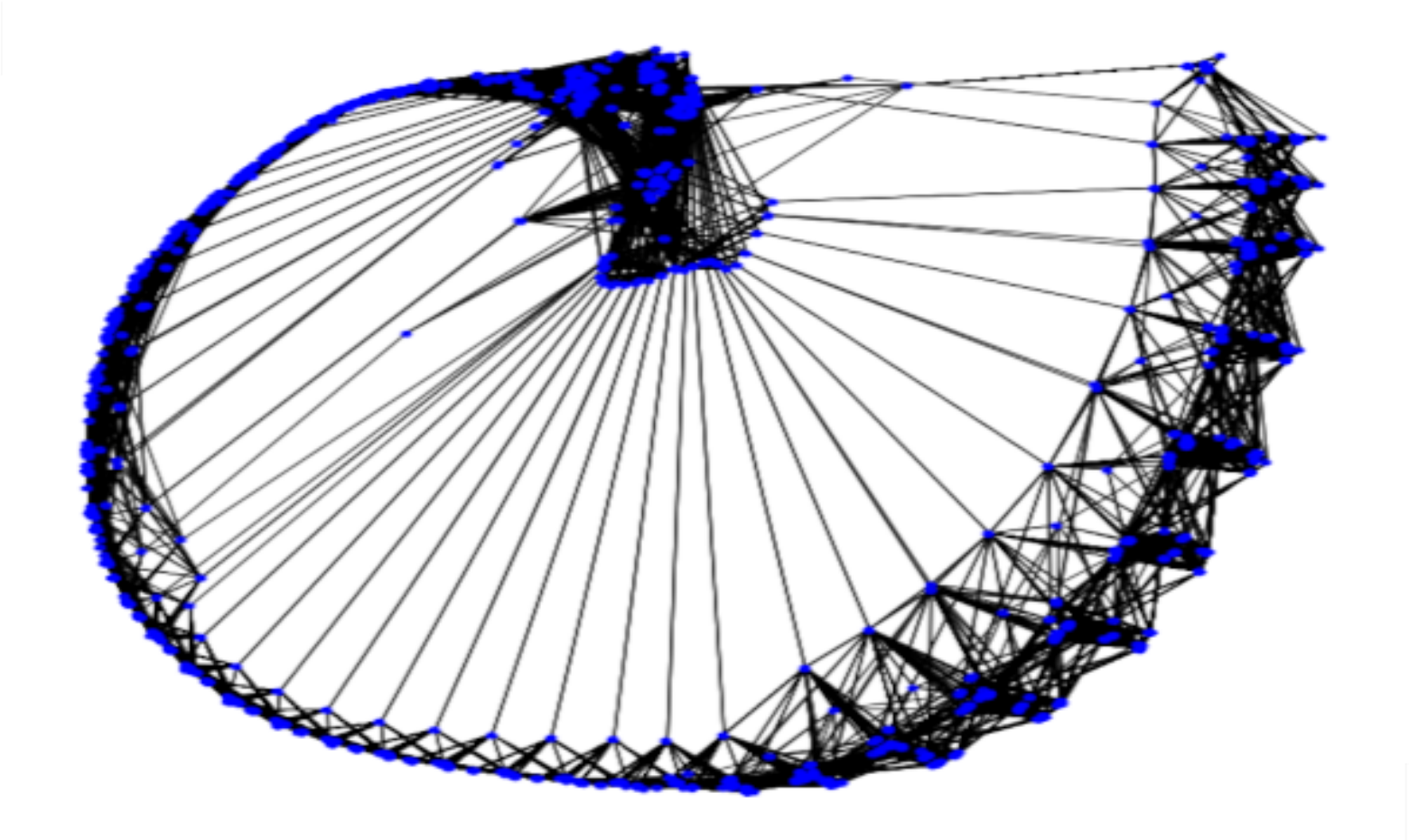}
         \includegraphics[width=0.24\columnwidth, trim = 20 20 20 20, clip]{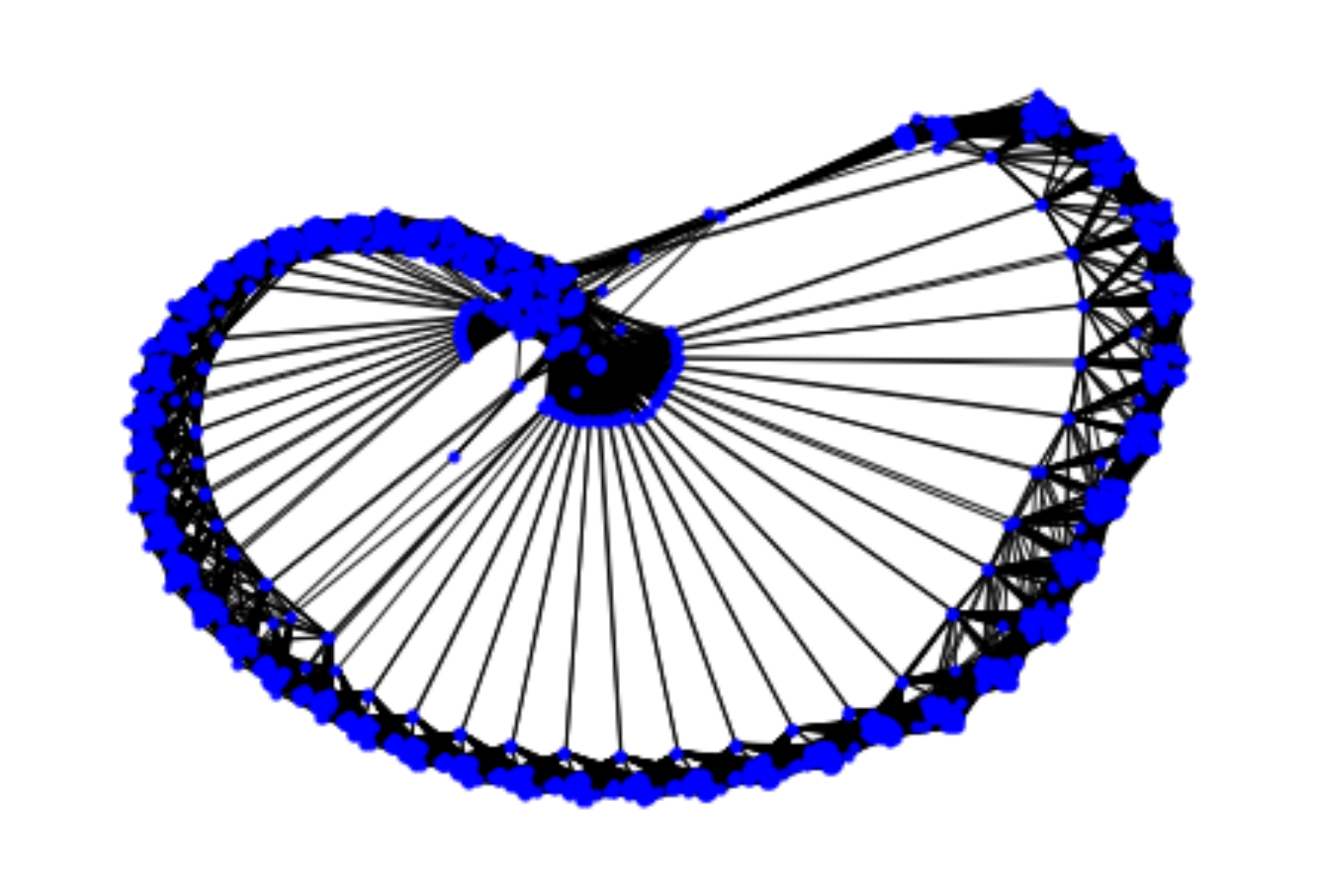}
         \includegraphics[width=0.24\columnwidth, trim = 20 20 20 20, clip]{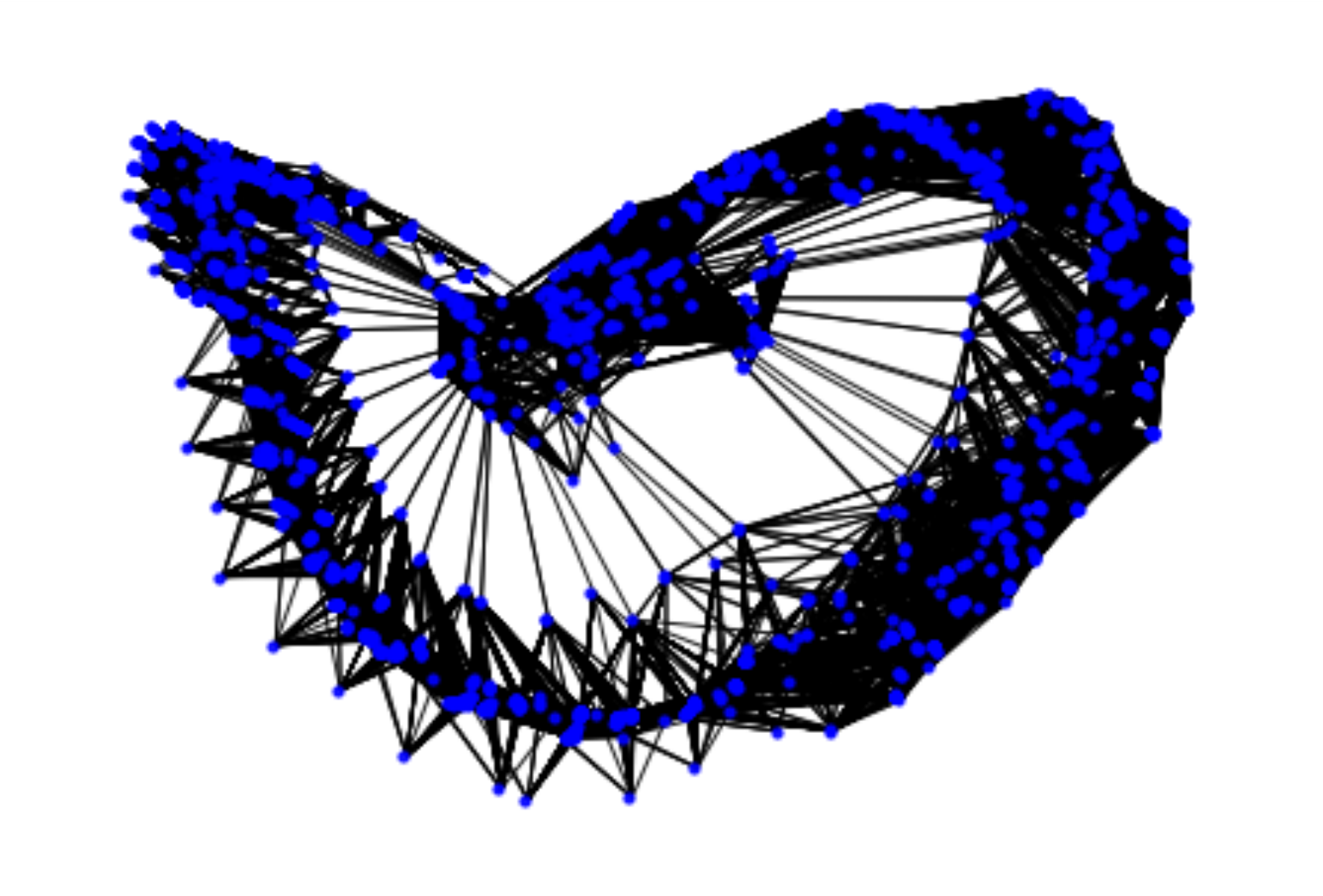}



    \caption{\mt{Large scale graphs from the SuiteSparse Matrix collection. Left to right: layouts produced by a GAT-based GND (trained to minimize stress on the Rome dataset), layout produced by the \texttt{sfdp} algorithm for large scale graphs and outcome of the PivotMDS method. We report for each row the name of the graph from the dataset collection, its order ($|\mathcal{V}|$) and size ($|\mathcal{E}|$).}}
    \label{fig:tamu}
\end{figure}

\mt{
\subsection{Scaling to bigger graphs}
\label{sec:bigger}
Common Graph Drawing techniques based on multidimensional scaling \cite{cox2008multidimensional} or SGD \cite{zheng2018graph} require ad-hoc iterative optimization processes for each graph to be drawn. Additionally, dealing with large scale graphs -- both in terms of number of nodes and number of involved edges -- decreases the time efficiency of these approaches. Conversely, once a GND has been learned, the graph layout generation consist solely in the extraction of Laplacian PE followed by a forward pass on the chosen GNN backbone. In this Section, we prove the ability of GND to scale to real-world graphs, providing quantitative results in terms of computational times and a qualitative analysis on the obtained graph layouts, with respect to SOTA Graph Drawing techniques. We employed the best performing GAT model trained to minimize the stress loss on the Rome dataset ( Section \ref{sec:stress}). We test the model inference performances on bigger scale graphs from the SuiteSparse Matrix Collection.\footnote{\url{https://sparse.tamu.edu}}  We report in Figure \ref{fig:timings} the computational times required by the different techniques to generate graph layouts of different scale,  from the \texttt{dwt\_n} graph family.  We analyze both the correlation on graph order (left -- varying number of nodes) and size (right -- varying number of edges). We compare the GND execution times against those of the NetworkX-GraphViz implementation of \texttt{neato} and \texttt{sfdp}, the latter being a multilevel force-directed algorithm that efficiently layouts large graphs. We also tested the Fruchterman-Reingold force-directed algorithm implemented in NetworkX (denoted with FR) and the PivotMDS implementation from the NetworKit \texttt{C++} framework \cite{staudt2016networkit}. The tests where performed in a Linux environment equipped with an Intel(R) Core(TM) i9-10900X CPU @ 3.70GHz, 128 GB of RAM and an NVIDIA GeForce RTX 3090 GPU (24 GB).
\begin{figure}
    \centering
    \includegraphics[width=0.99\columnwidth]{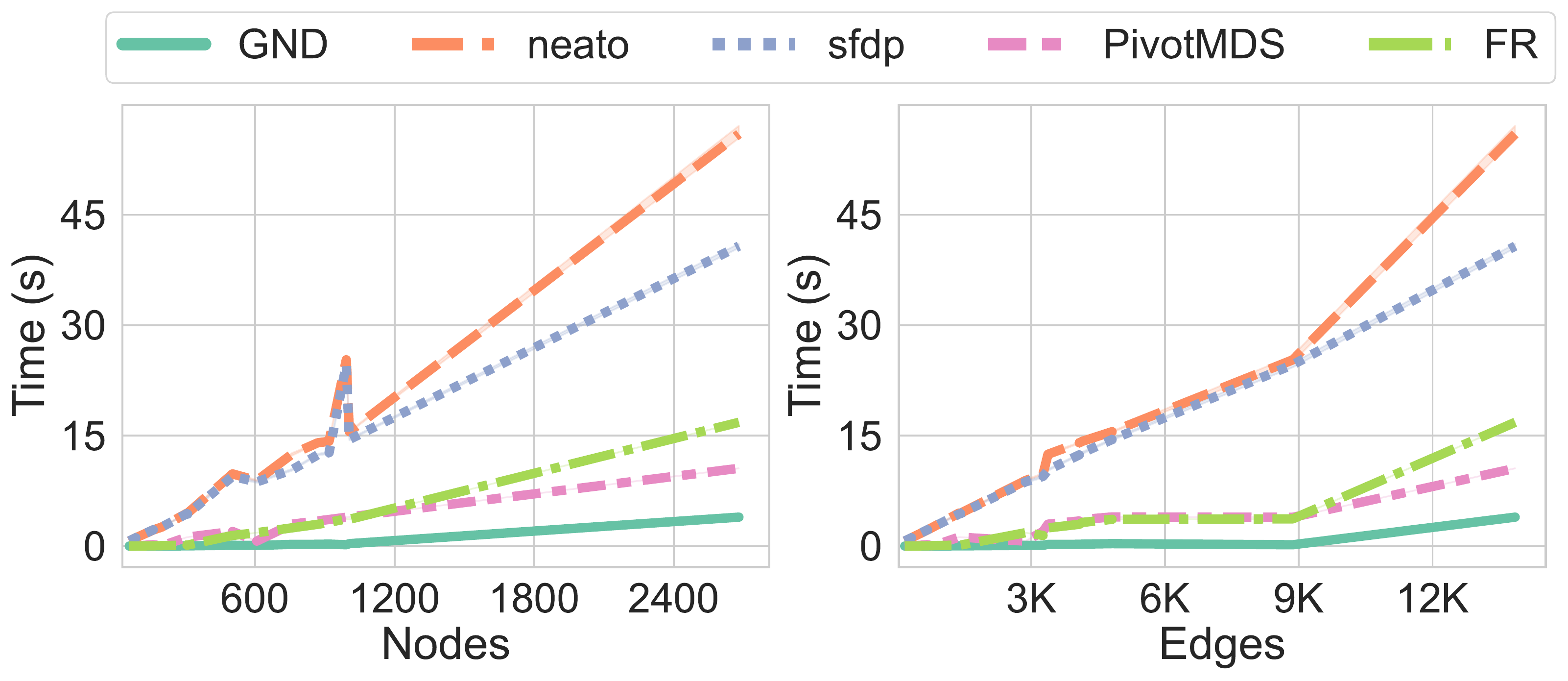}
    \caption{\mt{Computational time comparison on \texttt{dwt\_n} graphs. Left: correlation between the number of nodes in the graph and the layout generation timings of the analyzed Graph Drawing methods.  Right: correlation between number of edges available in the graphs and the corresponding layout generation timings.}}
    \label{fig:timings}
\end{figure}
We report the average execution times over three runs (we omit the variances due to their negligible values). 
These results confirm the advantages of the proposed approach. While all the competitors require expensive optimization process that increase their impact with bigger graph scale, the fast inference step carried on by GNDs assures small timings even with big graphs. Computing Laplacian PE is scalable and does not hinder the  time efficiency of the proposed method.  
To asses the quality of the generated layouts, we report in Figure \ref{fig:tamu} a comparison among the ones yielded by GND the framework, \texttt{sfdp} and PivotMDS on several graphs from the SuiteSparse collection (we report the graph name, its order $|\mathcal{V}|$ and size $|\mathcal{E}|$). While we remark that in this experiment we exploited a GND model trained on a smaller scale dataset (i.e., Rome), the performances show a significant ability of the model to generalize the learned laws (e.g., the stress minimization in this case) to unseen graphs, even when dealing with   diverging characteristics. However, we also remark that graphs having very diverse structures from the training distribution may be not correctly plotted. The causes of such performances drop are twofold. First, the intrinsic  dependance of neural models on the inductive biases learned during the training process leads to an inability to generalize to unseen graph topologies. On the other hand, the limitations of Laplacian PE to discriminate certain graph simmetries or structures \cite{dwivedi2020benchmarking} may be further compounded with larger scale datasets, which is an active area of research \cite{cui2021positional}.
}

\section{Conclusion}
\label{sec:conclusion}


Starting from some very interesting and promising results on the adoption of GNNs for graph drawing, which are mostly based on supervised learning, in this paper we proposed a general framework to emphasize the role of unsupervised learning schemes based on loss functions that enforce classic aesthetic measures. 
When working in such a framework, referred to as \textit{Graph Neural Drawers}, we open the doors towards the construction of a novel machine learning-based drawing scheme where the {\em Neural Aesthete} drives the learning of a GNN towards the optimization of beauty indexes.
While we have adopted the {\em Neural Aesthetes} only from learning to minimize arc intersections, the same idea can be used for nearly any beauty index.
We show that our framework is effective also for drawing unlabelled graphs. In particular, we rely on the adoption of Laplacian Eigenvector-based positional features~\cite{dwivedi2020benchmarking} for attaching information to the vertexes, which leads to very promising results.


%

\section*{Acknowledgment}
The authors would like to thank Giuseppe Di Battista for the insightful discussions and for useful suggestions on the Graph Drawing literature and methods.
This work was partly supported by EU Horizon 2020 project AI4Media, under contract no. 951911 (https://ai4media.eu/).

\ifCLASSOPTIONcaptionsoff
  \newpage
\fi



%
\bibliographystyle{IEEEtran}
\bibliography{refe.bib}

%




\end{document}